\pdfoutput=1

\documentclass[11pt]{article}
\usepackage{enumitem}
\usepackage{style/acl}\usepackage{todonotes}
\usepackage{float}

\usepackage{times}
\usepackage{latexsym}
\usepackage{graphicx}
\usepackage{adjustbox}
\usepackage{siunitx}
\usepackage{bbold}
\usepackage{array}
\usepackage{mfirstuc}
\usepackage{subcaption} 
\usepackage{longtable}
\usepackage{hyperref}
\usepackage{dutchcal}
\usepackage[acronym]{glossaries} 

\newglossarystyle{compactlist}{%
  \setglossarystyle{list}
    {\begin{description}[style=unboxed,leftmargin=2em,labelsep=0.5em,itemsep=0.05em]}%
    {\end{description}}%
}

\setglossarystyle{compactlist} 
\makeglossaries 

\usepackage[normalem]{ulem}
\usepackage{soul}
\usepackage{xcolor}
\usepackage{booktabs}
\usepackage{array} 
\usepackage{amsmath}
\usepackage{ulem}
\usepackage{cellspace}
\usepackage[utf8]{inputenc} 
\usepackage[T1]{fontenc}    
\usepackage{tabularx}    
\usepackage{multirow}
\usepackage{microtype}
\usepackage{longtable}

\usepackage{inconsolata}
\usepackage{listings}
\usepackage[most]{tcolorbox} 
\usepackage{todonotes}

\usepackage{subfiles} 

\usepackage[bb=ams]{mathalfa}

\makeatletter
\newcommand\footnoteref[1]{\protected@xdef\@thefnmark{\ref{#1}}\@footnotemark}
\makeatother

\title{
Discrepancy Detection at the Data Level:\\ Toward Consistent Multilingual Question Answering
}

\author{
\begin{tabular}{ccc}
\textbf{Lorena Calvo-Bartolomé}$^{C}$\thanks{Corresponding author: \texttt{lcalvo@pa.uc3m.es}} &
\textbf{Valérie Aldana}$^{M}$ &
\textbf{Karla Cantarero}$^{M}$ \\
\textbf{Alonso Madroñal de Mesa}$^{C}$ &
\textbf{Jerónimo Arenas-García}$^{C}$ &
\textbf{Jordan Boyd-Graber}$^{M}$ \\
\end{tabular} \\[0.75em]
\normalfont $^{C}$Universidad Carlos \textsc{III} de Madrid, Spain \quad
$^{M}$University of Maryland, College Park, USA
}

\newcommand{\mm}[0]{\abr{llm}}

\newcommand{\ling}[0]{\abr{mind}}
\newcommand{\pltm}[0]{\abr{pltm}}

\newcommand{\nd}[0]{\abr{nd}}
\newcommand{\con}[0]{\abr{c}}
\newcommand{\cd}[0]{\abr{cd}}
\newcommand{\nei}[0]{\abr{nei}}
\newcommand{\mallet}[0]{\abr{mallet}}

\definecolor{customblue}{RGB}{0, 84, 168}
\definecolor{customred}{RGB}{149, 55, 28}
\definecolor{promptBlue}{RGB}{186,225,255} 
\definecolor{promptGreen}{RGB}{186,255,201}   
\definecolor{promptRed}{RGB}{255,179,186}
\definecolor{contradiction}{HTML}{F95454}  
\definecolor{discrepancy}{HTML}{77CDFF}    
\definecolor{lgray}{RGB}{176, 179, 184}
\definecolor{tableblue}{HTML}{D9EAFD}
\definecolor{tablegray}{HTML}{F5F5F5}
\definecolor{lightbluegray}{RGB}{230, 236, 240}  
\definecolor{darkbluegray}{RGB}{70, 90, 110}     
\definecolor{yesbluegray}{RGB}{0, 85, 160}       
\definecolor{nobluegray}{RGB}{150, 40, 40}       
\definecolor{noblueorange}{RGB}{241, 120, 110}  

\definecolor{blueoutside}{RGB}{44, 84, 252}
\definecolor{blueinside}{RGB}{212, 227, 252}
\definecolor{bluetheother}{RGB}{150, 175, 252}
\definecolor{blue1}{HTML}{A2B9E0}
\definecolor{blue2}{HTML}{B5C8E7}
\definecolor{blue3}{HTML}{C8D6ED}

\newtcolorbox[list inside=prompt,auto counter,number within=section]{prompt}[1][]{
    colbacktitle=black!60,
    fonttitle=\bfseries\small,
    colbacktitle=blue1,
    coltitle=black,
    colback=blue3!10, 
    fontupper=\footnotesize,
    boxsep=3pt,
    left=0pt,
    right=0pt,
    top=0pt,
    bottom=0pt,
    boxrule=1pt,
    enhanced jigsaw,  
    breakable,    
    #1,
}

\newif\ifcomment\commenttrue

\usepackage[a-1b]{pdfx}

\usepackage{framed}
\usepackage{mdwlist}
\usepackage{siunitx}
\usepackage{latexsym}
\usepackage{colortbl}
\usepackage{xcolor}
\usepackage{nicefrac}
\usepackage{booktabs}
\usepackage{fnpct}
\usepackage{amsfonts}
\usepackage[T1]{fontenc}
\usepackage{bold-extra}
\usepackage{amsmath}
\usepackage{amssymb}
\usepackage{bm}
\usepackage{graphicx}
\usepackage{mathtools}
\usepackage{microtype}
\usepackage{multirow}
\usepackage{multicol}
\usepackage{xpatch}
\usepackage{latexsym,comment}
\usepackage[normalem]{ulem}

\newcommand*{\missingreference}{{\Huge \colorbox{red}{?reference?}}}
\newcommand*{\missingcitation}{{\Huge \colorbox{red}{?citation?}}}

\makeatletter
\xpatchcmd{\@setref}{\bfseries}{\missingreference}{}{}
\def\@citex[#1]#2{\leavevmode
    \let\@citea\@empty
    \@cite{\@for\@citeb:=#2\do
        {\@citea\def\@citea{,\penalty\@m\ }%
            \edef\@citeb{\expandafter\@firstofone\@citeb\@empty}%
            \if@filesw\immediate\write\@auxout{\string\citation{\@citeb}}\fi
            \@ifundefined{b@\@citeb}{\hbox{\reset@font\missingcitation}%
                \G@refundefinedtrue
                \@latex@warning
                {Citation `\@citeb' on page \thepage \space undefined}}%
            {\@cite@ofmt{\csname b@\@citeb\endcsname}}}}{#1}}
\makeatother

\newcommand{\gem}[1]{\mbox{\textsc{gem}}}
\newcommand{\abr}[1]{\textsc{#1}}

\newcommand{\lda}{\abr{lda}}

\renewenvironment{quote}
{\list{}{\rightmargin\leftmargin}%
    \item\relax\small\ignorespaces}
{\unskip\unskip\endlist}

\newcommand{\hidetext}[1]{}
\newcommand{\ignore}[1]{}

\ifcomment
    \newcommand{\pinaforecomment}[3]{\colorbox{#1}{\parbox{.8\linewidth}{#2: #3}}}

    \newcommand{\prtodo}[1]{\pinaforecomment{lightblue}{pr}{#1}}
    \newcommand{\prtodoi}[1]{\pinaforecomment{lightblue}{pr}{#1}}
\else
    \newcommand{\pinaforecomment}[3]{}
    \newcommand{\prtodo}[1]{}
    \newcommand{\prtodoi}[1]{}
\fi

\newcommand{\smallurl}[1]{ \begin{tiny}\url{#1}\end{tiny}}

\definecolor{lightblue}{HTML}{3cc7ea}
\definecolor{CUgold}{HTML}{CFB87C}
\definecolor{grey}{rgb}{0.95,0.95,0.95}
\definecolor{ceil}{rgb}{0.57, 0.63, 0.81}
\definecolor{UMDred}{HTML}{ed1c24}
\definecolor{UMDyellow}{HTML}{ffc20e}

\begin{document}

\maketitle
\begin{abstract}
Multilingual question answering (QA) systems must ensure factual consistency across languages, especially for objective queries such as {\it What is jaundice?}, while also accounting for cultural variation in subjective responses. We propose \ling{}, a user-in-the-loop fact-checking pipeline to detect factual and cultural discrepancies in multilingual QA knowledge bases. \ling{} highlights divergent answers to culturally sensitive questions (e.g., {\it Who assists in childbirth?}) that vary by region and context. We evaluate \ling{} on a bilingual QA system in the maternal and infant health domain and release a dataset of bilingual questions annotated for factual and cultural inconsistencies. We further test \ling{} on datasets from other domains to assess generalization. In all cases, \ling{} reliably identifies inconsistencies, supporting the development of more culturally aware and factually consistent QA systems.
\footnote{\texttt{\url{http://github.com/lcalvobartolome/mind}} contains all datasets and models used in our experiments, as well as a package to run \ling{} on new datasets.}

\end{abstract}

\section{Introduction} \label{sec:sections/10-intro} \begin{figure}[!ht]
    \centering
    \includegraphics[width=\columnwidth]{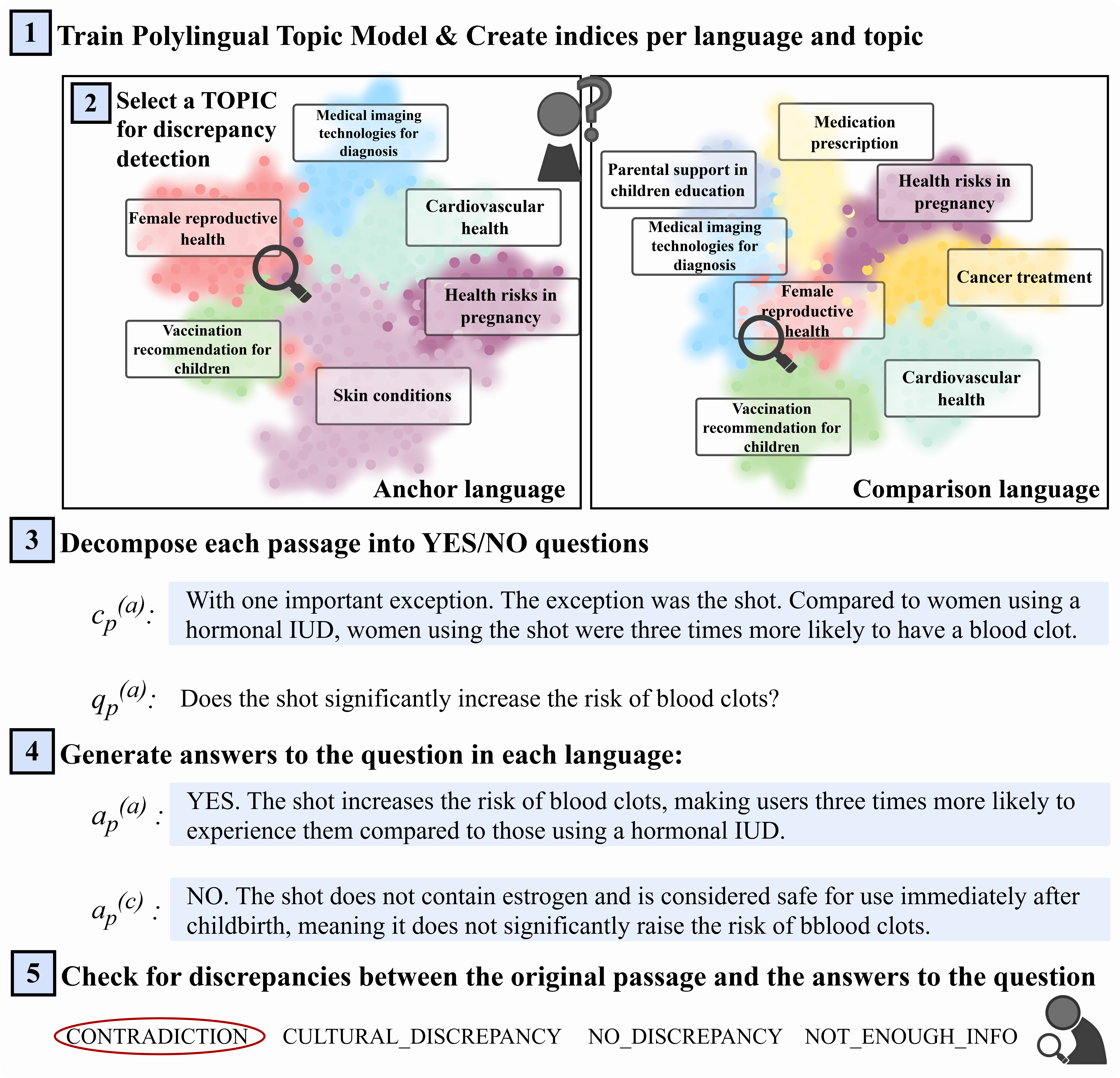}
    \caption{\ling{} overview. (\hyperref[subsec:pltm]{1}) Passages are aligned using topic modeling. (\hyperref[subsec:indexing]{2}) A topic is selected. (\hyperref[subsec:question_generation]{3})–(\hyperref[subsec:discrepancy_detection]{5}) An \mm{} aids question decomposition (\hyperref[subsec:question_generation]{3}), answering (\hyperref[subsec:retrieval]{4}), and discrepancy detection (\hyperref[subsec:discrepancy_detection]{5}).}
    \label{fig:topline}
\end{figure}

Multilingual QA systems face the dual challenge of ensuring factual accuracy while respecting cultural relevance. A correct answer in one language may not meet expectations in another due to differences in data availability, cultural practices, or local nuance. Although language and culture often overlap, they sometimes also misalign---cultural practices can vary significantly within the same language or span across multiple languages~\cite{hovy2021importance, hershcovich2022challenges}. The complexity grows when sources conflict, offering divergent information that leads to inconsistencies in model responses~\cite{palta2022investigating}.

In monolingual settings, several studies have explored how models should handle conflicting knowledge sources~\cite{pan2021contraqa,chen2022rich,park2024enhancing,kamath2020selective}. Here, we argue for a proactive approach: addressing gaps and inconsistencies upstream in the knowledge base simplifies the model's task and prevent inconsistencies from manifesting in model responses.

To identify these {\bf discrepancies}, we draw on previous work that incorporates question generation and answering into fact-checking systems~\cite{setty2024questgen,chen2023complex,schlichtkrull2023averitec}
through a claim-centric pipeline, where a known claim is verified through evidence retrieval and analysis. In contrast, our focus is on multilingual discrepancy detection, where claims are not predefined and divergences arise from factual or cultural differences across answers generated using language-specific information.
We introduce \ling{} (Multilingual Inconsistent Notion Detection), a four-stage \mm{}-aided pipeline that mirrors QA rather than classical fact-checking 
(see Fig.~\ref{fig:topline}). 
Distinctively, \ling{} begins by aligning multilingual documents using polylingual topic modeling~\citep[\pltm{},][]{mimno2009polylingual}. This alignment structures documents within a shared conceptual space, directing attention to relevant claims. We then detect discrepancies across languages by evaluating an answer's faithfulness in one language against its interpretation in another.


We focus on maternal and infant health, applying \ling{} to the bilingual QA dataset behind Rosie~\citep{mane2023practical}, a chatbot built independent from English and Spanish knowledge bases (\S\hyperref[sec:ablation]{\ref{sec:ablation}}). 
Although independent medically sound information should not contradict, in practice, the English and Spanish data often present incomplete, conflicting, or culturally divergent guidelines. This raises the question: {\em when both answers are grounded in credible evidence, which one should we trust?} 
We construct \texttt{ROSIE-MIND}, a collection of English–Spanish QA pairs labeled for discrepancies, serving as a multilingual QA benchmark.

To evaluate generalization, we apply \ling{} to two datasets: \texttt{FEVER-DPLACE-Q}, a controlled set with implicit discrepancies used in ablation studies of two-answer–per-question detection (\S\hyperref[sec:ablation]{\ref{sec:ablation}}), and \texttt{WIKI-EN-DE}, derived from aligned English–German Wikipedia pages, extending evaluation beyond English–Spanish and outside health (\S\hyperref[sec:mind_discrpancies]{\ref{sec:mind_discrpancies}}).  
In both cases, \ling{} surfaces inconsistencies beyond the original domain. 

%


 \label{sec:introduction}

\section{A data-level approach for cross-lingual discrepancy detection} \label{sec:sections/20-data_level_approach} 




Addressing contradictions in multilingual QA systems requires more than passively detecting discrepancies in answers; it necessitates a proactive approach to identify and address \emph{bad} advice before it reaches users. 
%
%
By \emph{bad} advice, we refer not only to factual inaccuracies, but also to inconsistencies in system responses when \emph{semantically equivalent user questions} yield conflicting answers. 

By definition, a contradiction occurs when two statements cannot be both true. However, in multilingual QA, contradictions often arise more subtly; \citet{de2008finding} proposes a more {\it loose} definition that aligns with human intuitions: {\it ``contradiction occurs when two sentences are extremely unlikely to be true simultaneously''}.
We adopt this broader view while distinguishing between
{\it contradictions} (direct factual opposition)
and {\it cultural discrepancies} (statements from different cultural or epistemological frameworks with conflicting meanings). 
We use the term \emph{cultural discrepancy} broadly: it covers {\em not only differences in values or worldviews} but {\em also variations in practices, policies, or institutional norms}, since these are culturally embedded and carry epistemic assumptions. 
Because discrepancies are varied and rooted in different frameworks, they cannot easily be resolved within a single epistemic frame. Identifying and interpreting them requires careful, bespoke analyses rather than mere intuition.
%
Our primary foundation is Hymes' theory of communicative competence~\citep{hymes1972communicative}: language is not only grammatical knowledge, as it encompasses different sociocultural knowledge and practice. 
We use what is considered valid knowledge~\citep{spivak2023can} to expose epistemic asymmetries~\citep{fanon1963wretched}.

For example, the following passages~\citep{mane2023practical} lead to opposing conclusions about the safety of the shot: 
\begin{quote}
\textbf{P1:} 
``The exception was the shot. Compared to women using a hormonal IUD, women on the \ul{shot were three times more likely to have a blood clot}.''

\textbf{P2:} ``Not all methods are safe for new moms. While hormonal methods that don't contain estrogen—\ul{the shot, the Mirena IUD, the implant and the mini-pill—are safe for women to use immediately after giving birth}[...].''
\end{quote}



A cultural discrepancy appears in the following childbirth descriptions~\cite{culturalBirthingPractices2020}:\footnote{
P1 describes Afghan home births with grandmothers; P2 reflects Burma's preference for Caesareans.
} 
\begin{quote}
\textbf{P1:} ``\ul{Most women give birth at home}; [...]
Usually \ul{a grandmother delivers the babies}.'' 

\textbf{P2:} ``\ul{Most women are encouraged to have Caesareans} and the doctor gives the date when to turn up at the hospital.''
\end{quote}

Given two responses, contradiction detection in Natural Language Inference (NLI) evaluates whether a hypothesis contradicts, entails, or is neutral given a premise. This setup, however, does not extend to multilingual QA, where 
evidence comes from diverse sources that may conflict, and documents are rarely thematically aligned or consistently structured across languages.
%

By systematically aligning information across languages and categorizing inconsistencies, we can reduce the likelihood of misleading responses. 
By surfacing these differences,
rather than allowing them to be masked or misinterpreted as contradictions,
we also detect
information gaps, where relevant data is missing in one language.
 \label{sec:data_level_approach}

\section{\textsc{MIND}: Multilingual Inconsistent Notion Detection} \label{sec:sections/40-pipeline} 
We designate the {\it anchor} corpus as the majority corpus
serving as the primary reference, while {\it comparison} corpora are analyzed against it. 
Importantly, we do not assume direct alignment between documents in the anchor and comparison corpora. Instead, we follow the assumption in \pltm{}~\citep{mimno2009polylingual} that documents in a tuple share the same topic distribution even if they are not translations of each other,
supporting {\it ``tuples of documents that are loosely equivalent to each other, but written in different languages''}.
In our setup, such loose alignments are created either from existing comparable corpora or via machine translation\footnote{Translation quality is not critical; it serves to establish topic-space alignment, then translations are discarded.}. 

\ling{} first organizes the 
corpora into a shared thematic space 
(\S\hyperref[subsec:pltm]{\ref{subsec:pltm}}), ensuring that topics are consistent across languages (\S\hyperref[appen:topics]{\ref{appen:topics}}), and clusters data by topic (\S\hyperref[subsec:indexing]{\ref{subsec:indexing}}) to focus comparisons on semantically related content. 
%
After topic alignment, \ling{} organizes documents by theme, and a \emph{user-in-the-loop} 
discards topics that are noisy or off-domain (e.g., web artifacts or {\it garbage} topics). 
The system then generates questions from {\it anchor} corpus passages for remaining topics (\S\hyperref[subsec:question_generation]{\ref{subsec:question_generation}}) and refines them into search queries (\S\hyperref[subsec:retrieval]{\ref{subsec:retrieval}}) to retrieve relevant content from the {\it comparison} corpora. The anchor-language answer comes from the passage that generated the question, while comparison-language answers are based on retrieved evidence (\S\hyperref[subsec:answer_generation]{\ref{subsec:answer_generation}}). We then analyze the relationship between answers given the question (\S\hyperref[subsec:discrepancy_detection]{\ref{subsec:discrepancy_detection}}). {\color{black} Finally, users review flagged discrepancies to confirm their validity.}

For simplicity in notation, we assume a bilingual setting with a single comparison corpus, but this approach naturally extends to multiple comparison corpora. Let:
\begin{align*}
    {\mathcal{C}}^{(a)} &= \{ \mathcal{c}_d^{(a)} \mid d = 1, \dots, D_a \}\\
    {\mathcal{C}}^{(c)} &= \{ \mathcal{c}_d^{(c)} \mid d = 1, \dots, D_c \}
\end{align*}
be the {\it anchor} (\(a\)) and {\it comparison} (\(c\)) corpora, respectively, where each document is segmented into a variable number of passages, 
resulting in a total of \(P^{(a)}\) and \(P^{(c)}\) passages: 
\begin{align*}
    C^{(a)} &= \{ c_p^{(a)} \mid p = 1, \dots, P^{(a)} \}\\
    C^{(c)} &= \{ c_p^{(c)} \mid p = 1, \dots, P^{(c)} \}.
\end{align*}
Calligraphic \(\mathcal{C}\) denotes complete documents, while regular \(C\) represent passages. 
We define \(\mathcal{D}(\cdot)\) as a function mapping passages to documents, e.g., \(\mathcal{D}(c_p^{(a)}) = \mathcal{c}_d^{(a)}\) indicates that \(\mathcal{c}_d^{(a)}\) is the document from which \(c_p^{(a)}\) originates.

\subsection{Polylingual Topic Modeling}\label{subsec:pltm}
\pltm{} trains a topic model using passages from 
\(C^{(a)}\) and \(C^{(c)}\), and their respective {\it loosely aligned} translations when $C^{(a)}$ and $C^{(c)}$ are not already comparable corpora, 
and provide two outputs:
\begin{enumerate}
    \item \textbf{Per-language word-topic distributions.} For each topic \(t_k, \; k = 1, \dots, K\), 
    \begin{align*}
        \boldsymbol{\beta}_k^{(a)} &= [\beta_{k,v}^{(a)} \mid v = 1, \dots, V^{(a)}],\\
        \boldsymbol{\beta}_k^{(c)} &= [\beta_{k,v}^{(c)} \mid v = 1, \dots, V^{(c)}].
    \end{align*}
    Each \(\beta_{k,v}^{(a)}\) is the probability of word \(w_v^{(a)}\) under topic \(t_k\).
    Since each language has its own vocabulary, topics are defined over different word distributions across languages.
    \item \textbf{Topic distributions.} 
    The topic representation of passages is \(\boldsymbol{\theta}_p^{(a)}\)
    , where 
    \(\theta_{p,k}^{(a)}\) 
    is the proportion of topic \(t_k\) in passage \(c_p^{(a)}\). The same definition applies to \(\boldsymbol{\theta}_p^{(c)}\).
\end{enumerate}

\subsection{Topic-based clustering}\label{subsec:indexing}
\ling{}
retrieves relevant passages from the comparison corpus by first filtering based on topic relevance.
The active passages for each topic \(t_k\) are:\footnote{A similar definition applies if analyzing discrepancies within the anchor corpus.}
\[
\mathcal{T}_{k}^{(c)} = \{ c_p^{(c)} \mid \theta_{p,k}^{(c)} > 0, \; p = 1, \dots, P^{(c)} \},
\]
Approximate (ANN) or exact nearest neighbors (ENN) searches are used to identify relevant passages within each topic. When ANN is applied, the set \(\mathcal{T}_{k}^{(c)}\) must be partitioned 
into \(\ell_k^{(c)}\) clusters,
\begin{equation}\label{eq:nr_clusters}
    \ell_k^{(c)} = \max \left( \left\lfloor \lambda \sqrt{\mid \mathcal{T}_{k}^{(c)} \mid} \right\rfloor, \ell_{min} \right),
\end{equation}
where \(\lambda\) and \(\ell_{min}\) are customizable parameters ensuring a minimum number of clusters.

%

\subsection{Questions generation} \label{subsec:question_generation}
Taking the anchor corpus as starting generation point, we assign each passage \(c_p^{(a)}\) to the topic \(t_k\) it is most strongly associated with based on the arg max of \(\boldsymbol{\theta}^{(a)}_p\).\footnote{Note that this assignment associates each passage with a single topic for generation purposes, but at query time, its topic-weighted representation is used (see \S\hyperref[appen:topic_distr_retrieval]{\ref{appen:topic_distr_retrieval}}).}
Each passage \(c_p^{(a)}\) is then represented by a series of \abr{yes/no} questions~\cite{chen2022generating}, \(\mathcal{Q}_p^{a} = \{q_{p,n}^{(a)}, n=1,\dots,N\}\), where all questions in \(\mathcal{Q}_p^{(a)}\) share the same evidence \(r_p^{(a)} = (c_p^{(a)},\mathcal{D}(c_p^{(a)}))\),
ensuring that all questions \(\mathcal{Q}_p^s\) are grounded in \(c_p^{a}\) while preserving their document-level context \(\mathcal{D}(c_p^{(a)})\)~\citep{choi2021decontextualization}. 

These question-evidence pairs \((q_p^{(a)}, r_p^{(a)})\) are the basis for detecting discrepancies in \(C^{(c)}\). 
We use a few-shot prompting strategy~\cite{chen2024complex,brown2020language,ousidhoum2022varifocal}, 
but rather than focusing on fact-checking, we prioritize generating questions that users might naturally ask for information-seeking purposes. 

To ensure relevance, the \mm{} assesses before generating the questions whether the passage contains objective information
---excluding author affiliations; people experiences, opinions; or subjective content such as {\it ``I'm a single mom, and homeschooling was hard for my kids''} (\ref{prompt:question_generation}).
As an additional filter, we adopt a similar approach to that of~\citet{ki-etal-2025-askqe}, employing an off-the-shelf NLI classifier~\citep{manakul-etal-2023-selfcheckgpt} to discard questions whose answers are not entailed by the anchor passage, i.e., those labeled as contradictory with respect to it.

\subsection{Topic-based Retrieval}\label{subsec:retrieval}
Since questions in \(\mathcal{Q}_p^{(a)}\) are grounded on evidence set \(r_p^{(a)}\), we can generate answers in the anchor language, but we need to retrieve relevant passages from \(\mathcal{C}^{(c)}\) for the comparison language. 

Rather than querying solely with \(\mathcal{Q}_p^{(a)}\), 
a multi-hop question answering approach~\citep{qi2019answering}
prompts an \mm{} to decompose each \(q_n \in \mathcal{Q}_p^{(a)}\) into a sequence of \(M\) queries \(\mathcal{S}_n=\{s_{n,m}, 1,\dots, M\}\) (\ref{prompt:query_generation}). Here, each \(s_{n,m}\) is a reformulated query incorporating contextual disambiguation, e.g., {\it Multisystem Inflammatory Syndrome in Children (MIS-C)} instead of {\it MIS-C}.

Given a search query \(s_{n,m}\) and the \(\boldsymbol{\theta}_p^{(a)}\) of the passage \(c_{p}^{(a)}\) that generated it, we retrieve passages from \(\mathcal{C}^{(c)}\) using either ANN or ENN searches within the most relevant topic clusters,
\begin{equation}
\mathcal{T}_{k}^{(c),\mathrm{rel}}
= \{\mathcal{T}_k^{(c)}:\theta^{(a)}_{p,k}>\epsilon\}, \; k=1,\dots,K.
\end{equation}
For each relevant topic \(t_k\), we compare the query embedding \(e({s_i})\) against the passages in \(\mathcal{T}_{k}^{(c),\mathrm{rel}}\) and retrieve the top-\(H\) nearest neighbors. Each retrieved \(c_{p}^{(c)}\) receives the score:  
\begin{equation}
    S(c_{p}^{(c)}) = \alpha \cdot \operatorname{sim}(e(s_i), e(c_{p}^{(c)})),
\end{equation}
where \(\operatorname{sim}(e(s_i), e(c_{p}^{(c)}))\) is the cosine similarity between the query embedding \(e(s_i)\) and the target passage embedding \(e(c_{p}^{(c)})\). Weight $\alpha = \theta^{(a)}_{p,k}$ if weighted similarity is applied; otherwise $\alpha = 1$.
For each $s_{n,m}$, we deduplicate by passage---it may appear under multiple topics---and rank candidates globally, keeping the top-$L$. Across all $M$ subqueries for a given anchor question, we merge and deduplicate the retrieved passages to form the final evidence set $\mathcal{R}_p^{(c)} = \{r_{p,\ell}^{(c)}, \ell=1,\dots,L'\}$, where $L'$ is the number of unique relevant passages.
We call this approach topic-based retrieval and distinguish between Topic-based ENN (TB-ENN) and Topic-based ANN, which apply ENN/ANN searches when retrieving the top-\(H\) nearest neighbors for each topic.

\subsection{Answer generation} \label{subsec:answer_generation}
We pair each question $q_p^{(a)}$ with its in-corpus evidence $r_p^{(a)}$ and with each retrieved passage in $\mathcal{R}_p^{(c)}$. This yields one anchor-side answer $a_p^{(a)}$ and a set of comparison-side answers $\{a_{p,\ell}^{(c)}, \ell=1, \cdots, L'\}$, each generated by prompting an \mm{} with the question and corresponding evidence (\ref{prompt:answer_generation}).
To ensure answers remain grounded in context, the model is instructed to abstain from responding if the passage does not contain sufficient information. 
We pose all questions in the anchor language\footnote{This controlled setup allows us to isolate evidence-based inconsistencies. In production, questions may differ across languages, introducing further variance not addressed here.}, regardless of the comparison corpus language. This avoids rephrasing or translation artifacts, ensuring that inconsistencies arise from the retrieved evidence, not from semantic drift in question formulation.

\subsection{Discrepancy detection} \label{subsec:discrepancy_detection}
We prompt an \mm{} (\ref{prompt:discrepancy_detection}) to determine whether 
answer \(a_p^{(a)}\) entails (\abr{no\_discrepancy}, \nd{}), contradicts 
(\abr{contradiction}, \con{}), or differs due to a cultural discrepancy 
(\abr{cultural\_discrepancy}, \cd{}) with \(a_p^{(c)}\), given \(q_p^{(a)}\). 
We also allow classifying them as \abr{not\_enough\_info} (\nei{}) if there is 
not enough information in \(C^{(c)}\) to answer the question. \label{sec:pipeline}

\section{Ablation Studies} \label{sec:sections/50-ablation} Due to \ling{}’s pipeline structure, overall quality---and even the feasibility of detecting discrepancies---depends heavily on individual components. We therefore begin with ablation studies to assess each component in isolation, describing our dataset choices, topic modeling configuration, evaluation metrics, and results.

\subsection{Datasets}\label{subsec:datasets}
We use two datasets for these ablations: our primary dataset built upon Rosie~\citep{mane2023practical}, and a synthetic dataset designed to contrast observed discrepancies with controlled ones.

\paragraph{\texttt{ROSIE}} The knowledge base of~\citet{mane2023practical}, with English and Spanish documents on maternal and infant health segmented into passages. We remove passages with bibliographic references, citations, or personal experiences (\S\hyperref[appen:filtering]{\ref{appen:filtering}}). 
This yields \num{542055} English passages (anchor corpus \(\mathcal{C}^{(a)}\)) and \num{333175} Spanish passages (comparison corpus \(\mathcal{C}^{(a)}\)).
Each corpus is translated into the other language using \texttt{OPUS-MT}~\cite{tiedemann2020opus}. 
All passages undergo language-specific NLP preprocessing (\S\hyperref[appen:preprocessing_details]{\ref{appen:preprocessing_details}}).

{\bf \texttt{FEVER-DPLACE-Q}} We construct a controlled dataset using \texttt{gpt-4o} with explicit entailments and discrepancies. First, we convert 50 REFUTES and 50 SUPPORTS FEVER-v1 claims~\cite{thorne-etal-2018-fever} into question-answer triplets, ensuring one answer supports / contradicts the claim (\ref{prompt:fever_conversion}). 
For example, given the FEVER claim {\it ``Beautiful is a 2000 robot''}, we generate {\it ``Is Beautiful a 2000 robot?'' / ``Yes, Beautiful is a 2000 robot.'' / ``No, Beautiful is a 2000 American comedy-drama film directed by Sally Field.''}.
%
Next, we adapt 50 D-PLACE~\cite[Ethnographic Atlas,][]{kirby2016d} definitions, mapping categorical codes to cultural discrepancies (\ref{prompt:dplace_conversion}). For example, given the dimension {\it ``Age or occupational specialization in the manufacture of earthenware utensils''} and its first two codes, we construct {\it ``Is the creation of earthenware utensils typically done by older adults beyond their prime?'' / ``No, it is primarily done by boys and girls before puberty.'' / ``Yes, it is mainly performed by older adults beyond their prime.''}. 
We additionally generate 35 \nei{} samples using D-PLACE definitions where one code is ``Missing data''. The final dataset comprises 185 samples manually reviewed by the authors.

\subsection{Topic-based retrieval configuration} 
\paragraph{Topic modeling} We train \texttt{ROSIE} using \mallet{}~\cite{mallet}'s \pltm{} with default parameters, except setting $K=30$ (\S\hyperref[appen:k_optimization]{\ref{appen:k_optimization}}).
We further prompt an \mm{} (\ref{prompt:labeling}) with the most probable words and representative documents to obtain a descriptive label for each topic~\citep{pham-etal-2024-topicgpt}. 
Evaluating all \texttt{ROSIE} passages would be cost prohibitive, 
so we report results for {\it Pregnancy} (\(t_{12}\)) and {\it Infant Care} (\(t_{16}\)). 

\paragraph{(TB)-ENN/ANN} Passages are encoded using \texttt{BAAI/bge-m3}~\citep{bge-m3}, which is optimized for asymmetric and cross-lingual retrieval. (TB)-ENN/ANN searches are implemented with FAISS~\cite{douze2024faiss} Inverted File Index (IVF). For TB-ANN, each set of active topics \(\mathcal{T}_{k}^{(c)}\) is divided into \(\ell_k^{(c)}\) clusters (Eq.~\ref{eq:nr_clusters}) with $\lambda = 4$ and $\ell_{\min} = 8$. The number of probed clusters in ANN searches is set as $\max\bigl(1, \, \lfloor 0.10 \cdot \max(1, \ell) \rceil \bigr)$. 
For the ANN baseline, the number of clusters is set equivalently to the number of clusters in the topic sets, i.e.,
\begin{align}
\ell = \max \left( \left\lfloor \lambda \sqrt{\mid N \mid} \right\rfloor, \ell_{min}\right),
\end{align}
where N is the number of data points. 
We explore different configurations of TB-ENN/ANN. Specifically, we vary: (i) the threshold $\epsilon$, using either a fixed value of 0 (static, S) or an automatically selected value per topic $t_k$ via an elbow-detection algorithm on $\boldsymbol{\theta}^{(c)}_p$ (dynamic, D); (ii) the number of retrieved passages, $L \in \{3, 5\}$; and (iii) whether to apply $\theta$-weighted similarity. 

\paragraph{LLMs} We use \texttt{qwen:32b}, \texttt{llama3.3:70b}, and \texttt{gpt-4o} (\S\hyperref[appen:infrastructure]{\ref{appen:infrastructure}}). \texttt{70B-Llama3-instruct} and \texttt{8B-Llama3.1-instruct}, together with the first two, also serve as judges for obtaining gold passages for retrieval.  
We set temperature to 0, \texttt{top\_p} to 0.1, \texttt{frequency\_penalty} to 0, and fix a random seed for reproducibility.

\subsection{Metrics}\label{subsec:metrics}
Below we detail the metrics for each step in the pipeline. When human judgments were required, Prolific crowdworkers with graduate or doctorate degrees in Health \& Welfare provided annotations, for £12/hour, with at least three annotators per task.

\paragraph{Questions \& answers.}  We assess questions on six criteria---{\bf V}erifiability (V), {\bf P}assage {\bf I}ndependence (PI), {\bf C}larity (C), {\bf T}erminology (T), {\bf S}elf-{\bf C}ontainment (SC), and {\bf N}aturalness (N)---and answers on five---{\bf F}aithfulness (F), {\bf P}assage {\bf D}ependence (PD), {\bf P}assage {\bf R}eference {\bf A}voidance (PRA), {\bf S}tructured {\bf R}esponse (SR), and {\bf L}anguage {\bf C}onsistency (LC) (\S\hyperref[appen:criteria_question_answer_quality]{\ref{appen:criteria_question_answer_quality}}). Annotators assign a point for each criterion met.

\paragraph{Retrieval.} We benchmark TB-ENN/ANN against ENN/ANN on retrieval time and metrics: Recall, Precision, Multiple Mean Reciprocal Rank~\citep[MMRR, ][]{kachuee2025improving}, and Normalized Discounted Cumulative Gain~\citep[NDCG, ][]{jarvelin2002cumulated}. 
To establish gold passages, we retrieve the top-10 for each question-method pair and retain those judged relevant by all \mm{}s (\ref{prompt:relevant_passages_identification}).

\paragraph{Discrepancy detection.} We first evaluate the discrepancy detector on \texttt{FEVER-DPLACE-Q}, then apply it to \texttt{ROSIE}. Detected discrepancies are integrated with \ling{}’s outputs and classified by annotators.
\subsection{Results}
We present results from applying \ling{} on 100 \(C^{(a)}\) passages, where the primary topic is \(t_{12}\) ({\it Pregnancy}) or \(t_{16}\) ({\it Infant Care}) (see Table~\ref{tab:statistics} for sample statistics), following the dimensions from \S\hyperref[subsec:metrics]{\ref{subsec:metrics}}.
\subsubsection{Question and answer generation}\label{subsec:question_generation_quality}
\begin{figure}[!t]
    \centering
    \begin{subfigure}[b]{\columnwidth}
        \centering
        \includegraphics[width=\linewidth]{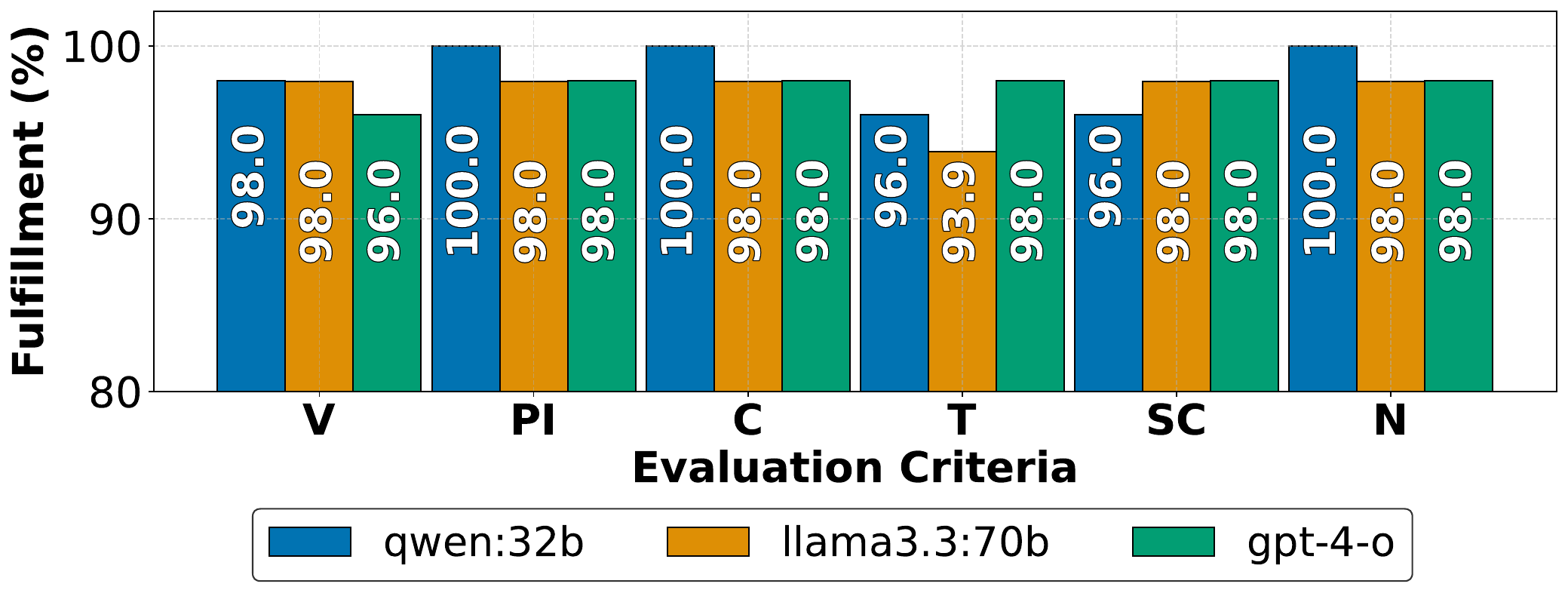}
        \caption{Question generation}
        \label{fig:question_quality_evaluation}
    \end{subfigure}
    \hfill
    \begin{subfigure}[b]{\columnwidth}
        \centering
        \includegraphics[width=\linewidth]{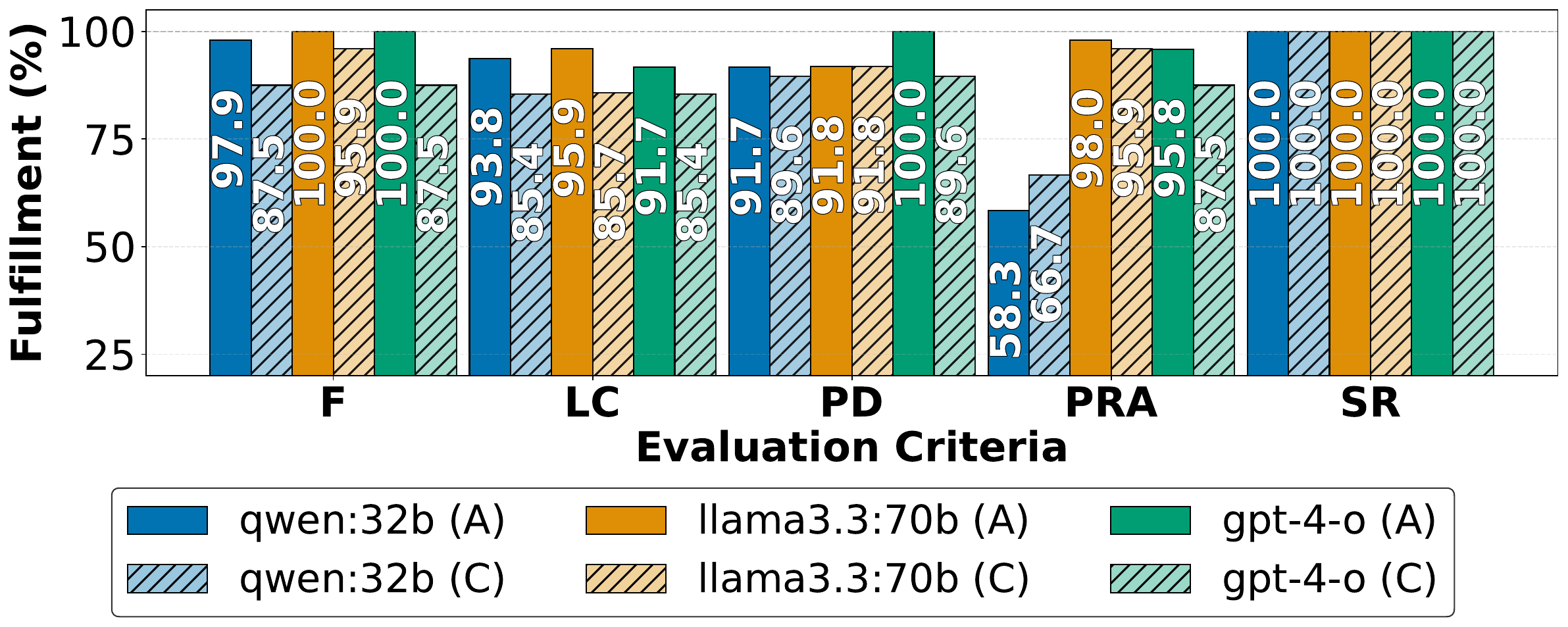}
        \caption{Answer generation}
        \label{fig:answer_quality_evaluation}
    \end{subfigure}
    \caption{Mean fulfillment rates for question (a) and answer (b) quality by model, based on majority vote from three annotators. Answer results are split by anchor vs. comparison corpora. Models generate questions reliably, but answer quality varies between $C^{(a)}$ and $C^{(c)}$.}
    \label{fig:question_answer_quality}
\end{figure}

As an initial step, \ling{} must generate well-formed questions from anchor passages and corresponding answers from both anchor and comparison corpora.
Fig.~\ref{fig:question_answer_quality} compares how well the evaluated \mm{}s fulfill the predefined criteria (\S\hyperref[subsec:metrics]{\ref{subsec:metrics}}) in generating questions (\subref{fig:question_quality_evaluation}) and answers (\subref{fig:answer_quality_evaluation}), based on 50 items per model. Each item was labeled by three annotators, with pass/fail determined by majority vote ($\geq 2/3$ positives). Percentages shown are the mean proportion of items passing. Inter-annotator agreement is calculated per criterion using Gwet's AC1 and macro-averaged per \mm{} (Table~\ref{tab:question_answer_agreement}).

Models perform strongly overall but vary in reliability and quality. In \underline{question generation}, fulfillment is uniformly high ($\geq\!94\%$ across criteria), yet agreement varies: annotators often disagree on {\bf S}elf-{\bf C}ontainment (\texttt{qwen:32b}, 0.62). 
{\bf C}larity (\texttt{llama3.3:70b}, 0.77) and {\bf N}aturalness (\texttt{gpt-4o}, 0.79) also show more variable judgments. 
For \underline{{\it anchor} answers}, fulfillment remains high across most criteria but drops for \underline{{\it comparison} answers}, with \texttt{gpt-4o} ranking highest or tied-highest on nearly all. \texttt{qwen:32b} underperforms with high agreement on {\bf P}assage {\bf R}eference {\bf A}voidance, while {\bf F}aithfulness and {\bf P}assage {\bf D}ependence agree less. %
%
The quality drop in \(C^{(c)}\) answers stems from the Spanish passages sometimes not fully aligning with the questions (see \S\hyperref[example:comp_answer_failures]{\ref{example:comp_answer_failures}}). 
\begin{table}[!t]
    \centering
    \resizebox{\columnwidth}{!}{%
    \begin{tabular}{lccccccc}
    \toprule
    & {\bf V} & {\bf PI} & {\bf C} & {\bf T} & {\bf SC} & {\bf N} & {\bf Macro} \\
    \midrule
    \texttt{qwen:32b}     & 0.88 & 0.90 & 0.88 & 0.86 & 0.62 & 0.83 & \cellcolor{gray!15}0.83 \\
    \texttt{llama3.3:70b} & 0.89 & 0.89 & 0.77 & 0.97 & 0.87 & 0.89 & \cellcolor{gray!15}0.88 \\
    \texttt{gpt-4-o}      & 0.94 & 0.88 & 0.79 & 0.94 & 0.88 & 0.79 & \cellcolor{gray!15}0.87 \\
    \midrule
    & {\bf F} & {\bf PD} & {\bf PRA} & {\bf SR} & {\bf LC} & \multicolumn{2}{c}{Macro} \\
    \midrule
    \texttt{qwen:32b}     & 0.82/0.50 & 0.66/0.47 & 0.92/0.92 & 0.99/0.99 & 0.87/0.84 & \multicolumn{2}{c}{\cellcolor{gray!15}0.85/0.74} \\
    \texttt{llama3.3:70b} & 0.99/0.69 & 0.84/0.56 & 0.82/0.81 & 1.00/1.00 & 0.96/0.85 & \multicolumn{2}{c}{\cellcolor{gray!15}0.92/0.78} \\
    \texttt{gpt-4-o}      & 0.97/0.69 & 0.96/0.67 & 1.00/0.93 & 1.00/1.00 & 0.92/0.88 & \multicolumn{2}{c}{\cellcolor{gray!15}0.97/0.83} \\
    \bottomrule
    \end{tabular}
    }
    \caption{Gwet's AC1 for question/ answer generation. For answers, numbers are anchor/comparison. 
    Agreement is high overall (\texttt{llama3.3:70b} leads on questions and \texttt{gpt-4o} on answers), confirming that questions and answers generally meet the criteria.
    }
    \label{tab:question_answer_agreement}
\end{table}

\begin{table*}[!t]
\centering
\resizebox{\textwidth}{!}{%
\begin{tabular}{c ccccccccc}
\arrayrulecolor{black}
\toprule
\textbf{Method} & \textbf{MRR@3} & \textbf{MRR@5} & \textbf{NDCG@3} & \textbf{NDCG@5} & \textbf{Precision@3} & \textbf{Precision@5} & \textbf{Recall@3} & \textbf{Recall@5} & \textbf{Time (s)} \\
\midrule
\textbf{ANN} & $0.640 \pm 0.034$ & $0.522 \pm 0.037$ & $0.151 \pm 0.042$ & $0.145 \pm 0.039$ & $0.138 \pm 0.039$ & $0.125 \pm 0.034$ & $0.058 \pm 0.020$ & $0.087 \pm 0.025$ & $\boldsymbol{0.015 \pm 0.000}$ \\
\textbf{ENN} & $0.724 \pm 0.035$ & $0.620 \pm 0.038$ & $0.410 \pm 0.051$ & $0.400 \pm 0.046$ & $0.351 \pm 0.046$ & $0.299 \pm 0.040$ & $0.223 \pm 0.037$ & $0.293 \pm 0.041$ & $0.128 \pm 0.001$ \\
\textbf{TB-ANN} & $0.691 \pm 0.036^{\dagger}$ & $0.585 \pm 0.039^{\dagger}$ & $0.342 \pm 0.050^{\dagger}$ & $0.337 \pm 0.048^{\dagger}$ & $0.301 \pm 0.045^{\dagger}$ & $0.269 \pm 0.040^{\dagger}$ & $0.167 \pm 0.032^{\dagger}$ & $0.232 \pm 0.039^{\dagger}$ & $0.017 \pm 0.000$ \\
\textbf{TB-ANN-D} & $0.691 \pm 0.036$ & $0.585 \pm 0.039$ & $0.342 \pm 0.050$ & $0.337 \pm 0.048$ & $0.301 \pm 0.045$ & $0.269 \pm 0.040$ & $0.167 \pm 0.032$ & $0.232 \pm 0.039$ & $0.017 \pm 0.000$ \\
\textbf{TB-ANN-W} & $0.656 \pm 0.034$ & $0.542 \pm 0.037$ & $0.199 \pm 0.046$ & $0.200 \pm 0.046$ & $0.172 \pm 0.041$ & $0.162 \pm 0.040$ & $0.091 \pm 0.027$ & $0.128 \pm 0.034$ & $0.018 \pm 0.000$ \\
\textbf{TB-ANN-W-D} & $0.656 \pm 0.034$ & $0.542 \pm 0.037$ & $0.199 \pm 0.046$ & $0.200 \pm 0.046$ & $0.172 \pm 0.041$ & $0.162 \pm 0.040$ & $0.091 \pm 0.027$ & $0.128 \pm 0.034$ & $0.019 \pm 0.000$ \\
\textbf{TB-ENN} & $\boldsymbol{0.725 \pm 0.035}$ & $0.620 \pm 0.039^{\dagger}$ & $0.413 \pm 0.051$ & $0.401 \pm 0.046^{\dagger}$ & $\boldsymbol{0.354 \pm 0.046}$ & $0.300 \pm 0.040$ & $\boldsymbol{0.225 \pm 0.037}$ & $0.291 \pm 0.040$ & $0.303 \pm 0.007$ \\
\textbf{TB-ENN-D} & $\boldsymbol{0.725 \pm 0.035}$ & $0.620 \pm 0.039$ & $0.413 \pm 0.051$ & $0.401 \pm 0.046$ & $\boldsymbol{0.354 \pm 0.046}$ & $0.300 \pm 0.040$ & $\boldsymbol{0.225 \pm 0.037}$ & $0.291 \pm 0.040$ & $0.307 \pm 0.007$ \\
\textbf{TB-ENN-W} & $0.724 \pm 0.034$ & $\boldsymbol{0.623 \pm 0.039^{\dagger\ddagger}}$ & $\boldsymbol{0.417 \pm 0.054}$ & $\boldsymbol{0.418 \pm 0.048^{\dagger\ddagger}}$ & $\boldsymbol{0.354 \pm 0.049}$ & $\boldsymbol{0.319 \pm 0.041^{\dagger}}$ & $0.220 \pm 0.036$ & $\boldsymbol{0.306 \pm 0.040^{\dagger}}$ & $0.313 \pm 0.008$ \\
\textbf{TB-ENN-W-D} & $0.724 \pm 0.034$ & $\boldsymbol{0.623 \pm 0.039^{\dagger\ddagger}}$ & $\boldsymbol{0.417 \pm 0.054}$ & $\boldsymbol{0.418 \pm 0.048^{\dagger\ddagger}}$ & $\boldsymbol{0.354 \pm 0.049}$ & $\boldsymbol{0.319 \pm 0.041^{\dagger}}$ & $0.220 \pm 0.036$ & $\boldsymbol{0.306 \pm 0.040^{\dagger}}$ & $0.325 \pm 0.009$ \\
\bottomrule
\end{tabular}}
\caption{Performance metrics for \texttt{gpt-4o}. Values are means over queries with 95\% bootstrap CIs for retrieval at $L \in \{3,5\}$. Results use relevant $c_p^{(c)}$ passages from 100 $t_{16}$ $c_p^{(a)}$ passages. Best values are bolded; $\dagger$ marks topic-based methods significantly outperforming baselines, and $\ddagger$ marks weighted topic-based methods significantly outperforming unweighted ones. TB-ENN-W achieves the highest retrieval performance overall.}
\label{tab:retrieval_performancet_15_qwen_gpt}
\end{table*}
\begin{figure*}[!t]
    \centering
    \includegraphics[width=0.7\textwidth]{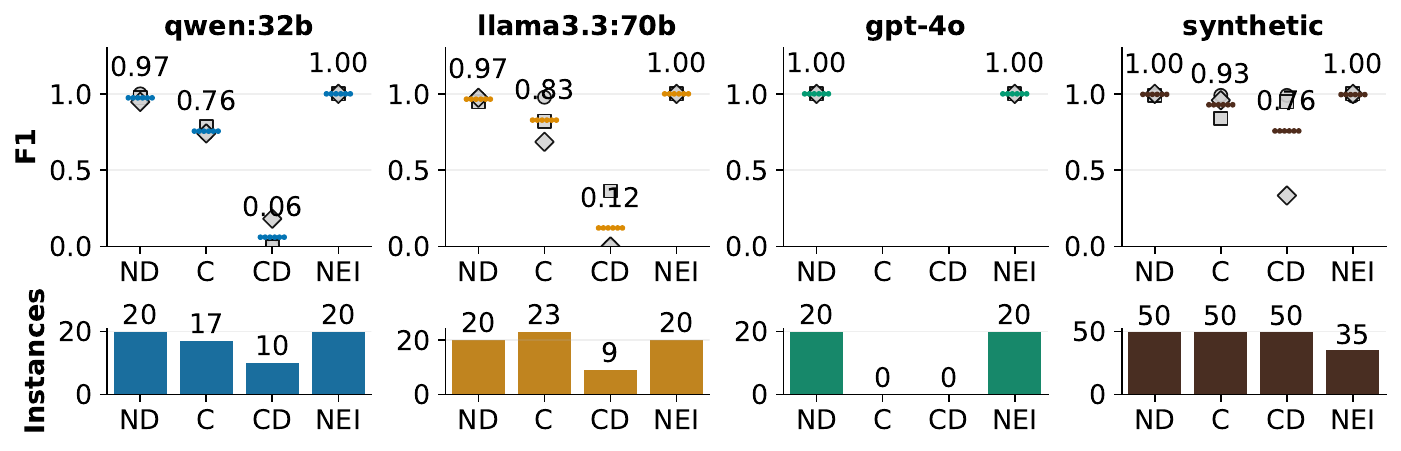}
    \caption{F1 scores per annotator and number of instances per category per model (for the controlled dataset, labeled synthetic, counts are actual instances). Dashed lines mark the mean across annotators. \texttt{llama3.3:70b} and \texttt{qwen:32b} cover all categories but yield more false positives, while \texttt{gpt-4o} predicts none.
    }
    \label{fig:discrepancy_evaluation}
\end{figure*}

\subsubsection{Retrieval evaluation}\label{subsec:retrieval_performance_results}
The retrieval step identifies candidate comparison passages used to generate questions that are later contrasted with the anchor answers. 
Results are reported at the query level, treating each query equally regardless of the relevant passages count (Table~\ref{tab:statistics}); means with 95\% bootstrap confidence intervals are estimated via query resampling. 
Statistical significance is assessed using the Friedman test across methods, followed by paired one-sided Wilcoxon signed-rank tests with Holm correction,
testing whether:
(i) topic-based methods 
improve ANN/ENN baselines, and
(ii) weighted topic-based methods  
improve unweighted counterparts.

Table~\ref{tab:retrieval_performancet_15_qwen_gpt} summarizes the results for \texttt{gpt-4o}\footnote{\S\hyperref[appen:more_retrieval_performance]{\ref{appen:more_retrieval_performance}} includes results for \(t_{16}\) using \texttt{qwen:32b} and \texttt{llama3.3:70b}, and the same analysis for \(t_{12}\).}. 
\underline{TB-ENN-W} consistently 
has best retrieval, 
particularly in Recall@5 and MMR@5, followed closely by TB-ENN. 
TB-ANN is slightly better than ANN across all metrics. 
As expected, all ANN-based methods are worse in ranking metrics but have the lowest retrieval times. 
Dynamically setting $\epsilon$ yields only marginal time gains (e.g., TB-ENN-W vs. TB-ENN-W-D for \texttt{qwen:32b} and \texttt{llama3.3:70b}; Table~\ref{tab:retrieval_performance_t15_llama}) with no clear effect for \texttt{gpt-4o}.
Overall, topic-based methods identify relevant passages reliably, with \texttt{gpt-4o} leading, \texttt{llama3.3:70b} following closely, and \texttt{qwen:32b} lagging with greater variability, though results vary across topics.

\subsubsection{Discrepancy detection quality}\label{subsec:discrepancy_detection_quality}
Having seemingly well-defined questions and answers in both languages, this step examines whether they can reveal discrepancies in the underlying passages that produced them.
\begin{table}[!t]
\centering
\resizebox{0.95\columnwidth}{!}{%
\begin{tabular}{lccc}
\toprule
\textbf{Label} & \texttt{qwen:32b} &\texttt{llama3.3:70b} & \texttt{gpt-4o} \\
\midrule
\abr{contradiction} & 0.935 & 0.883 & {\bf 0.962} \\
\abr{cultural\_discrepancy} & {\bf 0.889} & 0.809 & 0.869 \\
\abr{not\_enough\_info} &  0.825 & 0.889 & {\bf 0.889} \\
\abr{no\_discrepancy}  & {\bf 0.980} & 0.961 & 0.942 \\
\midrule
{\bf Weighted F1} &  0.914 & 0.885 & {\bf 0.918} \\
\bottomrule
\end{tabular}}
\caption{\mm{}s' F1 scores in \texttt{FEVER-DPLACE-Q}. \texttt{gpt-4o} outperforms in average, followed closely by \texttt{qwen:32b}.}
\label{tab:synthetic}
\end{table}

Crowdworkers classify \texttt{FEVER-DPLACE-Q} and \ling{}-detected discrepancies in 100-sample subsets from $t_{12}$ and $t_{16}$, with 20 \nd{}/\nei{} samples per \mm{} and topic. Inter-annotator agreement (Fleiss’s $\kappa$) is $0.743$.  
Table~\ref{tab:synthetic} presents the F1 scores from applying the discrepancy module on \texttt{FEVER-PLACE-Q} (see also Fig.~\ref{fig:confusion_fever_d_place_q}). Fig.~\ref{fig:discrepancy_evaluation} summarizes model–human agreement (category matches between the \mm{} and annotators) and human agreement with \texttt{FEVER-DPLACE-Q} gold labels, reported as F1-score per annotator.

\texttt{qwen:32b} is best on the controlled dataset in \cd{} and \nd{}, while \texttt{gpt-4o} leads in \con{} and \nei{}, with \texttt{llama3.3:70b} trailing \cd{}. Overall performance is strong but divided: \texttt{qwen:32b} is slightly better at handling discrepancies, and \texttt{gpt-4o} at factuality.
Annotator alignment with \nd{} and \nei{} is largely stable across \ling{} instances from all models and the controlled dataset. However, \texttt{gpt-4o} fails to detect any discrepancy type; annotators disagree on the \cd{}s flagged by \texttt{qwen:32b} and \texttt{llama3.3:70b}, with the latter performing slightly better; and agreement on \con{}s is higher for \texttt{llama3.3:70b}. In the synthetic dataset, one annotator severely misses \cd{}s but labels as \cd{}s some of the discrepancies flagged by \texttt{qwen:32b}, which the other two annotators do not.

Manual inspection of instances labeled as \cd{} by \ling{} shows that annotators often classify them as \con{}/\nd{}s instead. For example, to the question \textit{``Are all newborns required to undergo cardiac screening tests?''}, the anchor/comparison answers are: \textit{``Yes, all newborns are required [...] as part of standard newborn screenings''} and \textit{``No, not all newborns are required [...] \underline{California only requires healthcare professionals} \underline{to offer them}''}. Annotators labeled this case as \con{}. While the answers appear contradictory, the discrepancy arises from a regulatory exception in California rather than a fundamental contradiction in medical practice: {\bf both can be correct depending on jurisdiction, reflecting variation in U.S. state health regulations.}

A similar case occurs with \textit{``Is it necessary for a child with impetigo to wait more than 48 hours after starting antibiotics before returning to daycare or school if improving?''}, with answers: \textit{``No, it is not necessary [...], a child can return to daycare or school \underline{48 hours after starting} antibiotic treatment as long as there are signs of improvement.''} and \textit{``No, it's not necessary [...] They can usually return to school \underline{24 hours after beginning} treatment.''}. Here, annotators labeled the pair as \nd{}. While both responses are logically consistent, each rejecting the need to wait more than 48 hours, {\bf one specifies 24 hours and the other 48 hours, reflecting variation in institutional policies.}\footnote{In making these examples, the authors checked that the responses were faithful to the passages to ensure a true discrepancy across corpora (\S\hyperref[example:cd_failures]{\ref{example:cd_failures}}).}

These examples illustrate both the difficulty of consistently distinguishing cultural discrepancies and the subjectivity of annotator judgments. \ling{} effectively surfaces potential discrepancy cases, but fine-grained category boundaries remain unstable across models and annotators. This highlights the need for human supervision rather than blind reliance, and for annotators tailored to specific use cases and in-the-loop with the knowledge base under evaluation.
In some cases, introducing an additional category (e.g., contextual differences) may also help to capture cases that are not strictly cultural yet do not fit cleanly as contradictions.

 \label{sec:ablation}

\section{\textsc{MIND} the Discrepancies} \label{sec:sections/60-mind_discrepancies} 
%
Based on the ablation, we select \texttt{llama3.3:70b} as the final \mm{} {\it helper} (\texttt{qwen:32b} detects all discrepancy types and performs slightly better on the controlled dataset, but its questions/ answers are weaker and annotator agreement lower; the small gain in the controlled set does not outweigh these drawbacks) and apply \ling{} to a 500-sample of anchor passages, with majoritarian topics \(t_{12}\), \(t_{16}\), and \(t_{25}\) ({\it Pediatric Healthcare}).

Two authors (public health graduate students) revised the {\color{red}} discrepancies detected, plus an equal number of \nd{} cases, and refined them to create \texttt{ROSIE-MIND-v2}.\footnote{In a previous iteration, we generated \texttt{ROSIE-MIND-v1} using \texttt{quora-distilbert-multilingual}, and \texttt{qwen:32b} (\S\hyperref[appen:rosie_mind]{\ref{appen:rosie_mind}}).} 
\ling{} detected \num{71}, \num{136}, and \num{50} \con{}s; \num{35}, \num{70}, and \num{33} \cd{}s; \num{5990}, \num{4313}, and \num{3937} \nd{}s; and about \num{35}K \nei{}s in the analyzed passages for \(t_{12}\), \(t_{16}\), and \(t_{25}\), respectively. 
%
In what follows, we reference examples from \texttt{ROSIE-MIND} through hyperlinks to Table~\ref{tab:rosie_mind_examples} using the format \texttt{T-\{topic\_id\}-\{passage\_id\}} to illustrate failure cases and discrepancy patterns.

One such issue is that, in some cases, false discrepancies arise from the retrieved passage not fully aligning with the question, and \ling{}'s contextualization being insufficient for the \mm{} to classify it as irrelevant (\hyperlink{T_11_12176}{T-11-12176}).
In others, they occur when neither passage has enough information to answer the question (\hyperlink{T_11_7831}{T-11-7831}), or when the target chunk misses a specific detail, e.g., the question asks for a statistic from a particular institute, but the passage provides only a valid statistic from another source (\hyperlink{T_11_457}{T-11-457}). A further source of error involves questions generated from anecdotal or non-generalizable content---which occur frequently in \texttt{ROSIE}, such as {\it ``Will you receive the test results within 2 months?''} based on the passage{\it ``You can expect to get test results within 10 weeks.''}, which is unlikely to retrieve a reliable counterpart.
Yet, some detected contradictions are strong and should be addressed. For instance, one passage states that moderate alcohol consumption while breastfeeding is harmless, while another claims no level of alcohol in breast milk is safe (\hyperlink{T_15_470}{T-15-470}). Another pair presents conflicting advice on whether women should take a full zinc dose during pregnancy (\hyperlink{T_15_19913}{T-15-19913}).
There are also mild discrepancies: \hyperlink{T_11_455}{T-11-455} is labeled as \con{}, since the passages report different prevalence numbers for children with ASDs in the US. While contradictory, the difference may stem from statistics released in different years; without the year information, they remain a contradiction (\S\hyperref[appen:discrepancy_framing]{\ref{appen:discrepancy_framing}}).

{\bf Discrepancies by topic.} Contradictions are more frequent in domains with stronger medical guidelines (e.g., pregnancy), where opposing recommendations are common (\hyperlink{T_15_19913}{T-15-19913}, \hyperlink{T_15_470}{T-15-470}), while cultural discrepancies prevail more in child development, likely reflecting differing parenting practices (\hyperlink{T_24_15460}{T-24-15460}, \hyperlink{T_24_849}{T-24-849}). The high number of NEIs suggests many English passages lack a direct counterpart in Spanish.

{\bf Generalizability.} \ling{} applies to any language with loosely aligned or MT-corpora, provided multilingual embeddings exist.
To test this, we apply \ling{} to 150 samples from two topics of a 25-topic \pltm{} trained on \texttt{WIKI-EN-DE}, a collection of 600 Wikipedia German-English pairs, identifying illustrative discrepancies. 
For example, English sources describe Anglo-American Freemasonry as requiring belief in a supreme being, whereas German sources report that Liberal Freemasonry does not, reflecting a decision of the Grand Orient de France in 1877 (\hyperlink{E_T_3_6276}{E-T-3-6276}).\footnote{Hyperlinks refer to examples shown in Table~\ref{tab:ende_mind_examples}.} Misaligned pages can create direct contradictions: a user asking whether ``The Preservation of St Paul after a Shipwreck at Malta'' was painted by Benjamin West would see the English page affirming it, while the German page attributes a different work, ``St Paul auf der Melite'' (\hyperlink{E_T_3_854}{E-T-3-854}). Such cases show how cross-lingual inconsistencies can mislead users despite both sources being \emph{authoritative} (see also \S\hyperref[appen:wiki_ende]{\ref{appen:wiki_ende}}).

{\bf What Needs to Change?} Some discrepancies flagged by \ling{} are false positives---not due to passages truly conflicting, but because generated answers omit details or blur nuances. This reveals a deeper limitation of \mm{}-based QA: incomplete answers can make any system built on the same approach appear self-contradictory.  
To ensure consistency, QA systems like \texttt{ROSIE} must preserve context at retrieval to avoid false positives and guarantee access to relevant factual documents. A data-driven tool like \ling{} can help surface discrepancies, but fully supervising large knowledge bases is infeasible. An incremental strategy---documents in one language fill gaps in the other, similar to KnowledgeBase Guardian\footnote{\label{KnowledgeBase}\url{https://github.com/datarootsio/knowledgebase\_guardian}} but by topic---offers a path toward cross-lingual consistency, though human oversight will remain necessary.
 \label{sec:mind_discrpancies}

\section{Related Work} \label{sec:sections/90-related_work} \ling{} lies at the intersection of multilingual QA, fact-checking, and contradiction detection, unified through their reliance on NLI. QA can be framed as an entailment and vice versa~\citep{chen2021can,harabagiu2006methods,trivedi2019repurposing,kamath2020selective,mishra2021looking}, with contradiction signals shown to improve QA~\citep{fortier2023using}. 
\mm{}s are widely used for fact-checking~\citep{li2023self,rawte2024factoid,chen2024complex}, but lack blind reliability.
Benchmarks like FoolMeTwice~\citep[FM2,][]{eisenschlos-etal-2021-fool} stress-test retrieval and verification with human-written claims from Wikipedia. \citet{guan2024language} find model outputs can aid verification despite hallucinations, but \citet{si2024large} (on FM2) find \mm{}s sometimes produce convincing yet incorrect answers that humans over-trust, motivating human-in-the-loop methods~\citep{chen2024complex}.

Outside QA, contradiction detection alone is a long-studied problem~\citep{condoravdi2003entailment,harabagiu2006negation,schuster2022stretching, hsu2021wikicontradiction}. KnowledgeBase Guardian\footnoteref{KnowledgeBase} exemplifies an \mm{}-based approach to building contradiction-free knowledge bases.

Topic modeling has enhanced retrieval by combining \lda{}~\citep{blei2003latent} with word-level statistics~\citep{wei2006lda}, supporting nearest neighbor indexing, clustering~\citep{scherer2013topic}, and similarity search in reduced topic spaces~\citep{badenes2017efficient}. Interactive topic modeling~\citep{hu2014interactive} lets users refine topics, also in multilingual settings~\citep{yuan2018multilingual}. Yet systematic use for cross-lingual thematic alignment and retrieval within topic clusters is underexplored, though prior work has applied topic models to detect cultural differences across linguistic communities~\citep{gutierrez2016detecting}.

\label{sec:related_work}

\section{Conclusion} \label{sec:sections/100-conclusion} 
While not fully hands-off, \ling{} streamlines discrepancy detection in multilingual databases, achieving high agreement on non-discrepancy cases and reducing human supervision mainly to discrepancy cases---as reflected in our final output, \texttt{ROSIE-MIND}.
To further reduce human effort, we plan to incorporate active learning, prompting review only when model uncertainty is high~\citep{li-etal-2025-large-language}. To support this, we will fine-tune \ling{} 
using \texttt{ROSIE-MIND}. 
Looking ahead, we see potential for \ling{} to help surface cultural differences correlated with language. By identifying these through contradiction patterns, this knowledge could help mitigate bias by adapting models to better serve the cultural needs of underrepresented groups.
 \label{sec:conclusions}

\section{Limitations} \label{sec:sections/110-limitations} {\color{black}
Since \ling{} is designed for QA practitioners seeking to identify discrepancies in their knowledge bases, users are responsible for performing translation when parallel corpora are unavailable. However, translation quality is not critical---as long as documents are aligned thematically, even basic models like the one used in this paper are sufficient.
A second limitation concerns segmentation: as shown in the \texttt{ROSIE} results, poor segmentation can affect results, making careful preprocessing essential.
}
Lastly, our approach does not assume a direct correspondence between the anchor and comparison corpora, nor do we have prior knowledge of their content. Consequently, there are no ground-truth relevant passages for evaluating retrieval performance. However, since all retrieval methods are assessed using the same set of theoretically relevant passages---collected across all methods---the evaluation remains consistent across approaches.



 \label{sec:limitations}

\section{Acknowledgments} \label{sec:sections/130-acknowledgment} Many thanks are due to Neha Srikanth for discussion and insight during the initial planning stages of this work; to Heran Mane for her ready availability in answering questions about Rosie; and to Audrey Zarzuela for her thoughtful assistance in the early stages of annotation framing.
This work has been supported by Grant PID2023-146684NB-I00, funded by MICIU/AEI/10.13039/501100011033 and by ERDF/UE (Calvo-Bartolomé and Arenas-García) and the NIH Award No. R01MD016037 (Boyd-Graber).
Any opinions, findings, conclusions, or recommendations expressed in this material are those of the author and do not necessarily reflect the views of the National Institutes of Health.

\bibliography{bib/custom}

\clearpage \newpage
\section{Appendix} \label{sec:sections/200-appendix} \appendix

\section{Filtering of \texttt{ROSIE}}\label{appen:filtering}
As an initial filtering step, we removed \(33\,510\) and \(1\,886\) passages misclassified by language in the original authors' corpus, most of which contained bibliographic information and references. Since after this filtering many passages still remained irrelevant to the QA system's domain, we developed a score to identify additional passages for removal. 
The score relies on the word-topic distribution of a monolingual topic model, meaning that we train an \lda{} model per language (one on \(C^{(a)}\) and one on \(C^{(c)}\)) using \mallet{}. Let \(\boldsymbol{\beta}_{k}, \; \beta_{k,v}, v=1,..., V^{(a)'}\) where \(V^{(a)'}\) is the vocabulary in the language\footnote{Unlike \(V^{(a)}\), \(V^{(a)'}\) excludes the translated words from the other anchor corpus that are present in \(V^{(a)}\).}, be the topic distribution of a given topic \(t_k\). Note that here we have a separate word-topic matrix per language, as we train one \lda{} model per language.

To quantify the thematic relevance of a passage within the corpus, we use Eq.~\eqref{eq:beta_ds}~\cite{blei2009topic}, which re-ranks topic words by penalizing those that are commonly distributed across topics, favoring topic-specific terms. This re-ranking guides the selection of \(\mathcal{W}_p \subseteq V\)---the words retained after preprocessing \(c_p^{(a)}\)---while \(\mathcal{W}^*_p = V^{(a)'} \setminus \mathcal{W}_p\) denotes the excluded words.

Building on this, we define the passage score \(\xi_p\) (Eq.~\eqref{eq:doc_score}) as the normalized sum of the maximum word-topic weights for all words in \(\mathcal{W}_p\), with a penalty reflecting the proportion of excluded words \(\mathcal{W}^*_p\). The same process is applied independently to the model trained on the comparison corpus \(C^{(c)}\).

\begin{equation}
    \label{eq:beta_ds}
    \beta^{DS}_{k,v} = \beta_{k,v} \log\frac{\beta_{k,v}}{\left(\prod_{j=1}^K \beta_{j,v}\right)^{1/K}},
\end{equation}

\begin{equation}
    \label{eq:doc_score}
    \xi_p = \frac{\frac{1}{\lvert \mathcal{W}^*_p \rvert}\sum_{w_v \in \mathcal{W}_p} \max_{k \in \{1, \dots, K\} } \beta^{DS}_{k,v}}{\lvert \mathcal{W}_p \rvert}.
\end{equation}

Based on this score, we selected 1000 {\it bad} passage candidates per language (bottom 1\% percentile) and 1000 {\it good} candidates randomly sampled from the remainder. Of these, 200 were manually annotated by a Public Health Science graduate student (one of the authors), who marked passages as relevant (1) if they could answer user questions (e.g., from a \texttt{ROSIE} mum). The method achieved 87\% accuracy for English and 80\% for Spanish in identifying {\it bad} passages, but only 42\% and 52\% for {\it good} ones. This suggests the score is effective at filtering irrelevant content (e.g., author listings, bibliographies) but struggles with personal experience narratives. Table~\ref{tab:filtering_examples_english} shows English examples.
Using these annotations, we trained a SVM classifier per corpus using the document-topic and TF-IDF representation of the passages as features, and generated predictions for all passages, retaining those classified as relevant, obtaining the final \(542,055\) English and \(333,175\) Spanish passages.

\begin{table*}[!t]
\renewcommand{\arraystretch}{1.2}
\footnotesize
\centering
\resizebox{\textwidth}{!}{%
\begin{tabular}{p{12cm} p{2cm} p{1cm} p{1cm}}
\toprule
\textbf{Passage} & \(\xi_p\) & \textbf{Pred. label} & \textbf{Final label}\\ 
\midrule
MODERATOR: Thank you. We'll move on to the line of Jennifer Warner with WebMD. Please go ahead. & 0.0017 & 0 & 0\\
Di Maria MV, Goldberg DJ, Paridon S, Lubert A, Dragulescu A, Mackie AS, McCrary A, Weingarten A, Parthiban A, Goot B, Goldstein BH, Taylor C, Lindblade C, Petit C, Spurney C, Harrild DM, Urbina EM, Schuchardt E, Trachtenberg F, Kim GB, Yoon JK, Colombo JN, Wang K, Files MD, Schoessling M, Ermis PR, Wong P, Garg R, Swanson S, Menon SC, Srivastava S, Thorsson T, Johnson T, Krishnan U, Frommelt PC: Impact of Udenafil on Echocardiographic Indices of Single Ventricle Size and Function in FUEL Study Participants. Circulation 2020. & 0.0009 &  0 & 0\\
Reviewed on Feb 9, 2023: Dr. Novak seems very determined to help me, and I appreciate that.  & 0.0021 &  0 & 0\\
Join us as we recap the most popular posts of 2022, plus three posts you may have missed.  & 0.0031 &  0 & 0\\

Temperament includes behavioral traits such as sociability (outgoing or shy), emotionality (easy-going or quick to react), activity level (high or low energy), attention level (focused or easily distracted), and persistence (determined or easily discouraged). These examples represent a spectrum of common characteristics, each of which may be advantageous in certain circumstances. Temperament remains fairly consistent, particularly throughout adulthood.  & 0.0023 & 0 & 1\\
What are grief and grieving: Grief is a natural response to the loss of someone or something very important to you. The loss may cause sadness and may cause you to think of very little else besides the loss. The words sorrow and heartache are often used to describe feelings of grief. & 0.0016 & 0 & 1\\
Your peripheral nervous system has two main subsystems: autonomic and somatic. & 0.0021 & 0 & 1\\
Seek help if you have emotional ups and downs or feel depressed.  & 0.0022 & 0 & 1\\

\midrule
21-hydroxylase deficiency is one of a group of disorders known as congenital adrenal hyperplasias that impair hormone production and disrupt sexual development. 21-hydroxylase deficiency is responsible for about 95 percent of all cases of congenital adrenal hyperplasia. & 0.0262 & 1 & 1\\
How common is hypoplastic left heart syndrome: About 1 in 3,800 babies are born with hypoplastic left heart syndrome each year in the U.S. This condition accounts for about 2\% to 3\% of all congenital heart diseases (CHD). Hypoplastic left heart syndrome affects more men and people assigned male at birth (67\%) than women and people assigned female at birth. & 0.0562 & 1 & 1\\
Do infants get enough iron from breast milk: Most newborns have sufficient iron stored in their bodies for about the first 6 months of life depending on gestational age, maternal iron status, and timing of umbilical cord clamping. By age 6 months, however, infants require an external source of iron apart from breast milk. Breast milk contains little iron; therefore, parents of infants receiving only breast milk should talk to their infant’s health care provider about whether their infant needs iron supplements before 6 months of age. & 0.0223 & 1 & 1\\
Shoulder dislocations: A sudden impact to the shoulder can cause the top of the upper arm bone to dislocate from the socket of the shoulder blade. This is most common in young athletes who play contact sports. A dislocated shoulder is prone to repeated dislocation, which can cause damage to nerves, blood vessels, tendons or ligaments, and may require surgery to prevent further instability and restore range of motion. & 0.0256 & 1 & 1\\
Melissa: Dr. Doolin really pushed my mom and dad for me to have this procedure even when they weren't sure. He told her, you have to give her a chance in life. And it has really allowed me to live my life as normal as possible. I don't think of myself as abnormal in that area because I was able to have the pull-through procedure. & 0.007 & 1 & 0\\
Fetal Treatment Program: Resources: Below are links to other sites about a variety of fetal conditions. They range from support groups to professional societies, fetal treatment centers, sites about ongoing clinical trials and general information sites. & 0.0436 & 1 & 0\\
Dr. Jennifer Maniscalco's office is located at 601 - 5th St S St Petersburg, FL 33701. & 0.0058 & 1 & 0\\
Openshaw JJ, Swerdlow DL, Krebs JW, et al. Rocky Mountain spotted fever in the United States, 2000--2007: interpreting contemporary increases in incidence. Am J Trop Med Hyg 2010;83:174--82. & 0.0101 & 1 & 0\\
\bottomrule
\end{tabular}%
}
\caption{Examples of passages filtered from the English corpus. The top rows show documents initially classified as {\it bad} (0) based on \(\xi_p\), with manual labels confirming them as either {\it bad} (0) or {\it good} (1). The bottom rows show documents initially classified as {\it good} (1), with manual labels indicating whether they were indeed {\it good} or {\it bad}.}
\label{tab:filtering_examples_english}
\end{table*}

\section{Pre-processing details}\label{appen:preprocessing_details}
The preprocessing steps included: (1) Filtering of texts that do not belong to the language on which the topic model is to be constructed. (2) Expanding contractions and basic acronyms in the specified language. (3) Tokenization, removal of non-alphanumeric characters, and conversion to lowercase. (4) Elimination of basic and domain-specific stopwords. (5) Lemmatization according to part-of-speech (POS) tagging. 
We rely on the \texttt{NLPipe}\footnote{\url{https://github.com/lcalvobartolome/NLPipe}} library for this.

\section{Optimization of K}\label{appen:k_optimization}
We experimented with $K \in [5, 50]$ and selected $K = 30$ as the final number of topics, as it achieved the highest NPMI coherence score while maintaining reasonable overlap, and provided finer characterization of maternal and infant health by separating areas such as {\it Pregnancy}, {\it Infant Care}, {\it Pediatric Healthcare}, and {\it Childhood Vaccination}, which were less distinct in the 15- and 20-topic models (see Tables~\ref{tab:15_tpcs}, \ref{tab:20_tpcs}, and \ref{tab:30_tpcs} for full topic lists and labels).

\section{Criteria for question and answer generation quality}\label{appen:criteria_question_answer_quality}
Tables~\ref{tab:question_cirteria} and~\ref{tab:answer_cirteria} contain the criteria for evaluating question and answer quality, respectively.
\begin{table*}[!h]
\renewcommand{\arraystretch}{1.4}
\footnotesize
\centering
\resizebox{\textwidth}{!}{%
\begin{tabular}{p{3cm} p{14cm}}
\toprule
\textbf{Criteria} & \textbf{Definition} \\ 
\midrule
\textbf{Verifiability} & The question is a yes/no question about verifiable information from the passage (i.e., it does not ask for subjective opinions, personal experiences, or details about the author's background). \\
\textbf{Passage Independence} & The question does not explicitly reference the passage (e.g., avoiding phrases like ``According to the passage...''). \\
\textbf{Clarity} & The question avoids ambiguous language, such as pronouns (``it'', ``they'') or vague references (``the'') unless the entity has been previously introduced. \\
\textbf{Terminology} & The question avoids technical terms unless adequately contextualized (e.g., writes ``Multisystem Inflammatory Syndrome in Children (MIS-C)'' instead of just ``MIS-C''). \\
\textbf{Self-Containment} & The question can be answered based solely on the information in the given passage. \\
\textbf{Naturalness} & The question is phrased in a way that a general user would naturally ask, avoiding unnecessary technicality or excessive detail. \\
\bottomrule
\end{tabular}%
}
\caption{Criteria for Evaluating Questions}
\label{tab:question_cirteria}
\end{table*}

\begin{table*}[!h]
\renewcommand{\arraystretch}{1.4}
\footnotesize
\centering
\resizebox{\textwidth}{!}{%
\begin{tabular}{p{3cm} p{14cm}}
\toprule
\textbf{Criteria} & \textbf{Definition} \\ 
\midrule
\textbf{Faithfulness} & The answer accurately reflects the information provided in the passage. If the passage lacks sufficient information to provide a valid answer, or just contain personal experiences, the response is ``I cannot answer given the context''.  \\
\textbf{Passage Dependence} & The answer is solely based on the passage and does not incorporate external knowledge or speculation. If the passage lacks sufficient information to provide a valid answer, or just contain personal experiences, the response is ``I cannot answer given the context''.\\
\textbf{Passage Reference Avoidance} & The answer does not explicitly refer to the passage itself (e.g., avoiding phrases like ``The passage provides general...''). \\
\textbf{Structured Response} & The answer begins with \abr{yes/no}, followed by a concise explanation based on the passage. If the passage lacks sufficient information to provide a valid answer, or just contain personal experiences, the response is ``I cannot answer given the context''. \\
\textbf{Language Consistency} & The answer is fully in the same language as the question and does not contain unexpected characters. \\
\bottomrule
\end{tabular}%
}
\caption{Criteria for Evaluating Answers}
\label{tab:answer_cirteria}
\end{table*}

\section{Use of Topic Distributions in Retrieval}\label{appen:topic_distr_retrieval}
In our retrieval pipeline, we leverage full topic distributions rather than relying solely on the top-ranked topic for a document. This design ensures that even if the primary topic assignment is suboptimal, documents can still be retrieved under other semantically relevant topics where they carry significant weight.
For instance, consider the following passage and the topic model trained with \(K=30\):

\begin{quote}
``For a double uterus, some basic questions to ask your doctor include:
\begin{itemize}
\item What's likely causing my symptoms?
\item Could there be other possible causes for my symptoms?
\item Do I need any tests done?
\item Do I need treatment?
\item Are there any alternatives to the treatment you're suggesting?
\item Are there restrictions I need to follow?
\item Should I see a specialist?
\item Do you have any brochures or other printed material I can take with me? What websites do you recommend?''
\end{itemize}
\end{quote}

This passage is primarily assigned to Topic 0, \textit{``Healthcare Guidance''} (keywords: \textit{provider, care, health, healthcare, doctor, medical, treatment, symptom, visit, recommend, question, diagnose}), based on its general medical phrasing. However, its second most probable topic is Topic 10, \textit{``Reproductive Health''} (keywords: \textit{uterus, woman, vaginal, menstrual, bleeding, hormone, sex, body}), which is thematically more appropriate.

By incorporating the full topic distribution in the retrieval phase, the document is still considered under Topic 10, ensuring more accurate and context-aware retrieval even when the top topic alone might not be the best match.

\section{Additional results}\label{appen:more_retrieval_performance}
\begin{table*}[!ht]
\centering
\renewcommand{\arraystretch}{1.2}
\resizebox{0.65\textwidth}{!}{%
\begin{tabular}{lccccc}
\toprule
{\bf Model} & {\bf \# Questions} & {\bf Question Length} & {\bf \# Queries} & {\bf Queries Length} & {\bf \# Rel. Passages}   \\
\midrule
\multicolumn{6}{c}{\(\mathbf{t_{12}}\)\textbf{ : Pregnancy}} \\
\midrule
\texttt{qwen:32b}    & $237$ & $16.03 \pm 4.74$ & $2.43 \pm 0.83$ & $6.59 \pm 1.77$ & $5.699\pm 6.304$\\
\texttt{llama3.3:70b} & $396$ & $16.51 \pm 5.53$ & $2.30 \pm 0.64$ & $7.98 \pm 3.44$ & $7.019\pm 7.289$\\
\texttt{gpt-4o}       & $297$ & $15.93 \pm 4.98$ & $2.02 \pm 0.13$ & $7.51 \pm 1.59$ & $6.032\pm5.627$\\
\midrule
\multicolumn{6}{c}{\(\mathbf{t_{16}}\)\textbf{ : Infant Care}} \\
\midrule
\texttt{qwen:32b}     & $238$ & $14.69 \pm 4.88$ & $2.40 \pm 0.55$ & $5.96 \pm 1.51$ & $5.953 \pm 5.000$  \\
\texttt{llama3.3:70b} & $384$ & $14.92 \pm 5.17$ & $2.31 \pm 0.51$ & $7.35 \pm 2.54$ & $6.972\pm6.514$  \\
\texttt{gpt-4o}       & $268$ & $15.04 \pm 4.65$ & $2.02 \pm 0.14$ & $6.93 \pm 1.69$  & $6.029\pm4.947$  \\
\bottomrule
\end{tabular}}
\caption{Summary statistics, including the total number of generated questions, question length, total number of queries, query length, and the number of relevant retrieved passages. Statistics are based on a subsample of 100 passages per \mm{} and topic.}
\label{tab:statistics}
\end{table*}
\begin{table*}[h]
\centering
\resizebox{\textwidth}{!}{%
\begin{tabular}{c ccccccccc}
\arrayrulecolor{black}
\toprule
\textbf{Method} & \textbf{MRR@3} & \textbf{MRR@5} & \textbf{NDCG@3} & \textbf{NDCG@5} & \textbf{Precision@3} & \textbf{Precision@5} & \textbf{Recall@3} & \textbf{Recall@5} & \textbf{Time (s)} \\
\midrule
\multicolumn{10}{l}{\texttt{llama3.3:70b}} \\
\arrayrulecolor{black}\specialrule{0.5pt}{0pt}{0pt}\arrayrulecolor{black}
\textbf{ANN} & $0.644 \pm 0.032$ & $0.536 \pm 0.034$ & $0.117 \pm 0.032$ & $0.119 \pm 0.030$ & $0.112 \pm 0.031$ & $0.104 \pm 0.027$ & $0.051 \pm 0.019$ & $0.077 \pm 0.023$ & $\boldsymbol{0.015 \pm 0.000}$ \\
\textbf{ENN} & $0.707 \pm 0.031$ & $0.614 \pm 0.033$ & $0.339 \pm 0.043$ & $0.347 \pm 0.041$ & $0.297 \pm 0.039$ & $0.272 \pm 0.036$ & $\boldsymbol{0.191 \pm 0.034}$ & $\boldsymbol{0.263 \pm 0.037}$ & $0.129 \pm 0.002$ \\
\textbf{TB-ANN} & $0.683 \pm 0.031^{\dagger}$ & $0.584 \pm 0.034^{\dagger}$ & $0.285 \pm 0.042^{\dagger}$ & $0.281 \pm 0.040^{\dagger}$ & $0.266 \pm 0.039^{\dagger}$ & $0.238 \pm 0.035^{\dagger}$ & $0.135 \pm 0.026^{\dagger}$ & $0.183 \pm 0.030^{\dagger}$ & $0.018 \pm 0.000$ \\
\textbf{TB-ANN-D} & $0.683 \pm 0.031$ & $0.584 \pm 0.034$ & $0.285 \pm 0.042$ & $0.281 \pm 0.040$ & $0.266 \pm 0.039$ & $0.238 \pm 0.035$ & $0.135 \pm 0.026$ & $0.183 \pm 0.030$ & $0.018 \pm 0.000$ \\
\textbf{TB-ANN-W} & $0.663 \pm 0.031^{\dagger}$ & $0.558 \pm 0.033^{\dagger}$ & $0.176 \pm 0.038^{\dagger}$ & $0.167 \pm 0.036^{\dagger}$ & $0.158 \pm 0.035^{\dagger}$ & $0.132 \pm 0.031$ & $0.085 \pm 0.024^{\dagger}$ & $0.107 \pm 0.027^{\dagger}$ & $0.020 \pm 0.000$ \\
\textbf{TB-ANN-W-D} & $0.663 \pm 0.031^{\dagger}$ & $0.558 \pm 0.033^{\dagger}$ & $0.176 \pm 0.038^{\dagger}$ & $0.167 \pm 0.036^{\dagger}$ & $0.158 \pm 0.035^{\dagger}$ & $0.132 \pm 0.031$ & $0.085 \pm 0.024^{\dagger}$ & $0.107 \pm 0.027^{\dagger}$ & $0.019 \pm 0.000$ \\
\textbf{TB-ENN} & $0.707 \pm 0.031$ & $0.614 \pm 0.033^{\dagger}$ & $0.339 \pm 0.043$ & $0.348 \pm 0.041^{\dagger}$ & $0.297 \pm 0.038$ & $0.273 \pm 0.035$ & $0.189 \pm 0.033$ & $0.263 \pm 0.038$ & $0.377 \pm 0.010$ \\
\textbf{TB-ENN-D} & $0.707 \pm 0.031$ & $0.614 \pm 0.033$ & $0.339 \pm 0.043$ & $0.348 \pm 0.041$ & $0.297 \pm 0.038$ & $0.273 \pm 0.035$ & $0.189 \pm 0.033$ & $0.263 \pm 0.038$ & $0.304 \pm 0.006$ \\
\textbf{TB-ENN-W} & $\boldsymbol{0.708 \pm 0.030^{\dagger\ddagger}}$ & $\boldsymbol{0.615 \pm 0.034^{\dagger\ddagger}}$ & $\boldsymbol{0.348 \pm 0.044^{\dagger\ddagger}}$ & $\boldsymbol{0.351 \pm 0.042^{\dagger\ddagger}}$ & $\boldsymbol{0.304 \pm 0.039}$ & $\boldsymbol{0.274 \pm 0.035}$ & $0.184 \pm 0.032$ & $0.254 \pm 0.038$ & $0.382 \pm 0.011$ \\
\textbf{TB-ENN-W-D} & $\boldsymbol{0.708 \pm 0.030^{\dagger\ddagger}}$ & $\boldsymbol{0.615 \pm 0.034^{\dagger\ddagger}}$ & $\boldsymbol{0.348 \pm 0.044^{\dagger\ddagger}}$ & $\boldsymbol{0.351 \pm 0.042^{\dagger\ddagger}}$ & $\boldsymbol{0.304 \pm 0.039}$ & $\boldsymbol{0.274 \pm 0.035}$ & $0.184 \pm 0.032$ & $0.254 \pm 0.038$ & $0.317 \pm 0.007$ \\
\midrule

\multicolumn{10}{l}{\texttt{qwen:32b}} \\
\arrayrulecolor{black}\specialrule{0.5pt}{0pt}{0pt}\arrayrulecolor{black}
\textbf{ANN} & $0.630 \pm 0.035$ & $0.503 \pm 0.036$ & $0.098 \pm 0.033$ & $0.098 \pm 0.033$ & $0.087 \pm 0.030$ & $0.080 \pm 0.028$ & $0.044 \pm 0.017$ & $0.059 \pm 0.020$ & $\boldsymbol{0.015 \pm 0.000}$ \\
\textbf{ENN} & $0.688 \pm 0.037$ & $0.575 \pm 0.040$ & $0.308 \pm 0.053$ & $0.313 \pm 0.048$ & $0.269 \pm 0.047$ & $0.240 \pm 0.040$ & $0.176 \pm 0.039$ & $0.241 \pm 0.042$ & $0.145 \pm 0.004$ \\
\textbf{TB-ANN} & $0.660 \pm 0.037^{\dagger}$ & $0.542 \pm 0.039^{\dagger}$ & $0.229 \pm 0.051^{\dagger}$ & $0.228 \pm 0.044^{\dagger}$ & $0.211 \pm 0.048^{\dagger}$ & $0.187 \pm 0.040^{\dagger}$ & $0.114 \pm 0.030^{\dagger}$ & $0.162 \pm 0.036^{\dagger}$ & $0.017 \pm 0.000$ \\
\textbf{TB-ANN-D} & $0.660 \pm 0.037$ & $0.542 \pm 0.039$ & $0.229 \pm 0.051$ & $0.228 \pm 0.044$ & $0.211 \pm 0.048$ & $0.187 \pm 0.040$ & $0.114 \pm 0.030$ & $0.162 \pm 0.036$ & $0.017 \pm 0.000$ \\
\textbf{TB-ANN-W} & $0.642 \pm 0.036^{\dagger}$ & $0.519 \pm 0.038^{\dagger}$ & $0.151 \pm 0.043^{\dagger}$ & $0.150 \pm 0.039^{\dagger}$ & $0.138 \pm 0.042^{\dagger}$ & $0.123 \pm 0.035^{\dagger}$ & $0.067 \pm 0.022^{\dagger}$ & $0.102 \pm 0.028^{\dagger}$ & $0.018 \pm 0.000$ \\
\textbf{TB-ANN-W-D} & $0.642 \pm 0.036^{\dagger}$ & $0.519 \pm 0.038^{\dagger}$ & $0.151 \pm 0.043^{\dagger}$ & $0.150 \pm 0.039^{\dagger}$ & $0.138 \pm 0.042^{\dagger}$ & $0.123 \pm 0.035^{\dagger}$ & $0.067 \pm 0.022^{\dagger}$ & $0.102 \pm 0.028^{\dagger}$ & $0.019 \pm 0.000$ \\
\textbf{TB-ENN} & $0.688 \pm 0.037$ & $0.575 \pm 0.040$ & $0.308 \pm 0.054$ & $0.311 \pm 0.049$ & $0.271 \pm 0.048$ & $0.238 \pm 0.040$ & $0.175 \pm 0.039$ & $0.239 \pm 0.043$ & $0.300 \pm 0.008$ \\
\textbf{TB-ENN-D} & $0.688 \pm 0.037$ & $0.575 \pm 0.040$ & $0.308 \pm 0.054$ & $0.311 \pm 0.049$ & $0.271 \pm 0.048$ & $0.238 \pm 0.040$ & $0.175 \pm 0.039$ & $0.239 \pm 0.043$ & $0.304 \pm 0.009$ \\
\textbf{TB-ENN-W} & $\boldsymbol{0.692 \pm 0.038^{\dagger\ddagger}}$ & $\boldsymbol{0.579 \pm 0.040^{\dagger\ddagger}}$ & $\boldsymbol{0.329 \pm 0.054^{\dagger\ddagger}}$ & $\boldsymbol{0.326 \pm 0.050^{\dagger\ddagger}}$ & $\boldsymbol{0.289 \pm 0.047}$ & $\boldsymbol{0.251 \pm 0.041}$ & $\boldsymbol{0.178 \pm 0.037}$ & $\boldsymbol{0.241 \pm 0.043}$ & $0.313 \pm 0.009$ \\
\textbf{TB-ENN-W-D} & $\boldsymbol{0.692 \pm 0.038^{\dagger\ddagger}}$ & $\boldsymbol{0.579 \pm 0.040^{\dagger\ddagger}}$ & $\boldsymbol{0.329 \pm 0.054^{\dagger\ddagger}}$ & $\boldsymbol{0.326 \pm 0.050^{\dagger\ddagger}}$ & $\boldsymbol{0.289 \pm 0.047}$ & $\boldsymbol{0.251 \pm 0.041}$ & $\boldsymbol{0.178 \pm 0.037}$ & $\boldsymbol{0.241 \pm 0.043}$ & $0.325 \pm 0.011$ \\
\bottomrule

\end{tabular}}
\caption{Performance metrics per method on \(t_{16}\), computed separately for \texttt{qwen:32b} and \texttt{llama3.3:70b}. Values are means over queries with 95\% bootstrap CIs for retrieval at $L \in \{3,5\}$. Results use relevant $c_p^{(c)}$ passages from 100 $t_{16}$ $c_p^{(a)}$ passages. Best values are bolded; $\dagger$ marks topic-based methods significantly outperforming baselines, and $\ddagger$ marks weighted topic-based methods significantly outperforming unweighted ones.}
\label{tab:retrieval_performance_t15_llama}
\end{table*}

\begin{table*}[!h]
\centering
\resizebox{\textwidth}{!}{%
\begin{tabular}{c ccccccccc}
\arrayrulecolor{black}
\toprule
\textbf{Method} & \textbf{MRR@3} & \textbf{MRR@5} & \textbf{NDCG@3} & \textbf{NDCG@5} & \textbf{Precision@3} & \textbf{Precision@5} & \textbf{Recall@3} & \textbf{Recall@5} & \textbf{Time (s)} \\
\midrule
\multicolumn{10}{l}{\texttt{gpt-4o}} \\
\arrayrulecolor{black}\specialrule{0.5pt}{0pt}{0pt}\arrayrulecolor{black}
\textbf{ANN} & $0.632 \pm 0.038$ & $0.521 \pm 0.042$ & $0.187 \pm 0.049$ & $0.181 \pm 0.046$ & $0.165 \pm 0.044$ & $0.143 \pm 0.038$ & $0.088 \pm 0.030$ & $0.119 \pm 0.035$ & $\boldsymbol{0.015 \pm 0.000}$ \\
\textbf{ENN} & $0.701 \pm 0.038$ & $0.611 \pm 0.039$ & $0.413 \pm 0.051$ & $0.438 \pm 0.044$ & $0.358 \pm 0.047$ & $0.331 \pm 0.040$ & $0.263 \pm 0.043$ & $0.386 \pm 0.045$ & $0.138 \pm 0.003$ \\
\textbf{TB-ANN} & $0.683 \pm 0.038^{\dagger}$ & $0.585 \pm 0.041^{\dagger}$ & $0.353 \pm 0.053^{\dagger}$ & $0.354 \pm 0.048^{\dagger}$ & $0.312 \pm 0.048^{\dagger}$ & $0.281 \pm 0.043^{\dagger}$ & $0.203 \pm 0.039^{\dagger}$ & $0.269 \pm 0.042^{\dagger}$ & $0.019 \pm 0.000$ \\
\textbf{TB-ANN-D} & $0.683 \pm 0.038$ & $0.585 \pm 0.041$ & $0.353 \pm 0.053$ & $0.354 \pm 0.048$ & $0.312 \pm 0.048$ & $0.281 \pm 0.043$ & $0.203 \pm 0.039$ & $0.269 \pm 0.042$ & $0.017 \pm 0.000$ \\
\textbf{TB-ANN-W} & $0.659 \pm 0.037^{\dagger}$ & $0.553 \pm 0.041^{\dagger}$ & $0.272 \pm 0.054^{\dagger}$ & $0.263 \pm 0.051^{\dagger}$ & $0.237 \pm 0.050^{\dagger}$ & $0.212 \pm 0.045^{\dagger}$ & $0.127 \pm 0.031^{\dagger}$ & $0.169 \pm 0.035^{\dagger}$ & $0.020 \pm 0.000$ \\
\textbf{TB-ANN-W-D} & $0.659 \pm 0.037^{\dagger}$ & $0.553 \pm 0.041^{\dagger}$ & $0.272 \pm 0.054^{\dagger}$ & $0.263 \pm 0.051^{\dagger}$ & $0.237 \pm 0.050^{\dagger}$ & $0.212 \pm 0.045^{\dagger}$ & $0.127 \pm 0.031^{\dagger}$ & $0.169 \pm 0.035^{\dagger}$ & $0.018 \pm 0.000$ \\
\textbf{TB-ENN} & $0.701 \pm 0.038$ & $0.611 \pm 0.039$ & $0.413 \pm 0.051$ & $0.437 \pm 0.044$ & $0.358 \pm 0.047$ & $0.329 \pm 0.041$ & $0.263 \pm 0.043$ & $0.386 \pm 0.046$ & $0.333 \pm 0.010$ \\
\textbf{TB-ENN-D} & $0.701 \pm 0.038$ & $0.611 \pm 0.039$ & $0.413 \pm 0.051$ & $0.437 \pm 0.044$ & $0.358 \pm 0.047$ & $0.329 \pm 0.041$ & $0.263 \pm 0.043$ & $0.386 \pm 0.046$ & $0.318 \pm 0.010$ \\
\textbf{TB-ENN-W} & $\boldsymbol{0.707 \pm 0.037^{\dagger\ddagger}}$ & $\boldsymbol{0.619 \pm 0.038^{\dagger\ddagger}}$ & $\boldsymbol{0.435 \pm 0.053^{\dagger\ddagger}}$ & $\boldsymbol{0.454 \pm 0.045^{\dagger\ddagger}}$ & $\boldsymbol{0.375 \pm 0.047^{\dagger\ddagger}}$ & $\boldsymbol{0.339 \pm 0.042}$ & $\boldsymbol{0.278 \pm 0.044^{\dagger\ddagger}}$ & $\boldsymbol{0.396 \pm 0.047}$ & $0.366 \pm 0.013$ \\
\textbf{TB-ENN-W-D} & $\boldsymbol{0.707 \pm 0.037^{\dagger\ddagger}}$ & $\boldsymbol{0.619 \pm 0.038^{\dagger\ddagger}}$ & $\boldsymbol{0.435 \pm 0.053^{\dagger\ddagger}}$ & $\boldsymbol{0.454 \pm 0.045^{\dagger\ddagger}}$ & $\boldsymbol{0.375 \pm 0.047^{\dagger\ddagger}}$ & $\boldsymbol{0.339 \pm 0.042}$ & $\boldsymbol{0.278 \pm 0.044^{\dagger\ddagger}}$ & $\boldsymbol{0.396 \pm 0.047}$ & $0.344 \pm 0.012$ \\
\midrule

\multicolumn{10}{l}{\texttt{llama3.3:70b}} \\
\arrayrulecolor{black}\specialrule{0.5pt}{0pt}{0pt}\arrayrulecolor{black}
\textbf{ANN} & $0.624 \pm 0.032$ & $0.520 \pm 0.034$ & $0.168 \pm 0.039$ & $0.169 \pm 0.039$ & $0.158 \pm 0.039$ & $0.147 \pm 0.036$ & $0.071 \pm 0.021$ & $0.099 \pm 0.027$ & $\boldsymbol{0.014 \pm 0.000}$ \\
\textbf{ENN} & $0.688 \pm 0.032$ & $0.603 \pm 0.035$ & $0.397 \pm 0.043$ & $0.416 \pm 0.040$ & $0.351 \pm 0.039$ & $0.322 \pm 0.034$ & $0.227 \pm 0.036$ & $\boldsymbol{0.339 \pm 0.042}$ & $0.129 \pm 0.001$ \\
\textbf{TB-ANN} & $0.672 \pm 0.032^{\dagger}$ & $0.582 \pm 0.035^{\dagger}$ & $0.335 \pm 0.047^{\dagger}$ & $0.342 \pm 0.043^{\dagger}$ & $0.304 \pm 0.043^{\dagger}$ & $0.281 \pm 0.038^{\dagger}$ & $0.172 \pm 0.032^{\dagger}$ & $0.241 \pm 0.035^{\dagger}$ & $0.017 \pm 0.000$ \\
\textbf{TB-ANN-D} & $0.672 \pm 0.032$ & $0.582 \pm 0.035$ & $0.335 \pm 0.047$ & $0.342 \pm 0.043$ & $0.304 \pm 0.043$ & $0.281 \pm 0.038$ & $0.172 \pm 0.032$ & $0.241 \pm 0.035$ & $0.017 \pm 0.000$ \\
\textbf{TB-ANN-W} & $0.648 \pm 0.033^{\dagger}$ & $0.550 \pm 0.035^{\dagger}$ & $0.261 \pm 0.047^{\dagger}$ & $0.259 \pm 0.044^{\dagger}$ & $0.242 \pm 0.045^{\dagger}$ & $0.220 \pm 0.038^{\dagger}$ & $0.112 \pm 0.026^{\dagger}$ & $0.161 \pm 0.031^{\dagger}$ & $0.018 \pm 0.000$ \\
\textbf{TB-ANN-W-D} & $0.648 \pm 0.033^{\dagger}$ & $0.550 \pm 0.035^{\dagger}$ & $0.261 \pm 0.047^{\dagger}$ & $0.259 \pm 0.044^{\dagger}$ & $0.242 \pm 0.045^{\dagger}$ & $0.220 \pm 0.038^{\dagger}$ & $0.112 \pm 0.026^{\dagger}$ & $0.161 \pm 0.031^{\dagger}$ & $0.018 \pm 0.000$ \\
\textbf{TB-ENN} & $0.688 \pm 0.032$ & $0.603 \pm 0.035$ & $0.398 \pm 0.043$ & $0.416 \pm 0.039$ & $0.351 \pm 0.039$ & $0.321 \pm 0.034$ & $0.227 \pm 0.036$ & $0.338 \pm 0.042$ & $0.305 \pm 0.009$ \\
\textbf{TB-ENN-D} & $0.688 \pm 0.032$ & $0.603 \pm 0.035$ & $0.398 \pm 0.043$ & $0.416 \pm 0.039$ & $0.351 \pm 0.039$ & $0.321 \pm 0.034$ & $0.227 \pm 0.036$ & $0.338 \pm 0.042$ & $0.306 \pm 0.009$ \\
\textbf{TB-ENN-W} & $\boldsymbol{0.689 \pm 0.033}$ & $\boldsymbol{0.604 \pm 0.035^{\dagger}}$ & $\boldsymbol{0.403 \pm 0.046}$ & $\boldsymbol{0.419 \pm 0.043^{\dagger}}$ & $\boldsymbol{0.357 \pm 0.042}$ & $\boldsymbol{0.328 \pm 0.036}$ & $\boldsymbol{0.231 \pm 0.039}$ & $0.333 \pm 0.041$ & $0.332 \pm 0.011$ \\
\textbf{TB-ENN-W-D} & $\boldsymbol{0.689 \pm 0.033}$ & $\boldsymbol{0.604 \pm 0.035^{\dagger}}$ & $\boldsymbol{0.403 \pm 0.046}$ & $\boldsymbol{0.419 \pm 0.043^{\dagger}}$ & $\boldsymbol{0.357 \pm 0.042}$ & $\boldsymbol{0.328 \pm 0.036}$ & $\boldsymbol{0.231 \pm 0.039}$ & $0.333 \pm 0.041$ & $0.332 \pm 0.011$ \\
\midrule

\multicolumn{10}{l}{\texttt{qwen:32b}} \\
\arrayrulecolor{black}\specialrule{0.5pt}{0pt}{0pt}\arrayrulecolor{black}
\textbf{ANN} & $0.607 \pm 0.038$ & $0.488 \pm 0.041$ & $0.146 \pm 0.048$ & $0.146 \pm 0.046$ & $0.126 \pm 0.040$ & $0.115 \pm 0.038$ & $0.078 \pm 0.034$ & $0.100 \pm 0.037$ & $\boldsymbol{0.014 \pm 0.000}$ \\
\textbf{ENN} & $0.662 \pm 0.042$ & $0.559 \pm 0.044$ & $0.352 \pm 0.058$ & $0.361 \pm 0.054$ & $0.301 \pm 0.050$ & $0.270 \pm 0.042$ & $0.224 \pm 0.050$ & $0.301 \pm 0.054$ & $0.130 \pm 0.001$ \\
\textbf{TB-ANN} & $0.657 \pm 0.041^{\dagger}$ & $0.551 \pm 0.045^{\dagger}$ & $0.319 \pm 0.058^{\dagger}$ & $0.324 \pm 0.055^{\dagger}$ & $0.276 \pm 0.051^{\dagger}$ & $0.248 \pm 0.045^{\dagger}$ & $0.184 \pm 0.042^{\dagger}$ & $0.253 \pm 0.049^{\dagger}$ & $0.017 \pm 0.000$ \\
\textbf{TB-ANN-D} & $0.657 \pm 0.041$ & $0.551 \pm 0.045$ & $0.319 \pm 0.058$ & $0.324 \pm 0.055$ & $0.276 \pm 0.051$ & $0.248 \pm 0.045$ & $0.184 \pm 0.042$ & $0.253 \pm 0.049$ & $0.017 \pm 0.000$ \\
\textbf{TB-ANN-W} & $0.621 \pm 0.041^{\dagger}$ & $0.504 \pm 0.045^{\dagger}$ & $0.219 \pm 0.058^{\dagger}$ & $0.214 \pm 0.053^{\dagger}$ & $0.196 \pm 0.053^{\dagger}$ & $0.170 \pm 0.045^{\dagger}$ & $0.109 \pm 0.033^{\dagger}$ & $0.146 \pm 0.039^{\dagger}$ & $0.018 \pm 0.000$ \\
\textbf{TB-ANN-W-D} & $0.621 \pm 0.041^{\dagger}$ & $0.504 \pm 0.045^{\dagger}$ & $0.219 \pm 0.058^{\dagger}$ & $0.214 \pm 0.053^{\dagger}$ & $0.196 \pm 0.053^{\dagger}$ & $0.170 \pm 0.045^{\dagger}$ & $0.109 \pm 0.033^{\dagger}$ & $0.146 \pm 0.039^{\dagger}$ & $0.018 \pm 0.000$ \\
\textbf{TB-ENN} & $0.661 \pm 0.042$ & $0.558 \pm 0.045$ & $0.349 \pm 0.059$ & $0.357 \pm 0.055$ & $0.299 \pm 0.049$ & $0.267 \pm 0.043$ & $0.223 \pm 0.050$ & $0.298 \pm 0.054$ & $0.317 \pm 0.010$ \\
\textbf{TB-ENN-D} & $0.661 \pm 0.042$ & $0.558 \pm 0.045$ & $0.349 \pm 0.059$ & $0.357 \pm 0.055$ & $0.299 \pm 0.049$ & $0.267 \pm 0.043$ & $0.223 \pm 0.050$ & $0.298 \pm 0.054$ & $0.314 \pm 0.010$ \\
\textbf{TB-ENN-W} & $\boldsymbol{0.671 \pm 0.042^{\dagger\ddagger}}$ & $\boldsymbol{0.570 \pm 0.045^{\dagger\ddagger}}$ & $\boldsymbol{0.370 \pm 0.061^{\dagger\ddagger}}$ & $\boldsymbol{0.384 \pm 0.058^{\dagger\ddagger}}$ & $\boldsymbol{0.313 \pm 0.051}$ & $\boldsymbol{0.286 \pm 0.047^{\ddagger}}$ & $\boldsymbol{0.241 \pm 0.051}$ & $\boldsymbol{0.319 \pm 0.054^{\ddagger}}$ & $0.350 \pm 0.014$ \\
\textbf{TB-ENN-W-D} & $\boldsymbol{0.671 \pm 0.042^{\dagger\ddagger}}$ & $\boldsymbol{0.570 \pm 0.045^{\dagger\ddagger}}$ & $\boldsymbol{0.370 \pm 0.061^{\dagger\ddagger}}$ & $\boldsymbol{0.384 \pm 0.058^{\dagger\ddagger}}$ & $\boldsymbol{0.313 \pm 0.051}$ & $\boldsymbol{0.286 \pm 0.047^{\ddagger}}$ & $\boldsymbol{0.241 \pm 0.051}$ & $\boldsymbol{0.319 \pm 0.054^{\ddagger}}$ & $0.346 \pm 0.014$ \\
\bottomrule

\end{tabular}}
\caption{Performance metrics per method on \(t_{12}\), computed separately for each \mm{}. Values are means over queries with 95\% bootstrap CIs for retrieval at $L \in \{3,5\}$. Results use relevant $c_p^{(c)}$ passages from 100 $t_{11}$ $c_p^{(a)}$ passages. Best values are bolded; $\dagger$ marks topic-based methods significantly outperforming baselines, and $\ddagger$ marks weighted topic-based methods significantly outperforming unweighted ones.}
\label{tab:retrieval_performance_t11}
\end{table*}
\begin{figure*}[!h]
    \centering
    \includegraphics[width=0.75\textwidth]{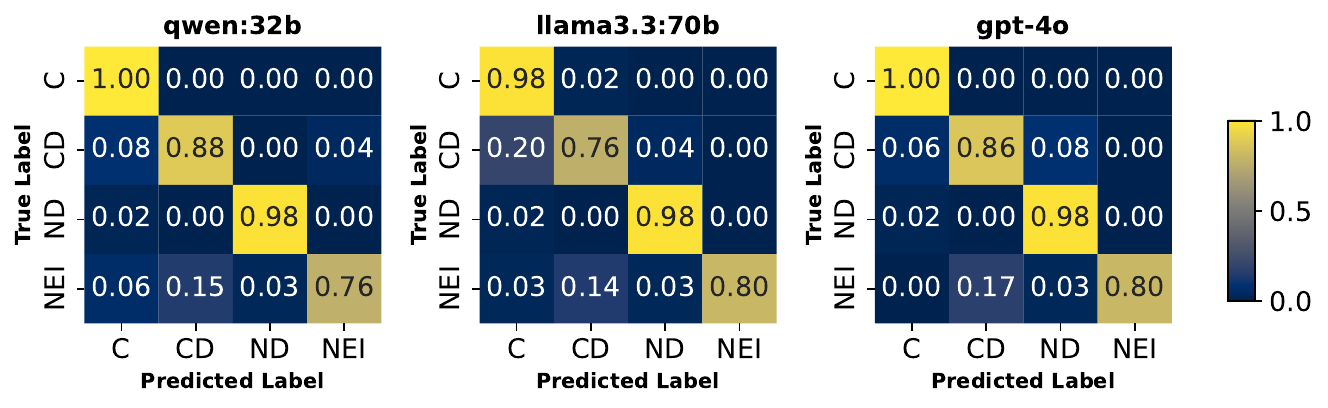}
    \caption{Confusion matrices for discrepancy classification across models on the \texttt{FEVER-DPLACE} dataset. 
    All models perform strongly, though \texttt{qwen:32b} and \texttt{gpt-4o} show more confusion between \con{} and \nd{}, while \texttt{llama3.3:70b} tends to mistake \cd{} for \con{} or \nd{}.
    }
    \label{fig:confusion_fever_d_place_q}
\end{figure*}

Table~\ref{tab:statistics} summarizes questions and queries generated from passages assigned to $t_{12}$ or $t_{16}$ in the 100-sample anchor corpus. \texttt{llama3.3:70b} generates the highest number of questions, while \texttt{gpt-4o} produces fewer but shorter questions. The number of queries per questions remains relatively stable across models, but query length varies, with \texttt{llama3.3:70b} generating the longest queries, particularly for \(t_{12}\). The number of relevant passages for questions from \texttt{gpt-4o} and \texttt{llama3.3:70b} is higher, especially for the latter with \(t_{12}\).

Table~\ref{tab:retrieval_performance_t15_llama} and~\ref{tab:retrieval_performance_t11} contains the retrieval performance for \texttt{qwen:32b} and \texttt{llama3.3:70b} for \(t_{16}\), and that of the three considered \mm{} for \(t_{25}\). 
TB-ENN-W achieves the best performance, with its dynamic version showing marginal improvements for \texttt{qwen:32b} in topics and for \texttt{llama3.3:70b} at \(t_{16}\). Moreover, ENN surpasses topic-based methods in Recall for \texttt{llama3.3:70b} queries at \(t_{16}\).

Fig.~\ref{fig:confusion_fever_d_place_q} compares discrepancy classification across the evaluated \mm{}s on \texttt{FEVER-DPLACE} by means of their confusion matrices, showing that \texttt{gpt-4o} maintains the most consistent balance across classes, while \texttt{llama3.3:70b} has greater difficulty with \cd{}, and \texttt{qwen:32b} is less reliable on \nei{}.

\section{Infrastructure and LLM deployment}\label{appen:infrastructure}
Experiments were conducted on a server equipped with an Intel(R) Xeon(R) Gold 5318Y CPU @ 2.10GHz (48 cores, 96 threads), 500 GB RAM, and three NVIDIA GeForce RTX 4090 GPUs. Running \ling{} on 500 anchor passages per topic took about one day (time varies with the relevant passages count and the underlying \mm{}).
Open-source models are deployed locally using \texttt{ollama}\footnote{\url{https://ollama.com/}} and OpenAI's \texttt{gpt-4o-2024-08-06} checkpoint is used.

\section{Examples from the main text}\label{appen:examples}
\subsection{Answers failing in the comparison corpus}\label{example:comp_answer_failures}
The following two answers from the comparison corpora illustrate why answer quality is sometimes lower than in the anchor corpora. In both cases, the retrieved passage is thematically or semantically relevant but fails to provide the information needed to answer the question directly.  \newline

\noindent {\bf Example A1:}
\begin{itemize}
    \item \textbf{\(\mathbf{q_p^{(a)}}\):} Is a lack of prenatal care associated with an increased risk of SIDS? 
    \item \textbf{\(\mathbf{c_p^{(c)}}\):} Mayor riesgo síndrome de muerte infantil súbita (SIDS, por sus siglas en inglés).
    \item \textbf{\(\mathbf{a_p^{(c)}}\):} Yes, a lack of prenatal care is associated with an increased risk of SIDS. The passage mentions that babies born to smoking mothers have a higher risk of sudden infant death syndrome (SIDS), which implies that prenatal exposure to smoking, a factor often linked to inadequate prenatal care, can increase the risk of SIDS.
\end{itemize}

\noindent {\bf Example A2:}
\begin{itemize}
    \item \textbf{\(\mathbf{q_p^{(a)}}\):} Are blood work and examinations part of the required monitoring after a transplant?    
    \item \textbf{\(\mathbf{c_p^{(c)}}\):} Monitorización de los sucesivos exámenes de rayos X y de sangre.
    \item \textbf{\(\mathbf{a_p^{(c)}}\):} Yes, blood work and examinations are part of the required monitoring, as indicated by the mention of ``exámenes de sangre'' (blood tests) and ``exámenes de rayos X'' (X-ray exams) in the monitoring process.
\end{itemize}

\subsection{Cultural Discrepancies missed by annotators}\label{example:cd_failures}
The following examples illustrate the full \ling{} process: generating a question from the anchor passage, checking discrepancies between the answers derived from the anchor and the retrieved comparison passage, and providing an explanation of the discrepancy (full documents omitted due to length) to support the discussion in \S\ref{sec:mind_discrpancies}. \newline

\noindent {\bf Example D1:}
\begin{itemize}
    \item \textbf{\(\mathbf{c_p^{(a)}}\):} All newborns undergo painless cardiac and hearing screening tests. The hearing test is performed by a newborn hearing screening technician and the cardiac test, which looks at your baby’s heart health, will be performed by your nurse. These tests will be done in the mother and baby unit before you go home.	
    \item \textbf{\(\mathbf{q_p^{(a)}}\):} Are all newborns required to undergo cardiac screening tests?	
    \item \textbf{\(\mathbf{a_p^{(a)}}\):} Yes, all newborns are required to undergo cardiac screening tests. These tests are performed in the mother and baby unit before discharge and are part of the standard newborn screenings, which also include hearing tests.
    \item \textbf{\(\mathbf{c_p^{(c)}}\):} Antes de darle de alta del hospital, a su bebé le hacen pruebas para detectar defectos cardíacos congénitos críticos como parte de las pruebas de detección para recién nacidos. Las pruebas de detección para recién nacidos detectan condiciones que son graves pero poco comunes y mayormente tratables. Estas incluyen una prueba de sangre, de audición y del corazón. Todos los estados exigen las pruebas de detección para recién nacidos y todos, excepto California, exigen las pruebas de detección de defectos cardíacos congénitos críticos. California requiere a los profesionales ofrecer la detección de los defectos cardíacos congénitos críticos.
    \item \textbf{\(\mathbf{a_p^{(c)}}\):} No, not all newborns are required to undergo cardiac screening tests because while almost all states require these screenings as part of newborn screening tests, there is an exception mentioned for California, which requires healthcare professionals to offer the screening but does not mandate it.	
    \item {\bf Discrepancy: } \cd{}
    \item {\bf Reason: } The answers provide information that seems contradictory at first glance, with \(a_p^{(a)}\) stating all newborns are required to undergo cardiac screening tests and \(a_p^{(c)}\) indicating not all are required due to an exception in California. However, upon closer examination, it becomes clear that the discrepancy arises from a specific regulatory difference in one region (California) rather than a fundamental contradiction in medical practice or recommendation across all contexts. Both answers can be true depending on the geographical location within the United States, reflecting variations in state health regulations.	
\end{itemize}

\noindent {\bf Example D2:}
\begin{itemize}
    \item \textbf{\(\mathbf{c_p^{(a)}}\):} Your child can return to daycare or school 48 hours after starting antibiotic treatment as long as there are signs of improvement.		
    \item \textbf{\(\mathbf{q_p^{(a)}}\):} Is it necessary for a child with impetigo to wait more than 48 hours after starting antibiotic treatment before returning to daycare or school if they are improving?	
    \item \textbf{\(\mathbf{a_p^{(a)}}\):} No, it is not necessary for a child with impetigo to wait more than 48 hours after starting antibiotic treatment before returning to daycare or school if they are improving. According to the information provided, a child can return to daycare or school 48 hours after starting antibiotic treatment as long as there are signs of improvement.	
    \item \textbf{\(\mathbf{c_p^{(c)}}\):} Los estudiantes con impétigo suelen poder volver a la escuela 24 horas después de comenzar el tratamiento.	
    \item \textbf{\(\mathbf{a_p^{(c)}}\):} No, it is not necessary for a child with impetigo to wait more than 48 hours after starting antibiotic treatment before returning to daycare or school if they are improving. They can usually return to school 24 hours after beginning treatment.	
    \item {\bf Discrepancy: } \cd{}
    \item {\bf Reason: } Both \(a_p^{(a)}\) and \(a_p^{(c)}\) agree that it is not necessary for a child with impetigo to wait more than 48 hours after starting antibiotic treatment before returning to daycare or school if they are improving. However, they provide slightly different timeframes (48 hours vs. 24 hours) after which the child can return. Since both answers essentially support the notion that a wait of more than 48 hours is unnecessary and only differ by 24 hours in their recommended return time, they do not directly contradict each other on the core issue (the necessity of waiting more than 48 hours), but rather reflect a difference due to variations in medical advice or school policies.
\end{itemize}

\section{\texttt{ROSIE-MIND}}\label{appen:rosie_mind}
The dataset generated from this study consists of two parts. \texttt{ROSIE-MIND-v1} was produced in an initial run using TB-ENN-W-D with the embedding model \texttt{quora-distilbert-multilingual} and \texttt{qwen:32b}, the best-performing configuration under those conditions. In this setting, we used \ling{} to analyze \num{4000} {\em anchor} passages and detected \num{75}, \num{72}, and \num{40} \con{}, \num{66}, \num{127}, and \num{206} \cd{}, and \num{2997}, \num{2342}, and \num{2507} \nd{} instances in the analyzed passages for \(t_{12}\), \(t_{16}\), and \(t_{25}\), respectively, with around \num{42}K \nei{} instances per topic. We then randomly selected 80 generated triplets, and refined them to create \texttt{ROSIE-MIND-v1}.

To test whether an asymmetric embedding model would improve results, we repeated the experiments with \texttt{BAAI/bge-m3}, whose results are reported in the main text (\S\hyperref[sec:mind_discrpancies]{\ref{sec:mind_discrpancies}}). Annotators found that this run produced more relevant retrieved passages and higher-quality answers, resulting in a total of 584 annotated samples. Table~\ref{tab:rosie_mind_examples} presents examples of \con{}, \nd{}, and \cd{} detected by \ling{} and annotated as such, along with cases where \ling{} failed (with reasons for failure given in parenthesis after the label). 

Ultimately, both the embedding model and the underlying \mm{} in \ling{} are configurable parameters; the configurations reported here reflect the best-performing options at the time of experimentation, while the framework remains open to further improvements with alternative choices.

In the creation of the final released dataset, instances in which false discrepancies arose due to failures to decontextualize content were excluded. This includes failures to remove anecdotal or non-generalizable references (e.g., \textit{``Was a tracheostomy placed in \textbf{the female patient} before she was born?''} or \textit{``Did \textbf{Liz} become pregnant after undergoing in vitro fertilization (IVF) treatments?''}) and vague contextual references (e.g., {\it ``Do you typically stay in the hospital for more than 3 days after {\bf the operation}?''} or {\it ``Do kids {\bf around this age} need at least 9 hours of sleep per night?''}).

\section{Note on discrepancies framing}\label{appen:discrepancy_framing}
Some of the discrepancies labelled here may not match the anthropological framing in the strict sense. In some cases, the distinction arises from the passages being rooted in different time frames, which is typical when scraping web sources, i.e., they were true at one point but not at another. For example, two answers may pose a factual contradiction as in \hyperlink{T_11_1655}{T-11-1655}, but on closer inspection they are better framed as a case of \underline{temporal} or \underline{historical variation}: the English source reflects modern direct-acting antivirals, which can cure hepatitis C, whereas the Spanish source reflects older, partially effective approaches. 
In other cases, the divergence comes not from time but from \underline{differences in emphasis or risk framing}. For instance, in \hyperlink{T_11_6088}{T-11-6088}, one passage (and its corresponding answer) states that such an attack is possible and stresses the urgency of preparedness, whereas the other considers it unlikely. Although the answers seem contradictory in isolation, they in fact represent different assessments of risk.  

This does not diminish the capabilities of \ling{}. First, the pipeline is easily adaptable to different discrepancy categories by simply modifying them in the corresponding prompt. Second, and ultimately, {\em it is the user who must determine whether a case constitutes a discrepancy within the scope of their study}. In the released dataset, these nuances are included as additional metadata.

\section{WIKI-EN-DE}\label{appen:wiki_ende}
This dataset was constructed {\em ad hoc} to test the generalizability of \ling{} across domains and languages. It consists of 600 pairs of Wikipedia articles in German and English, created by scraping articles that have corresponding entries in the other language, and spans a wide range of topics such as world history, geography, and politics.

Since each of the 600 articles can span multiple pages, we increase granularity from the article level to the paragraph level using a custom segmenter that splits the raw text at each newline and filters empty strings and ill-encoded characters to prevent error propagation. This results in an average of 27.48 passages per German article and 57.86 per English article. The final dataset consists of \num{17069}  English (anchor) and \num{8079} German (comparison) passages. Obviously, after this segmentation we no longer have one-to-one alignment, so we fulfill the {\em loose} alignment requirement of \pltm{} through MT using \texttt{OPUS-MT}, and NLP preprocessing followed the same procedure as for \texttt{ROSIE} (\S\ref{appen:preprocessing_details}).

We train several \pltm{} models with $K \in [5, 50]$ and select $K = 25$ as the final number of topics (see Table~\ref{tab:25_tpcs}). From these, we randomly sample 150 anchor passages whose primary topic is either {\it Christian Communion Practices} (\(t_{4}\)) or {\it Freemasonry Traditions} (\(t_{6}\)) to analyze with \ling{}, encompassing a total of 8 and 5 full articles inspected for topics \(t_{4}\) and \(t_{6}\), respectively. 

Following the convention used for \texttt{ROSIE}, we reference examples from \texttt{WIKI-EN-DE} (Table~\ref{tab:ende_mind_examples}) using the format \texttt{E-T-\{topic\_id\}-\{passage\_id\}}.
While more passages would need to be inspected to yield broader findings, this initial analysis already reveals meaningful examples of discrepancies. In addition to the examples shown in the main text (\S\hyperref[sec:mind_discrpancies]{\ref{sec:mind_discrpancies}}), other cases also illustrate the range of outputs: \hyperlink{E_T_3_8155}{E-T-3-8155} shows a \cd{} rooted in theology, with Catholic doctrine affirming the Eucharist as the same sacrifice as the Cross while Methodist sources describe it as a re-presentation. Obviously, Wikipedia pages are not always fully unfaithful to one another, as in \hyperlink{E_T_5_928}{E-T-5-928}, which yields \nd{} since both English and German passages confirm that graphic design principles shape effective map presentation. 

Still, as with \texttt{ROSIE}, \ling{} also generates false positive: for example, \hyperlink{E_T_3_6847}{E-T-3-6847} was incorrectly labelled as \cd{} since, based on the comparison passage {\it ``Berthold Seewald: Ihren ersten Krieg führten die USA gegen muslimische Piraten In: Die Welt veröffentlicht am 17. Februar 2019''}, it generated the answers {\it ``Yes, the United States of America have entered into a war or act of hostility against Muslim entities, specifically against Muslim pirates, as indicated by the mention of their first war being fought against Muslim pirates.''} despite the passage not containing such information.

\section{Prompts}
\ref{prompt:question_generation},~\ref{prompt:query_generation},~\ref{prompt:answer_generation},~\ref{prompt:discrepancy_detection} contain the prompts for the modules embedded within \ling{}. \ref{prompt:relevant_passages_identification} and~\ref{prompt:labeling} contain the prompts for identifying relevant passages for the retrieval evaluation and for generating topic labels, respectively. \ref{prompt:fever_conversion} and~\ref{prompt:dplace_conversion} contain the prompts for generating \texttt{FEVER-DPLACE-Q}.

\section{Instructions for annotators}\label{appen:instructions}
\begin{figure*}[!ht]
    \centering
    \includegraphics[width=0.7\textwidth]{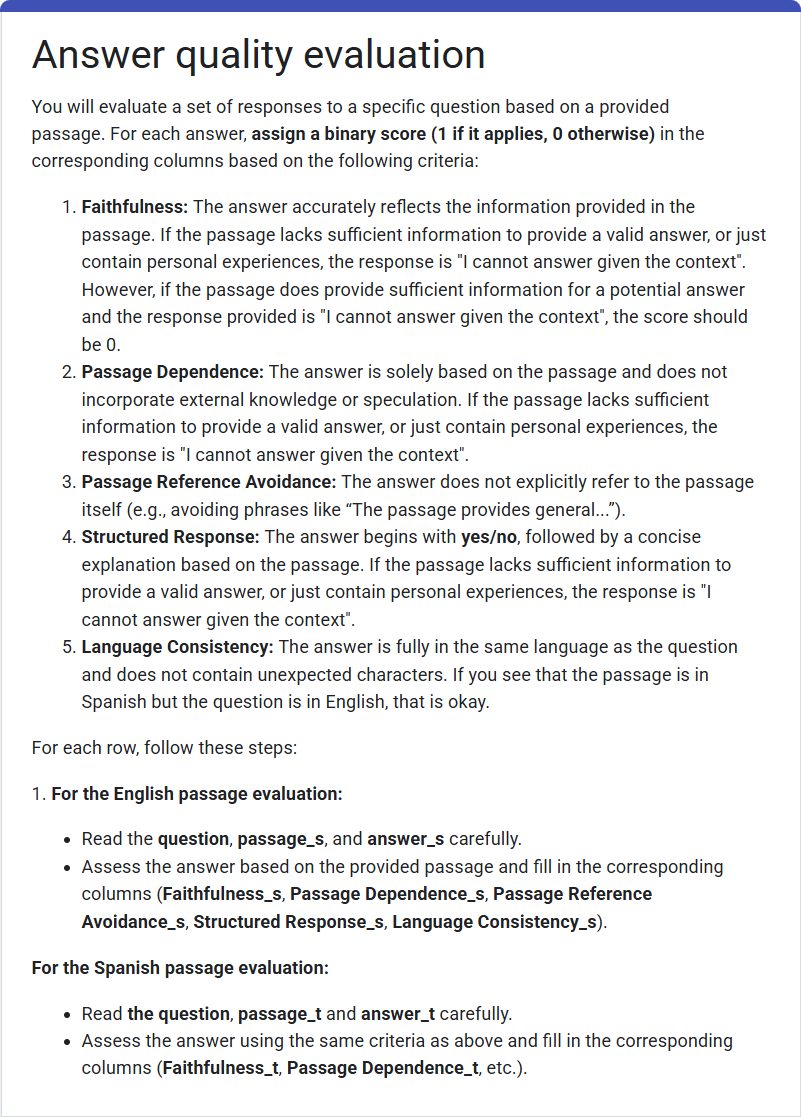}
    \caption{Instructions for Answer quality evaluation.}
    \label{fig:answer_instructions_for_annotators}
\end{figure*}

\begin{figure*}[!ht]
    \centering
    \includegraphics[width=0.7\textwidth]{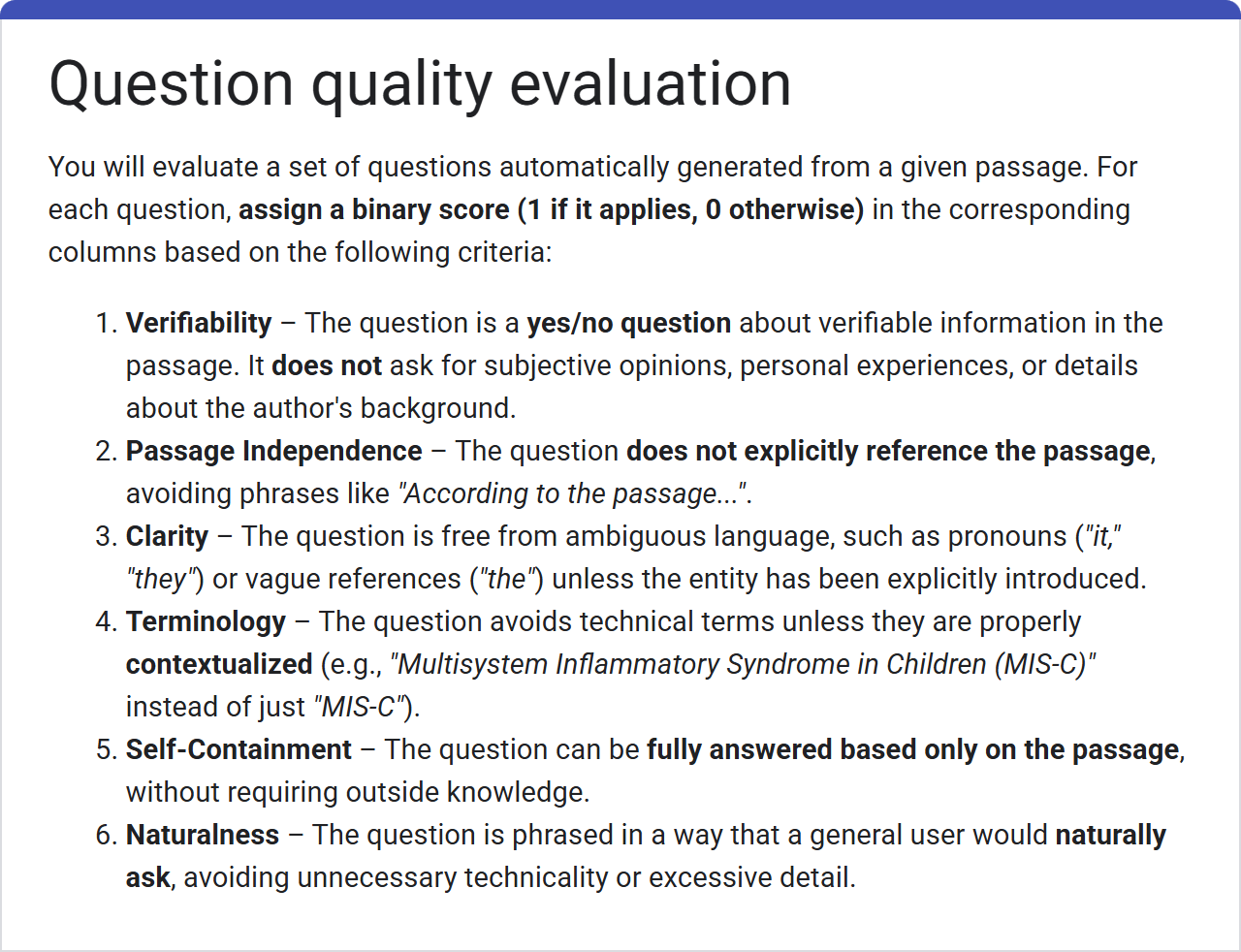}
    \caption{Instructions for Question quality evaluation.}
    \label{fig:question_instructions_for_annotators}
\end{figure*}

\begin{figure*}[!ht]
    \centering
    \includegraphics[width=0.7\textwidth]{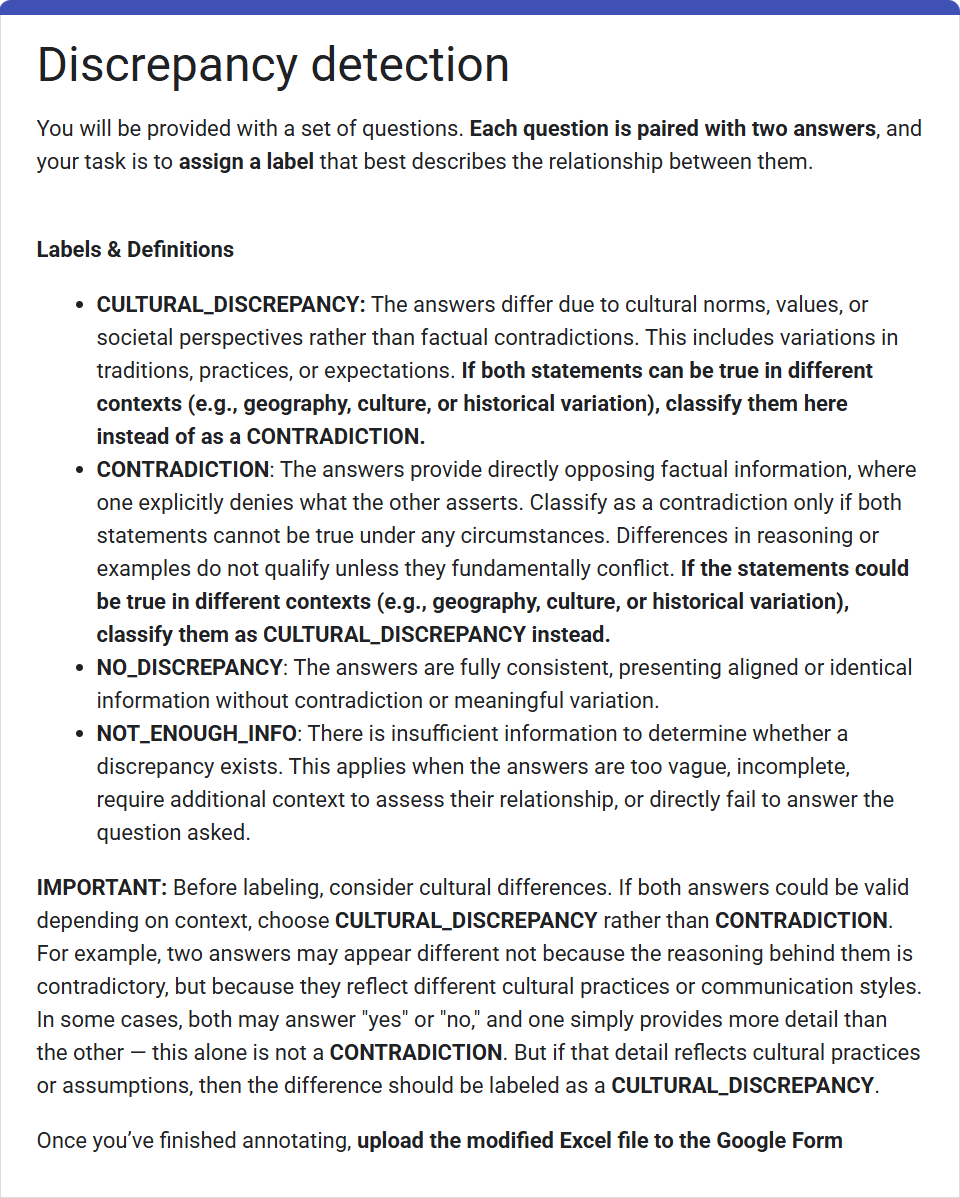}
    \caption{Instructions for Discrepancy classification and evaluation.}
    \label{fig:discrepancy_instructions_for_annotators}
\end{figure*}

\begin{figure*}[!ht]
    \centering
    \includegraphics[width=0.7\textwidth]{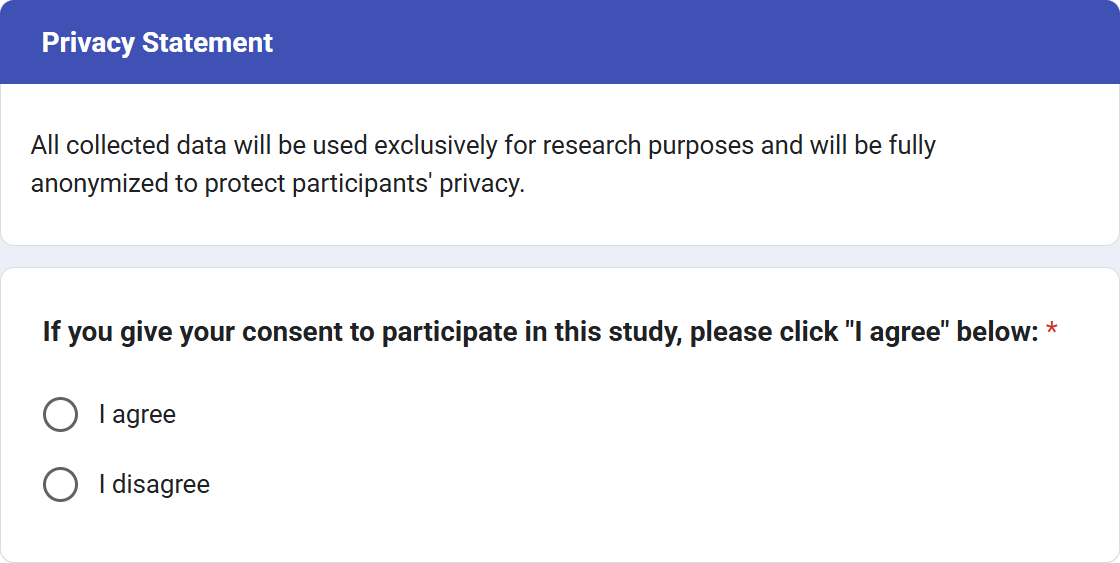}
    \caption{Consent Form.}
    \label{fig:consent}
\end{figure*}

Figs.~\ref{fig:discrepancy_instructions_for_annotators},~\ref{fig:question_instructions_for_annotators}, and~\ref{fig:answer_instructions_for_annotators} show the instructions provided to annotators for conducting the annotation tasks: question generation quality assessment, answer generation quality assessment, and discrepancy detection classification. 
Fig.~\ref{fig:consent} presents the consent form that annotators had to accept before participating in the study.

\section{MIND-WEB GUI}\label{appen:screenshots}
\begin{figure*}[!ht]
    \centering
     \adjustbox{fbox}{\includegraphics[width=0.7\textwidth]{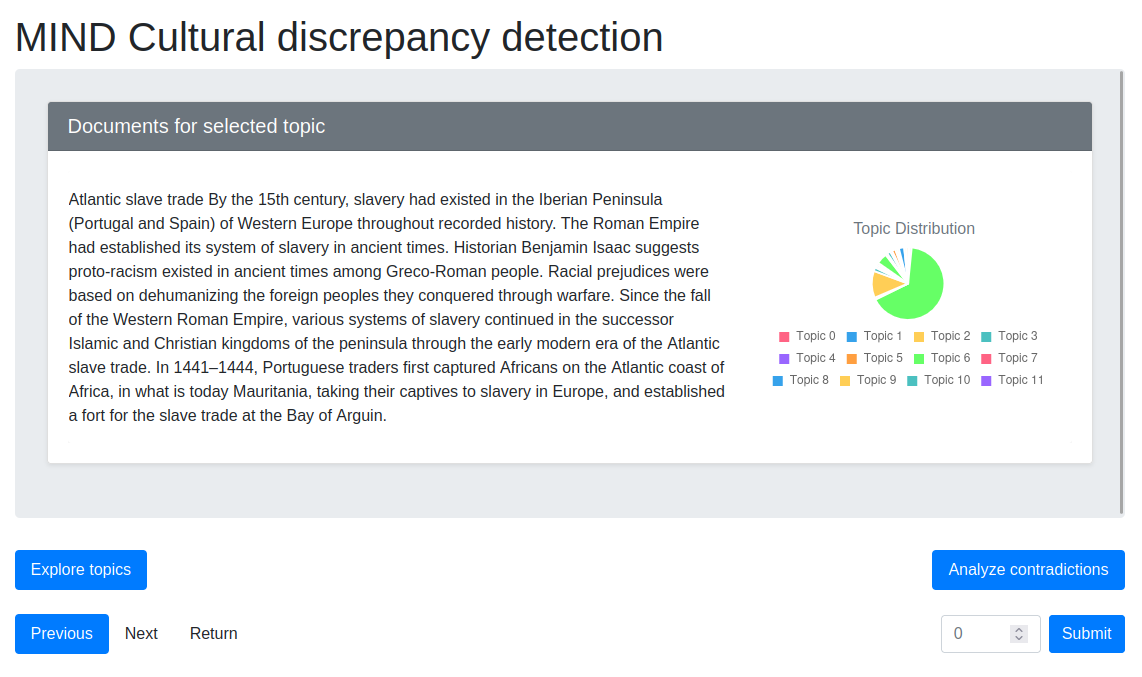}}
    \caption{Topic inspection view.}
    \label{fig:topic_gui}
\end{figure*}

\begin{figure*}[!ht]
    \centering
     \adjustbox{fbox}{\includegraphics[width=0.7\textwidth]{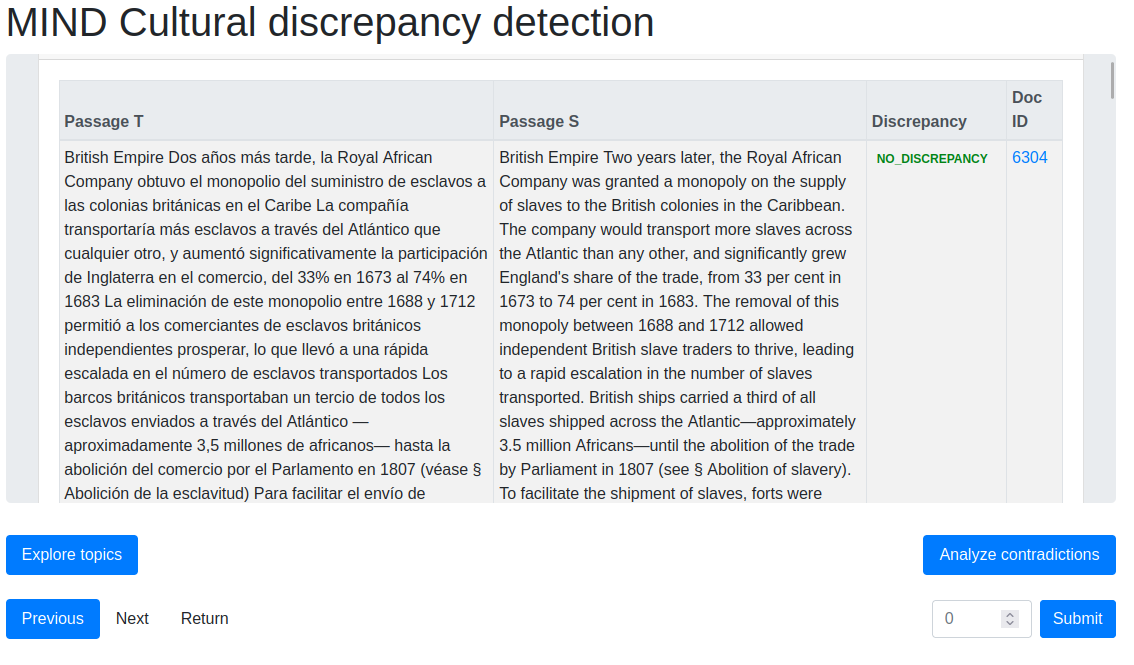}}
    \caption{Discrepancy detection view.}
    \label{fig:disc_gui}
\end{figure*}

To better support user-in-the-loop interaction, we complemented the CLI version of MIND with a GUI-based web tool (Figs.~\ref{fig:topic_gui} and \ref{fig:disc_gui}).

\section{Topic Models}\label{appen:topics}
Tables~\ref{tab:15_tpcs},~\ref{tab:20_tpcs}, and~\ref{tab:30_tpcs} show the 15-topic, 20-topic, and 30-topic \pltm{} models trained on the \texttt{ROSIE} dataset, respectively, while Table~\ref{tab:25_tpcs} shows the 25-topic model trained on \texttt{WIKI-EN-DE}.

\onecolumn

\renewcommand{\arraystretch}{1.3}
\setlength{\tabcolsep}{2pt}

\renewcommand{\arraystretch}{1.2}

\begingroup %
\footnotesize
\centering

\begin{longtable}{p{2cm} p{13cm}}
\caption{\texttt{ROSIE-MIND} examples.} \label{tab:rosie_mind_examples} \\

\toprule
\textbf{ID} & \textbf{Content} \\ 
\midrule
\endfirsthead

\multicolumn{2}{c}{\textit{(Continued from previous page)}} \\
\toprule
\textbf{ID} & \textbf{Content} \\ 
\midrule
\endhead

\bottomrule
\multicolumn{2}{r}{\textit{(Continued on next page)}} \\
\endfoot

\bottomrule
\endlastfoot

\hypertarget{T_11_455}{\textbf{T-11-455}}
& \textbf{\(\mathbf{q_p^{(a)}}\):} Do autism spectrum disorders (ASDs) affect at least 1 in every 88 children in the United States, according to the Centers for Disease Control and Prevention (CDC)? \\
& \textbf{\(\mathbf{c_p^{(a)}}\):} The Centers for Disease Control and Prevention (CDC) estimates that autism spectrum disorders affect 1 in every 88 children in the United States. Although ASDs are more common today than childhood cancer, juvenile diabetes and pediatric AIDS combined, little is known about their cause or the most effective treatment.\\
& \textbf{\(\mathbf{a_p^{(a)}}\):} Yes, according to the Centers for Disease Control and Prevention (CDC), autism spectrum disorders (ASDs) affect at least 1 in every 88 children in the United States. This information is directly provided, indicating a clear prevalence rate of ASDs among children in the U.S. as estimated by the CDC. \\
& \textbf{\(\mathbf{c_p^{(c)}}\):} ¿A quién afecta el ASD: Según los CDC, aproximadamente 1 de cada 68 niños en los EE.UU. tiene un trastorno del espectro autista. El trastorno se presenta con mucha más frecuencia en los niños que en las niñas. De 4 a 5 veces más niños que niñas tienen ASD. \\ 
& \textbf{\(\mathbf{a_p^{(c)}}\):} No, according to the information provided, autism spectrum disorders (ASDs) affect approximately 1 in every 68 children in the United States, not at least 1 in every 88, as stated by the Centers for Disease Control and Prevention (CDC). \\  
& \textbf{Discrepancy:} \abr{contradiction} \\  
\midrule

\hypertarget{T_11_457}{\textbf{T-11-457}}
& \textbf{\(\mathbf{q_p^{(a)}}\):} Do autism spectrum disorders (ASDs) affect at least 1 in every 88 children in the United States, according to the Centers for Disease Control and Prevention (CDC)? \\
& \textbf{\(\mathbf{c_p^{(a)}}\):} The Centers for Disease Control and Prevention (CDC) estimates that autism spectrum disorders affect 1 in every 88 children in the United States. Although ASDs are more common today than childhood cancer, juvenile diabetes and pediatric AIDS combined, little is known about their cause or the most effective treatment.\\
& \textbf{\(\mathbf{a_p^{(a)}}\):} Yes, according to the Centers for Disease Control and Prevention (CDC), autism spectrum disorders (ASDs) affect at least 1 in every 88 children in the United States. This information is directly provided, indicating a clear prevalence rate of ASDs among children in the U.S. as estimated by the CDC. \\
& \textbf{\(\mathbf{c_p^{(c)}}\):} Hasta 1 de cada 44 niños (o 23 cada 1000) en Estados Unidos tienen TEA. Los varones tienen cuatro veces más probabilidades de tener autismo que las mujeres; se da en 36.5 de cada 1000 varones y 8.8 de cada 1000 mujeres. Los síntomas del TEA pueden aparecer tan pronto como a los 12-18 meses de edad. \\ 
& \textbf{\(\mathbf{a_p^{(c)}}\):} No, the passage does not mention that autism spectrum disorders (ASDs) affect at least 1 in every 88 children according to the Centers for Disease Control and Prevention (CDC). Instead, it states that up to 1 in every 44 children in the United States have ASDs, but it does not provide a comparison to the 1 in 88 statistic mentioned in the question. \\  
& \textbf{Discrepancy:} \abr{not\_enough\_info} ({\it The original \(a_p^{(c)}\) was wrong: while the comparison chunk does contain information about the prevalence of ASDs in the United States, the information provided is not a statistic from the CDC, which is what the question asks for.})\\  
\midrule

\hypertarget{T_11_5100}{\textbf{T-11-5100}}
& \textbf{\(\mathbf{q_p^{(a)}}\):} Can HIV be transmitted to a child through breastfeeding? \\  
& \textbf{\(\mathbf{c_p^{(a)}}\):} HIV can be transmitted to the fetus or newborn during pregnancy, labor, or breastfeeding. \\
& \textbf{\(\mathbf{a_p^{(a)}}\):} Yes, HIV can be transmitted to a child through breastfeeding. \\
& \textbf{\(\mathbf{c_p^{(c)}}\):} El VIH puede transmitirse de una madre a su hijo durante el embarazo, el parto o la lactancia. Sin embargo, esa transmisión vertical puede prevenirse con intervenciones eficaces, como el TAR de la madre y un breve tratamiento del hijo con antirretrovíricos. Otras intervenciones eficaces incluyen medidas para evitar que las embarazadas contraigan el VIH y prácticas apropiadas de lactancia materna. \\
& \textbf{\(\mathbf{a_p^{(c)}}\):} Yes, HIV can be transmitted to a child through breastfeeding, but this transmission can be prevented with effective interventions such as antiretroviral treatment for the mother and brief treatment of the baby. Appropriate lactation practices are also important. \\  
& \textbf{Discrepancy:} \abr{no\_discrepancy} ({\it Initially labelled as \abr{cultural\_discrepancy}, but both answers acknowledge that HIV can be transmitted through breastfeeding but also mention preventive measures that can reduce the risk.}\\  
\midrule

\hypertarget{T_11_12176}{\textbf{T-11-12176}}
& \textbf{\(\mathbf{q_p^{(a)}}\):} Does it usually take more than 6 weeks for symptoms of scabies to appear in a child after contact with an infected person? \\
& \textbf{\(\mathbf{c_p^{(a)}}\):} It may take between 4 to 6 weeks for a child to develop symptoms of scabies after coming in contact with an infected person.\\
& \textbf{\(\mathbf{a_p^{(a)}}\):} No, it usually takes between 4 to 6 weeks for a child to develop symptoms of scabies after coming in contact with an infected person. \\
& \textbf{\(\mathbf{c_p^{(c)}}\):} El periodo de incubación suele durar entre 3 y 14 días. \\ 
& \textbf{\(\mathbf{a_p^{(c)}}\):} I cannot answer the question given the context. \\  
& \textbf{Discrepancy:} \abr{not\_enough\_info} ({\it The original \(a_p^{(c)}\) was wrong. Decontextualization did not work properly: the passages talks about West Nile Virus, while the questions asks about scabies.}) \\  
\midrule

\hypertarget{T_11_7831}{\textbf{T-11-7831}}
& \textbf{\(\mathbf{q_p^{(a)}}\):} Does capsaicin increase the risk of miscarriage? \\
& \textbf{\(\mathbf{c_p^{(a)}}\):} Miscarriage is common and can occur in any pregnancy for many different reasons. Studies have not been done to see if capsaicin increases the chance for miscarriage.\\
& \textbf{\(\mathbf{a_p^{(a)}}\):} No, studies have not been done to determine if capsaicin increases the risk of miscarriage. \\
& \textbf{\(\mathbf{c_p^{(c)}}\):} Un estudio en animales sugirió que la capsaicina podría afectar el crecimiento del embarazo en desarrollo. \\ 
& \textbf{\(\mathbf{a_p^{(c)}}\):} Yes, an animal study suggests that capsaicin may affect the growth of a developing pregnancy, which could potentially increase the risk of miscarriage. However, more research is needed to confirm these findings in humans. \\  
& \textbf{Discrepancy:} \abr{not\_enough\_info} ({\it \(c_p^{(a)}\) states that studies have not been done, while \(c_p^{(c)}\) states that studies in animals suggest that it could affect the pregnancy; yet neither of them confirm not refute the claim stated in the question.}) \\  
\midrule

\hypertarget{T_11_1655}{\textbf{T-11-1655}}
& \textbf{\(\mathbf{q_p^{(a)}}\):} Are medications available that can cure hepatitis C? \\
& \textbf{\(\mathbf{c_p^{(a)}}\):} Medications that cure hepatitis C offer the hope of eliminating the disease in the U.S., yet, today’s report suggests that millions are infected and have not benefited from these new treatment options. Expanded testing, treatment, and prevention services are urgently needed, especially in light of the surge in new infections linked to the opioid crisis. \\
& \textbf{\(\mathbf{a_p^{(a)}}\):} Yes, medications are available that can cure hepatitis C. The passage mentions that medications offering the hope of eliminating the disease are available, indicating the existence of curative treatment options for hepatitis C. \\
& \textbf{\(\mathbf{c_p^{(c)}}\):} ¿Cuál es el tratamiento de la hepatitis C: Los medicamentos antivirales, como el interferón utilizado solo o en combinación con ribavirina, están aprobados para el tratamiento. Algunas infecciones responden mejor al tratamiento que otras. En general, el tratamiento funciona bien en el 30 a 50\% de las personas que completan el tratamiento antiviral. \\ 
& \textbf{\(\mathbf{a_p^{(c)}}\):} No, medications that can cure hepatitis C are not mentioned in the available information. The treatment options discussed involve antiviral medications like interferón and ribavirina, which have a success rate of 30 to 50\% in treating the infection, but there is no mention of a cure. \\  
& \textbf{Discrepancy:} \abr{contradiction}\\  
\midrule

\hypertarget{T_11_6088}{\textbf{T-11-6088}} 
& \textbf{\(\mathbf{q_p^{(a)}}\):} Is a terrorist attack with smallpox considered possible in the current world? \\
& \textbf{\(\mathbf{c_p^{(a)}}\):} The president's decision to recommend that people on smallpox response teams voluntarily receive this vaccine as soon as possible was really based on the need to have a group of responders available very quickly to be able to assist the nation if needed. The urgency and the need for efficient action comes from the fact that we are in fact in a dangerous world, where a terrorist attack with smallpox is possible. We have to be prepared so that we can protect the American people. \\
& \textbf{\(\mathbf{a_p^{(a)}}\):} Yes, a terrorist attack with smallpox is considered possible in the current world. This is stated as a reason for the urgency and need for efficient action to have responders available who can assist if such an event occurs, emphasizing the importance of being prepared to protect the public. \\
& \textbf{\(\mathbf{c_p^{(c)}}\):} A pesar de que se ha hablado mucho sobre la posibilidad de que los terroristas utilicen el virus de la viruela como arma biológica, lo cierto es que esta posibilidad es improbable que ocurra por diversos motivos. En primer lugar, los terroristas deberían acceder a muestras de virus de la viruela y los pocos laboratorios de investigación que disponen de ellas cuentan con medidas de seguridad para protegerlas. Asimismo, sería sumamente difícil que a un grupo le diera tiempo a producir una cantidad lo bastante elevada de virus de la viruela sin ser detectado. \\ 
& \textbf{\(\mathbf{a_p^{(c)}}\):} No, a terrorist attack with smallpox is considered unlikely in the current world. This is because accessing the virus would be difficult due to security measures at research laboratories, and producing a sufficient amount of the virus without being detected would be extremely challenging. \\  
& \textbf{Discrepancy:} \abr{contradiction}\\  
\midrule

\hypertarget{T_15_14251}{\textbf{T-15-14251}} & 
\textbf{\(\mathbf{q_p^{(a)}}\):} Is it recommended for most infants and children to undergo routine hearing tests? \\  
& \textbf{\(\mathbf{c_p^{(a)}}\):} Why does my child need a hearing test: Routine hearing tests are recommended for most babies and children. Newborns are usually given hearing tests before they leave the hospital. If your baby doesn't pass this hearing test, it doesn't always mean serious hearing loss. But your baby should be retested within three months. \\  
& \textbf{\(\mathbf{a_p^{(a)}}\):} Yes, routine hearing tests are recommended for most babies and children. Newborns typically receive a hearing test before leaving the hospital, and if they don't pass, they should be retested within three months. This is important because normal hearing is crucial for language development in early childhood. \\  
& \textbf{\(\mathbf{c_p^{(c)}}\):} Los niños con una audición normal, deben seguir siendo sometidos a pruebas auditivas en sus revisiones regulares. Se suelen volver a hacer pruebas de cribado auditivo cuando los niños tienen 4, 5, 6, 8 y 10 años de edad, y luego en la preadolescencia y la adolescencia. El médico de su hijo le evaluará la audición en cualquier otro momento en que le preocupe. \\  
& \textbf{\(\mathbf{a_p^{(c)}}\):} Yes, it is recommended for most infants and children to undergo routine hearing tests during their regular check-ups. These tests are usually repeated at ages 4, 5, 6, 8, 10, and during adolescence. A doctor may also evaluate a child's hearing if there are any concerns. \\  
& \textbf{Discrepancy:} \abr{cultural\_discrepancy} \\  
\midrule

\hypertarget{T_15_19913}{\textbf{T-15-19913}}
& \textbf{\(\mathbf{q_p^{(a)}}\):} Should women continue taking the full dose of zinc during pregnancy? \\
& \textbf{\(\mathbf{c_p^{(a)}}\):} In most cases, doctors recommend that women continue to take the full dose of zinc during pregnancy. Experts recommend that women with Wilson disease do not breastfeed if they are taking chelating agents. Penicillamine is present in breast milk and can be harmful to a baby. Experts have little information about the safety of trientine and zinc in breast milk.\\
& \textbf{\(\mathbf{a_p^{(a)}}\):} Yes, women should continue taking the full dose of zinc during pregnancy, according to the passage. However, it is important to consult with a doctor for personalized advice and guidance.\\
& \textbf{\(\mathbf{c_p^{(c)}}\):} Vitamina C y zinc: Cuando tiene un resfriado, o cree que se avecina, podría tener la tentación de buscar vitamina C y zinc. Esto no se recomienda durante el embarazo. Primero, no hay evidencia suficiente de que la vitamina C o el zinc sean efectivos para prevenir o tratar los resfriados. En segundo lugar, las dosis de vitamina C y zinc en los suplementos para los resfriados son mucho más altas que la dosis recomendada para las personas embarazadas. La dosis de vitamina C recomendada para mujeres embarazadas es de 80 mg para adolescentes y 85 mg por día para adultos. La dosis recomendada de zinc es 12 mg para adolescentes embarazadas y 11 mg por día para adultas embarazadas. Si está tomando un prenatal, es muy probable que contenga la vitamina C y el zinc que necesita para el día. \\ 
& \textbf{\(\mathbf{a_p^{(c)}}\):} No, women should not take the full dose of zinc during pregnancy as it can exceed the recommended amount. The passage advises against taking high doses of zinc in supplements for colds, which are much higher than the 12 mg recommended for adolescent pregnant women and 11 mg for adult pregnant women. However, more research is needed to confirm these findings in humans. \\  
& \textbf{Discrepancy:} \abr{contradiction} \\  
\midrule

\hypertarget{T_15_470}{\textbf{T-15-470}} & 
\textbf{\(\mathbf{q_p^{(a)}}\):} Is it safe for a breastfeeding mother to consume alcohol every day? \\  
& \textbf{\(\mathbf{c_p^{(a)}}\):} Is it safe for mothers to breastfeed their infant if they have consumed alcohol? Not drinking alcohol is the safest option for breastfeeding mothers. Generally, moderate alcohol consumption by a breastfeeding mother (up to 1 standard drink per day) is not known to be harmful to the infant, especially if the mother waits at least 2 hours after a single drink before nursing. However, exposure to alcohol above moderate levels through breast milk could be damaging to an infant’s development, growth, and sleep patterns. Alcohol consumption above moderate levels may also impair a mother’s judgment and ability to safely care for her child. \\  
& \textbf{\(\mathbf{a_p^{(a)}}\):} Yes, moderate alcohol consumption by a breastfeeding mother (up to 1 standard drink per day) is generally not harmful to the infant, as long as she waits at least 2 hours after a single drink before nursing. \\  
& \textbf{\(\mathbf{c_p^{(c)}}\):} Estoy amamantando. ¿Puedo consumir bebidas alcohólicas? La lactancia y las bebidas alcohólicas no son una buena combinación. No hay ningún nivel de alcohol en la leche materna que se considere seguro para un bebé. \\  
& \textbf{\(\mathbf{a_p^{(c)}}\):} No, there is no level of alcohol in breast milk that is considered safe for a baby. \\  
& \textbf{Discrepancy:} \abr{contradiction} \\  
\midrule

\hypertarget{T_15_1636}{\textbf{T-15-1636}}
& \textbf{\(\mathbf{q_p^{(a)}}\):} Do most children continue taking an afternoon nap until about age 3? \\
& \textbf{\(\mathbf{c_p^{(a)}}\):} As babies get older, nap times typically become more predictable. For example:  
Ages 4 months to 1 year. After the newborn period, your baby will likely nap at least twice a day — once in the morning and once in the early afternoon. Some babies also need a late-afternoon nap. You might aim to have your baby nap at 9 a.m. and 1 p.m. Let your baby nap for as long as he or she wants, unless your baby has difficulty falling asleep at night. If your baby is taking a third nap in the late afternoon, try to eliminate that nap around age 9 months. This will help your baby be ready for an earlier bedtime. \\
& \textbf{\(\mathbf{a_p^{(a)}}\):} Yes, most children continue taking an afternoon nap until about age 3. \\
& \textbf{\(\mathbf{c_p^{(c)}}\):} Sueño. Los niños de la etapa preescolar duermen unas 10-13 horas por la noche. Muchos niños de 4 años dejan de hacer la siesta por la tarde, pero asegúrese de que su hijo dispone de un tiempo de descanso para relajarse y estar tranquilo durante el día. \\ 
& \textbf{\(\mathbf{a_p^{(c)}}\):} Yes, most children stop taking an afternoon nap around the age of 4. \\  
& \textbf{Discrepancy:} \abr{no\_discrepancy} ({\it Originally labeled as a \cd{}, this is not a discrepancy: while \(a_p^{(a)}\) emphasizes continuation until 3 and \(a_p^{(c)}\) cessation around 4, both can be true: children typically nap through the toddler years and drop naps sometime between ages 3–4.}) \\  
\midrule

\hypertarget{T_24_15460}{\textbf{T-24-15460}} & 
\textbf{\(\mathbf{q_p^{(a)}}\):} Is it recommended to limit screen time for 4-year-olds to a certain number of hours per day? \\  
& \textbf{\(\mathbf{c_p^{(a)}}\):} As the parent of a 4-year-old, you should: - Encourage and provide space for physical activity. - Show your child how to participate in and follow the rules of sporting activities. - Encourage play and sharing with other children. - Encourage creative play. - Teach your child to do small chores, such as setting the table. - Read together. - Limit screen time (television and other media) to 2 hours a day of quality programs. - Expose your child to different stimuli by visiting local areas of interest. \\  
& \textbf{\(\mathbf{a_p^{(a)}}\):} Yes, it is recommended to limit screen time for 4-year-olds to 2 hours a day of quality programs. \\  
& \textbf{\(\mathbf{c_p^{(c)}}\):} Los pediatras sugieren limitar el tiempo que los niños usan los medios digitales fuera del horario de las tareas escolares. Entre los 2 y los 5 años, no permita que los niños pasen más de 1 hora frente a la pantalla. Para los niños de 6 años en adelante, establezca límites consistentes en el tiempo de los medios. Eso le ayudará a asegurarse de que tengan suficiente tiempo para la actividad física, el juego, el sueño y otros comportamientos saludables. \\  
& \textbf{\(\mathbf{a_p^{(c)}}\):} Yes, it is recommended to limit screen time for children between 2 and 5 years old to no more than 1 hour per day.\\  
& \textbf{Discrepancy:} \abr{cultural\_discrepancy} \\  
\midrule

\hypertarget{T_24_849}{\textbf{T-24-849}}
& \textbf{\(\mathbf{q_p^{(a)}}\):} Can you help a small child reach the sink for handwashing by holding them up? \\
& \textbf{\(\mathbf{c_p^{(a)}}\):} Can’t reach the sink: If your child is small, hold them up to the sink. If your child can stand, use a safety step to boost them up to the faucet.\\
& \textbf{\(\mathbf{a_p^{(a)}}\):} Yes, you can help a small child reach the sink for handwashing by holding them up if the child is too small to reach the sink on their own. \\
& \textbf{\(\mathbf{c_p^{(c)}}\):} Esté preparado antes de la práctica: Antes de practicar el lavado de manos, asegúrese de tener todos los materiales que necesitará. Esto puede incluir jabón para manos, un taburete (si es necesario), una toalla de mano y un horario visual. Coloque estos artículos en el mismo lugar cada vez que practique para que su hijo pueda concentrarse en aprender cada paso en lugar de tratar de encontrarlo todo. \\ 
& \textbf{\(\mathbf{a_p^{(c)}}\):} Yes, you can help a small child reach the sink for handwashing by holding them up using a taburete (stool) if needed. This is to ensure they have access to the materials for handwashing and can focus on learning the process rather than searching for items. \\  
& \textbf{Discrepancy:} \abr{cultural\_discrepancy} \\  
\midrule

\hypertarget{T_24_9253}{\textbf{T-24-9253}}
& \textbf{\(\mathbf{q_p^{(a)}}\):} Can puberty start outside the typical age range for girls (7–13) and boys (9–15)? \\
& \textbf{\(\mathbf{c_p^{(a)}}\):} Puberty usually starts some time between age 7 and 13 in girls and 9 and 15 in guys. Some people start puberty a bit earlier or later, though. Each person is a little different, so everyone starts and goes through puberty on his or her body's own schedule. This is one of the reasons why some of your friends might still look like kids, whereas others look more like adults.\\
& \textbf{\(\mathbf{a_p^{(a)}}\):} Yes, people can start puberty outside the given age ranges for girls (7 to 13) and boys (9 to 15). Some individuals may start earlier or later than these typical age ranges, as everyone's body develops at its own pace.\\
& \textbf{\(\mathbf{c_p^{(c)}}\):} Información importante sobre la pubertad precoz independiente de la gonadotropina:
\noindent\textbullet\ La pubertad que se presenta de forma anticipada se llama pubertad precoz. Significa que los signos físicos de madurez sexual aparecen demasiado pronto en un niño. La pubertad precoz comienza antes de los 8 años en las mujeres y antes de los 9 años en los varones.

\noindent\textbullet\ Los signos pueden incluir el crecimiento de las mamas y del vello púbico, y los cambios en la voz.

\noindent\textbullet\ La pubertad precoz independiente de la gonadotropina ocurre a causa de los niveles altos de las hormonas sexuales. Entre ellas, se incluyen los andrógenos masculinos y los estrógenos femeninos. Las glándulas sexuales (los ovarios y los testículos) no maduran antes de tiempo.

\noindent\textbullet\ Puede deberse a tumores o masas en los ovarios, los testículos u otras partes del cuerpo. También, al contacto con cremas o pomadas con estrógenos o testosterona.

\noindent\textbullet\ El objetivo del tratamiento es detener la aparición anticipada de los signos de pubertad precoz. En algunos casos, es posible revertir los signos. El tratamiento puede incluir medicamentos que inhiben la acción de las hormonas sexuales.

\noindent\textbullet\ Si un tumor es la causa del problema, tal vez sea necesario hacer una cirugía para extirparlo.

\noindent\textbullet\ Sin tratamiento, un niño no alcanzará su estatura máxima prevista o sufrirá problemas sociales o emocionales por pasar por la pubertad antes que los demás niños de su misma edad.
\\ 
& \textbf{\(\mathbf{a_p^{(c)}}\):} No, puberty that begins before age 8 in girls and before age 9 in boys is considered precocious puberty. \\  
& \textbf{Discrepancy:} \abr{contradiction} \\

\end{longtable}
\endgroup


\renewcommand{\arraystretch}{1.3} 
\setlength{\tabcolsep}{2pt}

\renewcommand{\arraystretch}{1.2}

\begingroup %
\footnotesize
\centering

\begin{longtable}{p{2cm} p{13cm}}
\caption{\texttt{WIKI-EN-DE} examples.} \label{tab:ende_mind_examples} \\

\toprule
\textbf{ID} & \textbf{Content} \\ 
\midrule
\endfirsthead

\multicolumn{2}{c}{\textit{(Continued from previous page)}} \\
\toprule
\textbf{ID} & \textbf{Content} \\ 
\midrule
\endhead

\bottomrule
\multicolumn{2}{r}{\textit{(Continued on next page)}} \\
\endfoot

\bottomrule
\endlastfoot

\hypertarget{E_T_3_854}{\textbf{E-T-3-854}} & 
\textbf{\(\mathbf{q_p^{(a)}}\):} Does Freemasonry require its members to believe in a supreme being? \\  
& \textbf{\(\mathbf{c_p^{(a)}}\):} The religious requirement, initially revolutionary in the early 1700s for being strictly for Christians but welcoming all denominations during an era of sectarian strife, has gradually expanded. It evolved first to include all religions with a revealed scripture and the resurrection of the body after life, then to accept belief in any supreme being, and increasingly allows more personalised interpretations of spirituality whilst maintaining the requirement for some form of metaphysical belief. \\  
& \textbf{\(\mathbf{a_p^{(a)}}\):} Yes, Freemasonry requires its members to believe in a supreme being. The requirement has evolved over time but still maintains that members must have some form of metaphysical belief, including the acceptance of a supreme being. \\  
& \textbf{\(\mathbf{c_p^{(c)}}\):} Die Liberale Freimaurerei setzt aus Gründen der absoluten Gewissensfreiheit ausdrücklich keinen Glauben an ein Höchstes Wesen voraus. Dieses neue Konzept entstand auf dem Konvent des Grand Orient de France des Jahres 1877, bei dem auf Antrag des calvinistischen Pastors Frédéric Desmons das Symbol des Allmächtigen Baumeisters aller Welten in den Ritualen des Großorients abgeschafft wurde. Desmons argumentierte, dass die Freimaurerei wissenschaftlich und rational sei und daher keiner religiösen Bezüge bedürfe. In dieser Konsequenz ersetzte der GOdF die Heilige Schrift als das ``Buch des heiligen Gesetzes'' durch ein symbolisches ``weißes Buch''. Dies ging der Vereinigten Großloge von England zu weit und widersprach ihrer Vorstellung von Freimaurerei: Es kam zu einem Abbruch des Kontakts, 1913 beendete man die Beziehungen zum GOdF und bezeichnete ihn nicht länger als regulär. \\  
& \textbf{\(\mathbf{a_p^{(c)}}\):} No, Freemasonry does not universally require its members to believe in a supreme being, as there are forms of Freemasonry, like the Liberal Freemasonry mentioned, which do not require such a belief due to emphasizing absolute freedom of conscience. \\  
& \textbf{Discrepancy:} \abr{cultural\_discrepancy} \\   
\midrule

\hypertarget{E_T_3_6276}{\textbf{E-T-3-6276}} & 
\textbf{\(\mathbf{q_p^{(a)}}\)}: Was ``The Preservation of St Paul after a Shipwreck at Malta'' painted by Benjamin West?  \\  
& \textbf{\(\mathbf{c_p^{(a)}}\):} West also painted religious subjects, as in his huge work The Preservation of St Paul after a Shipwreck at Malta, at the Chapel of St Peter and St Paul at the Old Royal Naval College in Greenwich, and Christ Healing the Sick, presented to the National Gallery.  \\  
& \textbf{\(\mathbf{a_p^{(a)}}\):} Yes, ``The Preservation of St Paul after a Shipwreck at Malta'' was painted by Benjamin West. The information is directly stated, indicating that West created this work among his other religious subjects.  \\  
& \textbf{\(\mathbf{c_p^{(c)}}\):} In der Hofkapelle zu Windsor sind mehrere Altarbilder von ihm. Zu den Fenstergemälden fertigte er die Kartons, welche Forest von 1792 bis 1796 ausführte. Für die Audienzzimmer des Schlosses schuf er im Auftrag König George III. sechs große Gemälde, deren Gegenstände der Geschichte des Königs Eduard III. entnommen sind, jetzt samt vielen anderen Bildern Wests in einem Saal zu Hampton Court vereinigt. Ein Altarbild in der Hospitalkirche zu Greenwich von West stellt St. Paulus auf Melite vor, wie er die Viper von sich schleudert.  \\  
& \textbf{\(\mathbf{a_p^{(c)}}\):} No, the passage does not mention a painting called ``The Preservation of St Paul after a Shipwreck at Malta'' by Benjamin West, but it does describe an altarpiece depicting St. Paul on Melite, which could be related to the same biblical event. However, the title and exact subject matter do not match the question. \\  
& \textbf{Discrepancy:} \abr{contradiction} \\  
\midrule

\hypertarget{E_T_3_6847}{\textbf{E-T-3-6847}} &
\textbf{\(\mathbf{q_p^{(a)}}\)}: Have the United States of America ever entered into any war or act of hostility against any Muslim nation?  \\  
& \textbf{\(\mathbf{c_p^{(a)}}\):} As the Government of the United States of America is not, in any sense, founded on the Christian religion; as it has in itself no character of enmity against the laws, religion, or tranquility, of Mussulmen [Muslims]; and as the said States never entered into any war or act of hostility against any Mahometan [Mohammedan] nation, it is declared by the parties that no pretext arising from religious opinions shall ever produce an interruption of the harmony existing between the two countries.  \\  
& \textbf{\(\mathbf{a_p^{(a)}}\):} No, according to the statement, the United States of America has never entered into any war or act of hostility against any Muslim nation, as it declares that no pretext arising from religious opinions shall ever produce an interruption of the harmony existing between the two countries.  \\  
& \textbf{\(\mathbf{c_p^{(c)}}\):} Berthold Seewald: Ihren ersten Krieg führten die USA gegen muslimische Piraten In: Die Welt veröffentlicht am 17. Februar 2019  \\  
& \textbf{\(\mathbf{a_p^{(c)}}\):} I cannot answer given the context. \\  
& \textbf{Discrepancy:} \abr{not\_enough\_info} ({\it \(a_p^{(a)}\) claims there has been no such hostility based on a statement emphasizing harmony, while  \(a_p^{(c)}\) does not pertain information usable to answer the question since it states a real historical circumstance in the USA's war against Muslim pirates but does not contain relevant information about any hostility against a Muslim nation.}) \\  
\newpage

\hypertarget{E_T\_3\_8155}{\textbf{E-3-8155}} & 
\textbf{\(\mathbf{q_p^{(a)}}\)}: Is the Eucharist believed to be the same sacrifice that Jesus made on the cross?  \\  
& \textbf{\(\mathbf{c_p^{(a)}}\):} In the Eucharist the same sacrifice that Jesus made only once on the cross is believed to be made present at every Mass. According to Compendium of the Catechism of the Catholic Church, ``The Eucharist is the very sacrifice of the Body and Blood of the Lord Jesus which he instituted to perpetuate the sacrifice of the cross throughout the ages until his return in glory.''  \\  
& \textbf{\(\mathbf{a_p^{(a)}}\):} Yes, the Eucharist is believed to be the same sacrifice that Jesus made on the cross. It is thought to make this sacrifice present at every Mass, perpetuating the sacrifice of the cross throughout the ages until his return in glory. \\  
& \textbf{\(\mathbf{c_p^{(c)}}\):} Die aus der anglikanischen und arminianischen Tradition kommende Evangelisch-methodistische Kirche geht von der wirklichen, persönlichen und lebendigen Gegenwart Jesu Christi im Abendmahl aus, ohne diese vollständig erklären zu wollen. Die Lehre von der Transsubstantiation wird abgelehnt. Das Abendmahl ist Sakrament, Eucharistie (Dank), Gemeinschaft der Kirche, Erinnerung und Gnadenmittel. Es ist auch ein Opfer, allerdings nicht als Wiederholung des Opfers Christi, sondern als ``Wieder-Darstellung'', wobei sich die Kirche in Einheit mit Christus als Opfer gibt (Röm 12,1 ). \\  
& \textbf{\(\mathbf{a_p^{(c)}}\):} No, the Eucharist is not believed to be the same sacrifice that Jesus made on the cross, but rather a ``Wieder-Darstellung'' or re-presentation of it, where the church gives itself as an offering in unity with Christ. \\  
& \textbf{Discrepancy:} \abr{cultural\_discrepancy}\\  
\midrule

\hypertarget{E_T\_5\_928}{\textbf{E-T-5-928}} &
\textbf{\(\mathbf{q_p^{(a)}}\)}: Is graphic design used in the integration of elements on a map page?  \\  
& \textbf{\(\mathbf{c_p^{(a)}}\):} Layout: The map image must be placed on the page (whether paper, web, or other media), along with related elements, such as the title, legend, additional maps, text, images, and so on. Each of these elements have their own design considerations, as does their integration, which largely follows the principles of graphic design. \\  
& \textbf{\(\mathbf{a_p^{(a)}}\):} Yes, graphic design is used in the integration of elements on a map page. The placement and arrangement of various components such as the title, legend, maps, text, and images on the page follow the principles of graphic design to effectively communicate spatial information. \\  
& \textbf{\(\mathbf{c_p^{(c)}}\):} Lesbarkeit und Verständlichkeit einer Karte treffen zu, wenn der Kartenbenutzer sich beim Lesen der Karte schnell und leicht ein zutreffendes Bild von der dargestellten Wirklichkeit machen kann. Dies lässt sich an verschiedenen Merkmalen festmachen. Grafisch gut gestaltete Kartenzeichen mit hoher Symbolkraft und sinnvoller Farbgebung erleichtern das Verständnis („selbsterklärende“ Karte oder naturrealistische Karte). Eine gute Generalisierung trägt dazu bei, dass die wesentlichen und typischen Objekte oder Sachverhalte wiedergegeben werden und die Karte weder zu leer, noch zu überladen wirkt. (Dies darf nicht damit verwechselt werden, dass eine Karte auf Grund unterschiedlicher darzustellender Gegebenheiten – z. B. dicht besiedelte Gebiete und große Landwirtschaftsflächen – naturgemäß unterschiedlich dicht wirkt.) \\  
& \textbf{\(\mathbf{a_p^{(c)}}\):} Yes, graphic design is used in the integration of elements on a map page. This can be inferred from the mention of ``Grafisch gut gestaltete Kartenzeichen'' which translates to ``graphically well-designed map signs'', indicating that graphic design plays a role in creating understandable and effective maps. \\  
& \textbf{Discrepancy:} \abr{no\_discrepancy} \\

\end{longtable}
\endgroup

\begin{prompt}[title={\thetcbcounter: Topic Labeling}, label=prompt:labeling]
\begin{lstlisting}
You are given a set of keywords and five passages, each centered around a common topic. Your task is to analyze both the keywords and the content of the documents to create a clear, concise label that accurately reflects the overall theme they share. The label:  
- Must not be or include the word "LABEL".
- Must match the language of the keywords.
- Should be broad yet relevant, capturing the overall theme in a general way rather than focusing on specific components or processes

Your response should **only** be the label (no additional text).  

Keywords: {keywords}
Documents: {docs}
Label:
\end{lstlisting}
\end{prompt}

\begin{prompt}[title={\thetcbcounter: Question Generation}, label=prompt:question_generation]
\begin{lstlisting}
You will be given a PASSAGE and an excerpt from the FULL_DOCUMENT where it appears. Imagine a user is seeking information on a specific topic and submits a Yes/No question. Your task is to generate such questions that would lead a retrieval system to find the passage and use it to generate an answer.

#### TASK BREAKDOWN ####
1. Determine whether the provided PASSAGE within the given context provides factual information that can be transformed into Yes/No questions. Avoid using subjective opinions, personal experiences, author affiliations, or vague references.
2. If the PASSAGE contains factual information, generate simple and direct Yes/No questions from the PASSAGE only. Do not use the FULL_DOCUMENT to generate questions. Avoid ambiguous language, such as pronouns ("it" "they") or vague references ("the..."), and always define acronyms and abbreviations within each question (e.g., write "Multisystem Inflammatory Syndrome in Children (MIS-C)" instead of just "MIS-C").
3. Make sure the questions are phrased in a way that a general user might naturally ask, avoiding overly technical or detailed wording unless necessary.
4. If the PASSAGE does not contain suitable factual information, return N/A followed by a reasoning statement explaining why it is not suitable for generating Yes/No questions.

#### EXAMPLES ####

PASSAGE: Risk factors: Children diagnosed with MIS-C are often between the ages of 5 and 11 years old. But cases are reported among children ages 1 to 15. A few cases have also happened in older kids and in babies.

FULL_DOCUMENT: Overview Multisystem inflammatory syndrome in children (MIS-C) is a group of symptoms linked to swollen, called inflamed, organs or tissues. People with MIS-C need care in the hospital [...]

QUESTIONS: Can children with Multisystem Inflammatory Syndrome in Children (MIS-C) be as young as 1 year old?
Are most cases of Multisystem Inflammatory Syndrome in Children (MIS-C) found in children between 5 and 11 years old?
Have there been cases of Multisystem Inflammatory Syndrome in Children (MIS-C) in babies?


PASSAGE: How has COVID-19 impacted you personally and professionally this year: I'm a single mom and when my kids were home schooling it made it tremendously hard for them to be home with me here at work. That was a big challenge. I think it's difficult for all of us health care providers, who are taking care of the sickest patients and working with stressed out families. It adds an additional challenge for us.

FULL_DOCUMENT: Stacey Stone, M.D., began walking the halls of All Children's Hospital well before she wore a doctor's white coat. Touring the neonatal intensive care unit (NICU) as a teenager, she became smitten with the idea [...]

QUESTIONS: N/A, the passage provides subjective information about the personal and professional impact of COVID-19 on the author, which is not suitable for generating Yes/No questions.

#### YOUR TASK ####
PASSAGE: {passage}
FULL_DOCUMENT: {full_document}
QUESTIONS:

\end{lstlisting}
\end{prompt}

\begin{prompt}[title={\thetcbcounter: Search Queries Generation}, label=prompt:query_generation]
\begin{lstlisting}
You will receive a PASSAGE and a QUESTION based on the passage. Your task is to generate a concise search queries that effectively capture the user's intent to retrieve relevant information from a different database or search engine to look for contradictory information to the given passage.

#### TASK BREAKDOWN ####
1. Focus on the main concepts and avoid extraneous details.
2. Expand acronyms and abbreviations to their full forms (e.g., write Multisystem Inflammatory Syndrome in Children instead of MIS-C).
3. Ensure the query aligns with the **intent** of the user's question while maintaining brevity.
4. If you generate more than one query, separate them with a semicolon.

#### EXAMPLES ####

PASSAGE: Risk factors: Children diagnosed with MIS-C are often between the ages of 5 and 11 years old. But cases are reported among children ages 1 to 15. A few cases have also happened in older kids and in babies.

QUESTION: Can children with Multisystem Inflammatory Syndrome in Children (MIS-C) be as young as 1 year old?

SEARCH_QUERY: "Multisystem Inflammatory Syndrome in Children (MIS-C) age range; youngest reported case of Multisystem Inflammatory Syndrome in Children (MIS-C)"

PASSAGE: Removing the catheter:\\n- In the morning, remove the catheter.\\n- First, take the water out of the balloon. Place a syringe on the colored balloon port and let the water fill the syringe on its own. If water is not draining into the syringe, gently pull back on the syringe stopper. Do not use force.\\n- Once the amount of water inserted the night before is in the syringe, gently pull out the Foley catheter.\\n- Continue the normal cathing scheduled during the day.\\n- Wash the Foley catheter with warm, soapy water. Then, rinse and lay it on a clean towel to dry for later use.

QUESTION: Is it necessary to reuse a Foley catheter after cleaning it?

SEARCH_QUERY: "Foley catheter reuse safety; guidelines for single-use vs. reusable Foley catheters"

#### YOUR TASK ####

PASSAGE: {passage}
QUESTION: {question}
SEARCH_QUERY:
\end{lstlisting}
\end{prompt}

\begin{prompt}[title={\thetcbcounter: Relevant Passages Identification}, label=prompt:relevant_passages_identification]
\begin{lstlisting}
Determine whether the passage contains information that directly answers the question. If it does, return Yes; otherwise, return No. Respond with only Yes or No in uppercase.

#### EXAMPLES ####

PASSAGE: Puede empezar a ofrecerle a su hijo pequeñas cantidades de agua alrededor de los 6 meses. Si vive en un lugar con agua fluorada, esto le ayudará a prevenir la aparición de caries. Pregúntele al proveedor de atención médica de su hijo cuánta agua debe darle.

QUESTION: Are feeding tubes typically employed for infants, and do they have any potential benefits for older children or teenagers in specific cases?

RELEVANT: No


PASSAGE: La alimentación por sonda también puede ser una buena idea para aquellos niños que pueden comer de forma segura, pero no logran comer suficiente por la boca (incluso con suplementos) como para mantener un peso saludable. En estos casos, la alimentación por sonda puede complementar la rutina habitual de las comidas. Estas alimentaciones también son útiles cuando un niño está creciendo rápidamente y necesita más nutrientes mientras desarrolla habilidades para comer o tiene una enfermedad grave. Después, el tubo puede usarse con menos frecuencia o quitarse.

QUESTION: Are feeding tubes typically employed for infants, and do they have any potential benefits for older children or teenagers in specific cases?

RELEVANT: Yes

#### YOUR TASK ####

PASSAGE: {passage}
QUESTION: {question}
RELEVANT:
\end{lstlisting}
\end{prompt}

\begin{prompt}[title={\thetcbcounter: Answer Generation}, label=prompt:answer_generation]
\begin{lstlisting}
You will be given a QUESTION, a PASSAGE, and an excerpt from the FULL_DOCUMENT where the passage appears. If the PASSAGE contains information that directly answers the QUESTION, your task is to provide a Yes/No answer to the QUESTION, followed by a brief explanation based only on the PASSAGE.

#### TASK INSTRUCTIONS ####

- If the passage does not contain enough information to answer the question or the information only contain personal experiences, respond with "I cannot answer the question given the context.".
- Answer as if you were a chatbot responding to a user.
- Do not mention "the passage," "the text," or refer to where the information comes from.
- Keep the response natural, direct, and informative.
- Do not use outside knowledge or the FULL_DOCUMENT to answer the question.

#### EXAMPLES ####

QUESTION: Can children with Multisystem Inflammatory Syndrome in Children (MIS-C) be as young as 1 year old?  

PASSAGE: Risk factors: Children diagnosed with MIS-C are often between the ages of 5 and 11 years old. But cases are reported among children ages 1 to 15. A few cases have also happened in older kids and in babies.  

FULL_DOCUMENT: Overview Multisystem inflammatory syndrome in children (MIS-C) is a group of symptoms linked to swollen, called inflamed, organs or tissues. People with MIS-C need care in the hospital [...]  

ANSWER: Yes, children as young as 1 year old can be affected by MIS-C. Although it is most common in children aged 5 to 11, cases have been reported in younger children, including 1-year-olds and even infants.


QUESTION: Does a baby's Apgar score help identify infants who may have trouble breathing after birth?

PASSAGE: En el año 1963, se acuñó el acrónimo APGAR en inglés para el sistema de puntuación como ayuda nemónica de aprendizaje: Apariencia (color de la piel), Pulso (frecuencia cardíaca), "Grimace" o mueca (irritabilidad del reflejo), Actividad (tono muscular) y Respiración.

FULL_DOCUMENT: Misión Virginia Apgar Ante todo, Virginia Apgar era una incontenible y carismática defensora de los bebés, cuyo ingenio y vivaz personalidad cautivaba a todos los que conocía en su constante pugna por mejorar la salud materno-infantil. La prueba Apgar Virginia Apgar nació el 7 de junio de 1909 en Westfield, Nueva Jersey. Asistió a Mount Holyoke Collage en Massachusetts. En los años 30, estudió medicina en la Facultad de Médicos y Cirujanos de Columbia Universito en Nueva York con la intención de convertirse en cirujana. [...]

ANSWER: I cannot answer the question given the context.

#### YOUR TASK ####

QUESTION: {question}  
PASSAGE: {passage}  
FULL_DOCUMENT: {full_document}  
ANSWER:

Before answering, consider whether the passage contains information that directly answers the question. If it does not, respond with: "I cannot answer given the context."
\end{lstlisting}
\end{prompt}

\begin{prompt}[title={\thetcbcounter: Discrepancy Detection}, label=prompt:discrepancy_detection]
\begin{lstlisting}
You will be given a QUESTION along with two responses (ANSWER_1 and ANSWER_2). Your task is to classify the relationship between the two answers, given the question, into one of the following categories:

1. CULTURAL_DISCREPANCY: The answers reflect differences that stem from cultural norms, values, or societal perspectives rather than factual contradictions. This includes variations in common practices, traditions, or expectations that depend on cultural context. If both statements can be valid in different regions, societies, or traditions, classify them here rather than as CONTRADICTION.

2. CONTRADICTION: The answers provide directly opposing factual information, meaning one explicitly denies what the other asserts. A contradiction occurs only if both statements cannot be true in any context. Differences in reasoning, examples, or perspectives do not count as contradictions unless they fundamentally conflict. If both statements could be true in different settings (e.g., due to geography, culture, or historical variation), classify them as CULTURAL_DISCREPANCY instead.

3. NoT_ENoUGH_INFO: There is insufficient information to determine whether a discrepancy exists. This applies when the answers are too vague, incomplete, require additional context to assess their relationship, or directly fail to answer the question asked.

4. No_DISCREPANCY: The answers are fully consistent, presenting aligned or identical information without any conflict or variation in framing.

Response Format:  
- REASON: [Briefly explain why you selected this category]  
- DISCREPANCY_TYPE: [Choose one of the five categories above]

#### EXAMPLE ####

QUESTION: Does the shot significantly increase the risk of blood clots? 
ANSWER_1: Yes. The shot increases the risk of blood clots, making users three times more likely to experience them compared to those using a hormonal IUD.
ANSWER_2: No. The shot does not contain estrogen and is considered safe for use immediately after childbirth, meaning it does not significantly raise the risk of blood clots.
REASON: The answers provide directly opposing factual information on the risk of blood clots associated with the shot.
DISCREPANCY_TYPE: CONTRADICTION

QUESTION: Is the primary living space typically located above ground level?
ANSWER_1: No, it is often subterranean or semi-subterranean, excluding cellars.
ANSWER_2: Yes, it is typically at ground level.
REASON: The answers reflect differences in cultural practices and norms regarding the location of primary living spaces, rather than factual contradictions.
DISCREPANCY_TYPE: CULTURAL_DISCREPANCY

QUESTION: Does the rectangular fold method involve folding the diaper into a rectangle?	
ANSWER_1: Yes, the rectangular fold method involves folding the diaper into a rectangle. The passage describes this process in detail, mentioning to fold the diaper into a rectangle and potentially making an extra fold for added coverage in certain areas.	
ANSWER_2: No, the triangular fold method involves folding the diaper into a triangle, not a rectangle.	
REASON: The two answers refer to different diaper-folding techniques (rectangular vs. triangular) rather than directly contradicting each other. ANSWER_2 does not dispute the first answer's claim about the rectangular fold but instead describes a separate method.
DISCREPANCY_TYPE: NoT_ENoUGH_INFO

#### YOUR TASK ####  

QUESTION: {question}  
ANSWER_1: {answer_1}  
ANSWER_2: {answer_2}  

Before answering, consider whether the answers could be true in different cultural contexts.
\end{lstlisting}
\end{prompt}

\begin{prompt}[title={\thetcbcounter: FEVER conversion}, label=prompt:fever_conversion]
\begin{lstlisting}

You will receive a CLAIM, EVIDENCE, and a LABEL indicating whether the EVIDENCE REFUTES or SUPPORTS the claim. Your task is to generate a triplet consisting of a Yes/No QUESTION and two corresponding Yes/No ANSWERS.

#### TASK BREAKDOWN ####  
1. Generate a Yes/No QUESTION that directly asks about the CLAIM in a way that the expected answer would naturally follow with "Yes" and the CLAIM.  
2. ANSWER1 should be a reformulation of the CLAIM as a Yes response to the QUESTION.  
3. ANSWER2 depends on the LABEL:  
   - If the LABEL is "REFUTES", ANSWER2 should explicitly contradict ANSWER1 using the EVIDENCE.  
   - If the LABEL is "SUPPORTS", ANSWER2 should reinforce ANSWER1 with relevant information from the EVIDENCE.  
4. Keep the QUESTION and ANSWERS concise, factual, and directly tied to the EVIDENCE, avoiding unnecessary details.

#### EXAMPLES ####  

CLAIM: Tony Blair is not a leader of a UK political party.  
LABEL: REFUTES  
EVIDENCE: Tony Blair was elected Labour Party leader in July 1994, following the sudden death of his predecessor, John Smith.

QUESTION: Is Tony Blair not a leader of a UK political party?  
ANSWER1: Yes, Tony Blair is not a leader of a UK political party.  
ANSWER2: No, Tony Blair was elected Labour Party leader in July 1994.

CLAIM: The industry that The New York Times is part of is declining.  
LABEL: SUPPORTS  
EVIDENCE: The late 2000s-early 2010s global recession, combined with the rapid growth of free web-based alternatives, has helped cause a decline in advertising and circulation, as many papers had to retrench operations to stanch the losses. 

QUESTION: Is the industry that The New York Times is part of declining?
ANSWER1: Yes, the industry that The New York Times is part of is declining.
ANSWER2: Yes, the industry is declining due to the late 2000s-early 2010s global recession and the rise of free web-based alternatives, which led to a drop in advertising and circulation, forcing many papers to cut operations.


#### YOUR TASK ####  

CLAIM: {claim}  
LABEL: {label}  
EVIDENCE: {evidence}
\end{lstlisting}
\end{prompt}

\begin{prompt}[title={\thetcbcounter: DPLACE conversion}, label=prompt:dplace_conversion]
\begin{lstlisting}
You will receive a DEFINITION of a cross-cultural difference, and two EXAMPLES of this difference. Your task is to generate a Yes/No QUESTION that asks about the DEFINITION and two corresponding Yes/No ANSWERS so that the responses imply a cultural discrepancy given the question.

#### TASK BREAKDOWN ####  
1. Generate a Yes/No QUESTION that directly asks about the DEFINITION.
2. ANSWER1 should be a reformulation of EXAMPLE1, and ANSWER2 should be a reformulation of EXAMPLE2, both in the form of a Yes/No ANSWER.
3. Keep the QUESTION and ANSWERS concise, factual, and directly tied to the EVIDENCE, avoiding unnecessary details.

#### EXAMPLES ####  

DEFINITION: Floor level of the prevailing type of dwelling.
EXAMPLE1:Subterranean or semi-subterranean, ignoring cellars beneath the living quarters
EXAMPLE2:Floor formed by or level with the ground itself.

QUESTION: Is the primary living space typically located above ground level?
ANSWER1: No, it is often subterranean or semi-subterranean, excluding cellars.
ANSWER2: Yes, it is typically at ground level.

DEFINITION: Age or occupational specialization in the actual building of a permanent dwelling or the erection of a transportable shelter; not including the acquisition or preliminary preparation of the materials used.	
EXAMPLE1: Junior age specialization, i.e., the activity is largely performed by boys and/or girls before the age of puberty
EXAMPLE2: Senior age specialization, i.e., the activity is largely performed by men and/or women beyond the prime of life

QUESTION: Is the construction of dwellings typically performed by adults past their prime?
ANSWER1: No, it is primarily carried out by boys and girls before puberty.
ANSWER2: Yes, it is mainly done by older adults beyond their prime.


#### YOUR TASK ####  

DEFINITION: {definition}  
EXAMPLE1: {example1}  
EXAMPLE2: {example2}
\end{lstlisting}
\end{prompt}


\renewcommand{\arraystretch}{1.2}
\footnotesize
\begin{longtable}{p{4cm} p{11cm}}
\caption{\texttt{ROSIE} 15-topics model} \label{tab:15_tpcs} \\

\toprule
\textbf{Topic} & \textbf{Description} \\
\midrule
\endfirsthead

\multicolumn{2}{c}{\textbf{Continuation of Table \ref{tab:15_tpcs}}} \\
\toprule
\textbf{Topic} & \textbf{Description} \\
\midrule
\endhead

\midrule
\multicolumn{2}{r}{\textit{Continued on next page}} \\
\midrule
\endfoot

\bottomrule
\endlastfoot

{\bf Anatomy} & bone muscle injury technology nerve surgery joint ear tooth pain exercise spinal foot leg head arm fracture activity knee \\
         & lesi\'on hueso m\'usculo cirug\'ia pie nervio articulaci\'on dolor ejercicio o\'ido diente pierna fractura espinal brazo actividad rodilla cabeza columna \\
{\bf Health Disparities in the United States} & patient health risk study increase care report woman age factor disease result organization associate rate death clinical diagnosis include \\
         & paciente riesgo salud a\~no enfermedad prueba mujer alto diagn\'ostico factor persona atenci\'on caso tasa cl\'inico detecci\'on muerte edad dato \\
{\bf Hormonal Regulation} & cell protein body blood acid gene produce level function technology normal hormone immune gland result thyroid tissue enzyme mutation \\
         & c\'elula prote\'ina cuerpo producir \'acido llamado gen sistema nivel sangre normal hormona funci\'on gl\'andula gl\'obulo tipo tejido causar mutaci\'on \\
{\bf Infectious Diseases} & infection disease hiv treatment virus person people antibiotic infect bacteria organization treat health prevent hepatitis risk tuberculosis spread drug \\
         & infecci\'on persona enfermedad tratamiento vih virus causar bacteria infectado antibi\'otico transmisi\'on sexual riesgo hepatitis prevenir tratar grave tuberculosis caso \\
{\bf Medication} & provider doctor medication medicine care health technology treatment healthcare treat talk medical symptom dose prescribe follow injection drug day \\
         & m\'edico medicamento proveedor atenci\'on tomar tratamiento tratar dosis prueba inyecci\'on examen necesitar s\'intoma ayudar secundario m\'edica salud hora hablar \\
{\bf Nutrition} & food eat technology water weight diet healthy drink alcohol vitamin fat child product body people avoid smoke day milk \\
         & alimento comer agua dieta peso alcohol saludable producto vitamina cantidad ayudar grasa fumar beber mantener comida contener consumo evitar \\
{\bf Pregnancy \& Birth} & baby pregnancy woman birth pregnant technology health risk infant organization week mother increase bear defect breast delivery sex vaginal \\
         & beb\'e embarazo mujer embarazado riesgo parto nacimiento nacer nacido semana problema madre defecto \'utero sexual aumentar materno vaginal quedar \\
{\bf Health Symptoms} & symptom pain sleep feel technology sign severe people day experience include fever headache time asthma mild common hour occur \\
         & s\'intoma dolor persona causar grave sentir problema signo sue\~no cabeza dificultad respiratorio fiebre experimentar asma respirar leve hora dormir \\
{\bf Genetic Disorders} & condition disorder syndrome disease people affect brain gene genetic symptom technology develop child common occur seizure mutation include family \\
         & s\'indrome trastorno enfermedad persona afecci\'on afectar causar condici\'on gen\'etico s\'intoma problema gen cerebro cerebral desarrollar tipo causa com\'un ni\~no \\
{\bf Cardiovascular Health} & blood heart disease pressure kidney technology lung diabetes artery level vessel risk body valve flow condition pulmonary oxygen insulin \\
         & sangre coraz\'on card\'iaco enfermedad diabetes presi\'on sangu\'ineo arterial nivel pulmonar arteria renal vaso riesgo pulm\'on alto cuerpo tipo v\'alvula \\
{\bf Child Development} & child health care technology organization family treatment parent clinic time school team life hospital talk learn people mayo activity \\
         & ni\~no hijo ayudar salud tratamiento atenci\'on problema padre m\'edico cuidado equipo vida necesitar familia adolescente importante hospital persona mayo \\
{\bf Cancer Treatment} & cancer treatment tumor patient therapy cell breast radiation treat risk technology surgery chemotherapy stage study disease transplant spread clinical \\
         & c\'ancer tratamiento tumor paciente c\'elula tipo terapia riesgo cirug\'ia quimioterapia tratar canceroso mama enfermedad trasplante radiaci\'on a\~no cl\'inico pron\'ostico \\
{\bf Medical Procedures} & surgery procedure technology remove image tube tissue fluid bladder doctor urine tomography surgeon intestine ray perform sample magnetic stomach \\
         & cirug\'ia prueba m\'edico peque\~no l\'iquido examen tubo intestino tejido orina vejiga im\'agenes cirujano muestra est\'omago aguja biopsia quir\'urgico cat\'eter \\
{\bf Vaccination Practices} & vaccine dose child age vaccination health organization receive person month influenza recommend virus adult flu risk administer report vaccinate \\
         & vacuna dosis ni\~o a\~no persona vacunaci\'on recibir gripe edad mes virus enfermedad adulto administrar caso riesgo sarampi\'on recomendar unidos \\
{\bf Hygiene \& Skin Care} & skin eye technology hand vision hair mouth contact light rash body color nose clean dry wear red remove wash \\
         & piel ojo visi\'on causar mano ocular color boca \'area peque\~no contacto luz nariz cabello zona persona aire cuerpo aplicar \\

\end{longtable}%


\renewcommand{\arraystretch}{1.2}
\footnotesize
\centering
\begin{longtable}{p{4cm} p{11cm}}
\caption{\texttt{ROSIE} 20-topics model} \label{tab:20_tpcs} \\

\toprule
\textbf{Topic} & \textbf{Description} \\
\midrule
\endfirsthead

\multicolumn{2}{c}{\textbf{Continuation of Table \ref{tab:20_tpcs}}} \\
\toprule
\textbf{Topic} & \textbf{Description} \\
\midrule
\endhead

\midrule
\multicolumn{2}{r}{\textit{Continued on next page}} \\
\midrule
\endfoot

\bottomrule
\endlastfoot

{\bf Medication Adherence} & medication medicine doctor technology treat drug treatment dose injection prescribe day prescription pain time follow opioid reduce talk hour \\
         & medicamento tomar m\'edico tratar tratamiento dosis inyecci\'on secundario hora oral dolor ayudar dejar administrar receta v\'ia reducir indicar funcionar \\
{\bf Genetic Disorders} & cell gene protein mutation acid genetic function change technology deoxyribonucleic chromosome result enzyme normal role form produce lead development \\
         & c\'elula gen prote\'ina mutaci\'on gen\'etico \'acido llamado causar funci\'on producir desoxirribonucleico cambio celular normal cromosoma enzima tipo papel proporcionar \\
{\bf Child Development} & child technology sleep parent time feel health disorder school organization family activity stress anxiety behavior people depression learn talk \\
         & ni\~no hijo ayudar padre problema sue\~no trastorno sentir actividad persona ansiedad comportamiento dormir adolescente depresi\'on vida estr\'es familia escuela \\
{\bf Infectious Diseases} & infection hiv virus person disease health people infect risk organization bacteria transmission antibiotic prevent hepatitis spread contact illness human \\
         & infecci\'on persona enfermedad virus vih transmisi\'on sexual riesgo infectado bacteria causar antibi\'otico hepatitis prevenir contacto caso humano brote exposici\'on \\
{\bf Nutrition} & food eat diet healthy drink weight vitamin technology fat alcohol water day milk avoid body supplement product include allergy \\
         & alimento comer dieta peso saludable vitamina ayudar cantidad grasa beber comida agua alcohol producto leche mantener evitar suplemento tomar \\
{\bf Minimally Invasive Medical Procedures} & surgery procedure image technology remove surgeon tube ray tomography tissue sample ultrasound perform magnetic resonance doctor biopsy insert needle \\
         & cirug\'ia prueba peque\~no cirujano im\'agenes tejido muestra tubo m\'edico colocar examen biopsia aguja l\'iquido cat\'eter radiograf\'ia dispositivo quir\'urgico incisi\'on \\
{\bf Maternal Health \& Childbirth} & baby pregnancy birth woman pregnant technology infant health week risk mother bear defect organization delivery period breast uterus newborn \\
         & beb\'e embarazo mujer embarazado parto nacimiento nacer semana nacido riesgo madre defecto \'utero problema materno quedar mes per\'iodo anticonceptivo \\
{\bf Injury \& Musculoskeletal Health} & bone pain injury muscle technology joint exercise leg foot activity surgery arm fracture knee head arthritis hand hip spine \\
         & lesi\'on hueso dolor pie m\'usculo ejercicio articulaci\'on pierna actividad cirug\'ia brazo fractura dedo rodilla columna artritis \'oseo mano cadera \\
{\bf Neurological Disorders} & syndrome brain condition disorder symptom affect people disease nerve seizure muscle technology severe occur include develop nervous life common \\
         & s\'indrome trastorno persona s\'intoma cerebro enfermedad afecci\'on afectar cerebral causar problema nervioso condici\'on grave convulsi\'on nervio tipo ni\~no sistema \\
{\bf Vaccination Schedule} & vaccine dose child age vaccination health receive organization month person influenza recommend flu adult vaccinate coverage report administer measles \\
         & vacuna ni\~o dosis a\~no vacunaci\'on gripe recibir persona edad mes sarampi\'on administrar adulto caso cobertura recomendar enfermedad virus unidos \\
{\bf Cardiovascular Health} & heart blood pressure lung artery disease technology vessel valve flow pulmonary oxygen stroke chest attack body clot vein cardiac \\
         & coraz\'on card\'iaco presi\'on sangre arterial sangu\'ineo pulmonar pulm\'on arteria enfermedad vaso v\'alvula ox\'igeno flujo ataque respiratorio problema cuerpo co\'agulo \\
{\bf Chronic Conditions} & risk health increase people age disease factor woman organization death adult chronic study rate condition diabetes associate reduce percent \\
         & riesgo persona enfermedad a\~no mujer factor aumentar salud edad alto muerte diabetes adulto hombre cr\'onico tasa fumar probabilidad relacionado \\
{\bf Allergies and Sensory Symptoms} & symptom eye pain ear loss technology sign severe feel fever vision headache include reaction people common throat experience mild \\
         & s\'intoma dolor ojo causar p\'erdida grave visi\'on signo cabeza fiebre o\'ido problema persona sentir reacci\'on dificultad ocular leve hinchaz\'on \\
{\bf Medical Research and Treatment} & patient treatment study clinical result therapy diagnosis trial report day evidence drug disease month organization follow evaluate positive receive \\
         & paciente tratamiento cl\'inico diagn\'ostico prueba caso ensayo enfermedad terapia evaluar mes evidencia dato positivo recibir demostrar alto tasa observar \\
{\bf Pediatric Healthcare Services} & care health treatment child hospital clinic program patient team medical include mayo center pediatric improve organization service treat plan \\
         & tratamiento atenci\'on ni\~no salud m\'edico hospital cuidado programa equipo paciente mayo centro clinic mejorar servicio proporcionar enfermedad ayudar unidad \\
{\bf Endocrine System Dysregulation} & blood level body kidney cell disease liver technology hormone diabetes sugar gland insulin thyroid transplant normal produce glucose people \\
         & sangre nivel cuerpo enfermedad diabetes renal producir hormona tipo ri\~on h\'igado c\'elula alto trasplante sangu\'ineo az\'ucar gl\'obulo gl\'andula insulina \\
{\bf Cancer} & cancer treatment tumor cell therapy breast radiation technology chemotherapy stage surgery treat spread grow tissue lymph risk body node \\
         & c\'ancer tratamiento tumor c\'elula tipo quimioterapia canceroso mama terapia cirug\'ia radiaci\'on tratar radioterapia linf\'atico tejido estadio cuello riesgo ganglio \\
{\bf Digestive System} & technology bladder stomach intestine surgery urine tract urinary bowel cyst abdominal pelvic kidney stool organ fluid colon esophagus common \\
         & intestino vejiga est\'omago cirug\'ia causar orina problema intestinal tracto urinario l\'iquido biliar abdominal \'organo delgado es\'ofago p\'elvico conducto hez \\
{\bf Healthcare} & provider health care doctor treatment healthcare symptom child medical talk condition diagnose check technology recommend exam professional diagnosis tooth \\
         & m\'edico proveedor atenci\'on tratamiento prueba s\'intoma examen hijo salud diagnosticar necesitar m\'edica ayudar profesional diagn\'ostico problema afecci\'on recomendar hablar \\
{\bf Hygiene \& Skin Care} & skin technology water hand hair body clean mouth exposure wash rash temperature wound heat avoid child dry wear remove \\
         & piel agua mano causar \'area peque\~no mantener boca color cabello aire exposici\'on temperatura evitar herida producto cuerpo zona calor \\

\end{longtable}


\renewcommand{\arraystretch}{1.2}
\footnotesize
\centering
\begin{longtable}{p{4cm} p{11cm}}
\caption{\texttt{ROSIE} 30-topics model} \label{tab:30_tpcs} \\

\toprule
\textbf{Topic} & \textbf{Description} \\
\midrule
\endfirsthead

\multicolumn{2}{c}{\textbf{Continuation of Table \ref{tab:30_tpcs}}} \\
\toprule
\textbf{Topic} & \textbf{Description} \\
\midrule
\endhead

\midrule
\multicolumn{2}{r}{\textit{Continued on next page}} \\
\midrule
\endfoot

\bottomrule
\endlastfoot

{\bf Healthcare Guidance} & provider care health healthcare doctor medical treatment symptom talk visit professional condition recommend question check child diagnose follow history  \\
         & médico proveedor atención tratamiento síntoma salud necesitar médica profesional examen hijo prueba hablar pregunta visita diagnosticar ayudar recomendar problema  \\
{\bf Cardiovascular System} & heart blood pressure artery vessel valve technology flow stroke clot vein body oxygen cardiac attack pulmonary coronary left lung   \\
         & corazón cardíaco sangre presión sanguíneo arterial arteria vaso válvula flujo vena coágulo ataque cuerpo oxígeno accidente pulmonar cerebrovascular cardíaca 
 \\
{\bf Genetic Syndromes} & syndrome gene condition genetic disorder mutation people affect individual change chromosome inherit family result occur develop factor abnormality associate   \\
         & síndrome gen genético trastorno mutación causar persona afección condición afectado afectar cambio individuo cromosoma caso tipo desarrollar característica factor  \\
{\bf Pediatric Healthcare} & treatment care clinic child team health mayo treat pediatric center hospital specialist medical patient doctor program include surgery disease 
  \\
         & tratamiento médico equipo mayo clinic niño atención cuidado especialista centro hospital paciente pediátrico tratar diagnóstico programa cirugía enfermedad terapia  \\
{\bf Childhood Vaccination} & vaccine dose vaccination age child receive influenza organization month person recommend health flu vaccinate administer adult virus immunization coverage 
  \\
         & vacuna dosis vacunación gripe niño año recibir persona edad mes administrar recomendar adulto virus vacunar cobertura serie sarampión inmunización  \\
{\bf Symptoms of Illness} & symptom pain feel sign severe reaction headache fever allergy experience include mild allergic asthma nausea common occur vomiting cough \\
         & síntoma dolor grave signo causar reacción sentir cabeza fiebre alergia experimentar dificultad leve hinchazón alérgico vómito asma médico náusea  \\
{\bf Treatment Options} & patient treatment study clinical therapy trial drug result evidence receive regimen organization disease month follow diagnosis milligram report evaluate 
  \\
         & paciente tratamiento clínico terapia ensayo diagnóstico caso recibir enfermedad evidencia mes evaluar demostrar fármaco mg observar fase inicial tasa \\
{\bf Global Disease Prevention and Control} & health care country risk outbreak prevention program transmission report united control cdc reduce public include prevent person disease strategy \\
         & salud caso país prevención riesgo programa unidos brote enfermedad transmisión reducir cdc persona control estrategia público atención alto servicio  \\
{\bf Hygiene} & water technology tooth hand clean mouth wash temperature child dental avoid heat remove exposure air prevent wear chemical product 
  \\
         & agua diente mano dental boca mantener aire producto evitar temperatura exposición calor ropa químico lavar contener niño baño frío  \\
{\bf Nutrition} & food eat level diet blood sugar vitamin fat drink body healthy insulin glucose technology diabetes calcium cholesterol supplement product 
  \\
         & alimento comer nivel dieta sangre azúcar vitamina grasa diabetes cantidad insulina saludable alto glucosa comida cuerpo ayudar calcio beber \\
{\bf Reproductive Health} & technology hormone woman control period vaginal method birth uterus sex vagina egg menstrual bleeding change male female pill body 
  \\
         & mujer hormona método útero vaginal sexual anticonceptivo uterino vagina período sangrado hormonal menstrual hombre cuello ovario pene estrógeno testículo  \\
{\bf Pregnancy} & pregnancy birth woman baby pregnant risk health defect bear infant increase delivery week organization mother technology fetal fetus congenital 
  \\
         & embarazo bebé mujer embarazado parto riesgo nacimiento nacer defecto semana madre nacido problema aumentar probabilidad feto prematuro fetal quedar  \\
{\bf Sexually Transmitted Infections} & infection hiv person virus risk hepatitis organization health infect hpv sex transmission partner exposure contact transmit sexual testing positive 
 \\
         & infección persona vih virus sexual riesgo prueba hepatitis transmisión infectado vph tratamiento exposición contacto pareja herpes sífilis anticuerpo prevenir  \\
{\bf Mental Health Disorders} & disorder health alcohol mental people depression anxiety behavior symptom stress drug opioid technology treatment include organization experience change life 
  \\
         & trastorno problema persona alcohol salud mental depresión ansiedad síntoma comportamiento tratamiento estrés droga vida consumo sustancia afectar físico cambio  \\
{\bf Medical Imaging} & blood image sample diagnose doctor ray tomography result ultrasound detect biopsy magnetic resonance check technology diagnosis measure imaging exam 
  \\
         & prueba examen sangre médico muestra diagnóstico análisis amágenes detectar diagnosticar biopsia laboratorio radiografía determinar mostrar medir detección magnético tomografía  \\
{\bf Infant Care} & baby month technology hospital week day breast infant milk care time breastfeed unit child stay feed newborn start hour \\
         & bebé mes semana hospital leche necesitar unidad materno lactancia cuidado nacido hora recibir alimentación mayoría comenzar amamantar recuperación casa  \\
{\bf Infectious Diseases} & infection antibiotic bacteria disease virus people spread technology illness person common treat bacterial animal human respiratory fever infect prevent 
  \\
         & infección enfermedad causar bacteria persona antibiótico virus grave común animal humano respiratorio fiebre causado caso propagar neumonía infectado tratar  \\
{\bf Gender Disparities in Health} & risk age increase woman factor health study rate death adult child report percent associate organization obesity estimate cost prevalence 
  \\
         & riesgo año mujer edad factor niño alto aumentar tasa adulto muerte hombre aumento obesidad menor grupo persona unidos nivel  \\
{\bf Skin Conditions} & skin eye technology vision hair light color rash red loss body change form common layer contact patch spot sun 
  \\
         & piel ojo visión ocular color causar luz cabello rojo pequeño área aparecer pérdida capa forma retina erupción lesión común  \\
{\bf Chronic Health Conditions} & disease people condition health chronic organization risk life technology symptom affect complication develop diabetes severe treatment common person illness 
  \\
        & enfermedad persona afección vida riesgo diabetes condición crónico grave síntoma complicación afectar desarrollar tipo común año problema causa sistema  \\
{\bf Orthopedic Health} & bone joint injury foot technology pain surgery fracture leg arm knee hand hip spine muscle shoulder finger head tissue 
  \\
         & hueso lesión pie articulación dolor dedo cirugía fractura óseo pierna brazo rodilla columna mano cadera vertebral músculo hombro tejido  \\
{\bf Medication and Treatment} & medication medicine treat doctor treatment drug prescribe technology reduce pain prevent prescription control talk include class counter injection risk 
  \\
         & medicamento tomar médico tratar tratamiento secundario ayudar reducir medicamentos funcionar recetar dolor prevenir controlar prescribir aliviar dejar disminuir llamado  \\
{\bf Surgical Procedures} & surgery procedure tube remove technology bladder surgeon intestine stomach surgical incision tissue catheter urinary bowel insert perform tract esophagus 
  \\
         & cirugía intestino tubo vejiga cirujano estómago pequeño quirúrgico delgado incisión tejido catéter esófago intestinal abdomen urinario colon colocar orina  \\
{\bf Organ Failure} & kidney liver lung disease technology transplant blood body organ fluid damage smoke failure chronic urine renal function airway duct 
  \\
         & renal enfermedad riñón pulmón hígado pulmonar trasplante hepático respiratorio fumar órgano causar sangre líquido cuerpo insuficiencia daño biliar vía  \\
{\bf Child Development} & child technology parent school family time organization talk feel kid health learn age friend adult support understand play teen 
  \\
         & niño hijo padre ayudar familia escuela hablar importante sentir adolescente necesitar problema edad año pequeño cosa adulto aprender amigo  \\
{\bf Cancer Treatment} & cancer tumor treatment cell radiation breast therapy gland thyroid chemotherapy technology stage surgery spread hormone body treat prostate lymph 
  \\
         & cáncer tumor tratamiento célula tipo glándula canceroso quimioterapia mama cirugía radiación radioterapia linfático cuerpo hormona tejido estadio terapia ganglio  \\
{\bf Genetic Enzyme Functions} & cell protein blood acid body gene function immune produce normal enzyme deoxyribonucleic red anemia tissue role bone process marrow 
  \\
         & célula proteína ácido cuerpo llamado producir gen glóbulo función sistema normal anemia celular desoxirribonucleico enzima inmunitario rojo tipo producción  \\
{\bf Nervous System} & brain nerve muscle ear seizure loss spinal cord movement technology nervous damage affect injury hear control hearing body epilepsy 
  \\
         & cerebro cerebral nervioso nervio pérdida oído convulsión problema causar muscular espinal movimiento sistema músculo lesión médula afectar daño auditivo  \\
{\bf Medication Administration} & technology doctor day medicine sleep dose hour medication time injection follow intravenous inject mouth direct tablet pharmacist prescription label 
  \\
         & medicamento tomar médico hora dosis inyección sueño vía oral administrar instrucción indicar frecuencia dormir líquido tome recibir inyectar farmacéutico  \\
{\bf Physical Well-being} & activity exercise physical weight technology muscle healthy time reduce body change improve avoid prevent injury sport walk stress rest 
  \\
         & actividad ejercicio ayudar peso físico mantener reducir mejorar músculo evitar saludable lesión estilo cambio prevenir caminar activo aumentar deporte  \\
                                                                                          
\end{longtable}

\renewcommand{\arraystretch}{1.2}
\footnotesize
\centering
\begin{longtable}{p{4cm} p{11cm}}
\caption{\texttt{WIKI-EN-DE} 25-topics model} \label{tab:25_tpcs} \\

\toprule
\textbf{Topic} & \textbf{Description} \\
\midrule
\endfirsthead

\multicolumn{2}{c}{\textbf{Continuation of Table \ref{tab:25_tpcs}}} \\
\toprule
\textbf{Topic} & \textbf{Description} \\
\midrule
\endhead

\midrule
\multicolumn{2}{r}{\textit{Continued on next page}} \\
\midrule
\endfoot

\bottomrule
\endlastfoot

{\bf Geographical Regions and Waterways} & river florida west land north border technology county lake united south water mile include virginia east region delaware canada \\
         & florida staat river grenze gebiet land vereinigt county usa virginia stadt liegen westen kanada region delawar siedler south mississippi \\
         
{\bf U.S. Presidential Politics} & trump bush president clinton obama ford campaign office house presidential ultrasound election january white presidency official percent donald report \\
         & trump bush präsident clinton ford obama trumps januar usa haus prozent weiß präsidentschaft november donald nixon dezember oktober ehemalig \\
         
{\bf American Historical Figures and Literature} & franklin write hamilton john jefferson lincoln poe technology washington thomas publish health organization adams benjamin philadelphia james york time \\
         & franklin john schreiben hamilton lincoln jefferson poe washington thomas veröffentlichen benjamin adams james philadelphia alexander william erscheinen werk rede \\
         
{\bf Christian Communion Practices} & church religious technology book catholic prayer anglican religion communion god common christ christian eucharistic lord bread service faith supper \\
         & kirche religiös anglikanisch religion katholisch buch heilig gemeinsam brot gebet glaube eucharistie jahrhundert abendmahl kommunion verwenden christi christlich england  \\
         
{\bf French Revolution and Political Turmoil} & health organization lafayette french day king paris technology louis july revolution death france directory assembly people national return soldier  \\
         & französisch lafayette de juli paris frankreich soldat revolution juni april könig september oktober la louis november mai august lassen \\
         
{\bf Freemasonry Traditions} & lodge grand freemasonry masonic technology map century distillation form master masons lodges process doi jurisdiction degree freemason body object  \\
         & lodge grand freimaurerei freimaurer mitglied entwicklung karte verwenden jahrhundert verschieden loge de log destillation organisation bezeichnen entwickeln alt england  \\
         
{\bf American Colonial Resistance and British Rule} & british colony american parliament boston england king independence colonial government north lord george english colonist britain massachusetts tea london  \\
         & kolonie britisch amerikanisch boston parlament lord england könig kolonist großbritannien regierung george massachusett englisch london kolonial loyalist unabhängigkeit act  \\
         
{\bf Military Leadership and Command Structure} & army military officer chief president commander force appoint rank serve service war armed command minister united forces head staff \\
         & präsident armee ernennen dienen offizier general streitkraft oberbefehlshaber dienst rang militärisch army krieg vereinigt position united tragen artikel aufgabe  \\
         
{\bf Epilepsy and Seizure Disorders} & epilepsy blood seizure technology disease treatment include condition people occur medical death result barton risk health suffer time illness \\
         & epilepsie krankheit fall führen behandlung anfälle barton verursachen person ursache anfall schwer medizinisch arzt zeigen tod patient form kind \\
         
{\bf U.S. Foreign Military Policy} & united war eisenhower ultrasound policy foreign regulation roosevelt council country military american commission organization union adopt proposal march international \\
         & usa vereinigt staat eisenhower roosevelt amerikanisch regierung land krieg international märz erklären nation führen beziehung militärisch china dezember politik  \\
         
{\bf George Washington's Family and Descendants} & washington george family die father health bear son organization virginia child county wife brother death marry daughter house charles  \\
         & washington george sterben vater familie sohn virginia kind leben county tod frau gebären haus tochter bruder john mutter lee  \\
         
{\bf Vocational Education and Training} & woman school black education health organization white training washington age carver technology civil american era americans time african life \\
         & frau schule schwarz weiß bildung ausbildung carver washington arbeit leben helfen amerikaner arbeiten organisation beginnen fähigkeit öffentlich schüler gruppe  \\
         
{\bf Slavery and Slave Trade} & slave trade slavery population african people century health dowry black organization enslave southern family property european africa technology south  \\
         & sklave sklaverei bevölkerung sklavenhandel jahrhundert prozent afrikanisch land mitgift europäisch versklavt afrikaner eigentum barbado zahl familie schwarz leben handel \\
         
{\bf Political Revolution and Social Change} & political technology american people historian view revolution government idea power social health organization argue time influence century change lead  \\
         & politisch amerikanisch historiker revolution regierung volk freiheit idee glauben jahrhundert argumentieren rolle politik ansicht sehen sozial stark frage gesellschaft \\
         
{\bf European Wars and Treaties} & fort york washington arnold british french city war allen indian lead expedition force wayne quebec organization ohio indians health  \\
         & krieg frankreich großbritannien französisch britisch amerikanisch vereinigt vertrag staat spanien deutsch spanisch reich führen amerika karl armee königreich könig \\
         
{\bf Colonial Military Conflicts} & war france french britain british treaty american united spain empire german charles republic america spanish peace germany independence power  \\
         & fort washington york arnold brite indianer wayne general expedition franzose quebec city britisch mai stadt juli ohio indisch französisch  \\
         
{\bf United States Constitutional Development} & congress constitution convention house delegate government united committee technology power vote virginia elect propose continental confederation federal plan amendment  \\
         & staat verfassung kongreß kongress delegierter vereinigt wählen mitglied virginia senat mason kontinentalkongress änderung konvent verabschieden artikel regierung präsident madison \\
         
{\bf American Revolutionary War} & british battle army troop washington american soldier force attack continental militia regiment command americans cornwallis march retreat wound lead  \\
         & truppe schlacht britisch washington amerikanisch armee brite soldat general amerikaner regiment miliz angriff kontinentalarmee howe cornwallis offizier führen kommando \\
         
{\bf Naval Warfare in the Age of Sail} & ship french british fleet navy island battle naval admiral rodney royal sail port technology grasse war captain return april  \\
         & schiff de französisch britisch flotte marine rodney schlacht admiral royal navy linie island insel hafen august april farragut grasse  \\

{\bf American Historical Publications} & american isbn history press book university york volume revolution america george life film historical publish david edition german archive  \\
         & geschichte amerikanisch isbn buch revolution york band george presse american oclc amerika auflage david deutsch leben press universität john \\

{\bf American Higher Education and Academic Societies} & university college school boston american society gray study include award student united art academy honor receive graduate science william  \\
         & university college boston universität gray school american erhalten mitglied william society gründen art national spielen academy harvard besuchen willard  \\
         
{\bf Economic and Financial Policy} & company technology pay government tax debt money increase carnegie business economic bell bank fund financial cost dollar budget reduce  \\
         & million dollar unternehmen geld carnegie bell regierung bank kosten schuld erhöhen erhalten zahlen company steuer wirtschaft gründen milliarde bezahlen   \\
         
{\bf Historic Architecture and Infrastructure} & city washington national building baltimore york memorial bridge park technology street hall george build albany house historic alexandria construction  \\
         & stadt washington national baltimore park york gebäude george street city albany bauen befinden alexandria brücke hall memorial bridge denkmal  \\
         
{\bf Constitutional Law and Amendments} & court law supreme amendment federal technology united government clause rule justice judge protection decision legal person hold constitution constitutional  \\
         & staat gesetz gerichtshof oberer fall gericht vereinigt entscheiden richter änderung person verbieten entscheidung änderungsantrag regierung erklären act staatlich urteil  \\

{\bf Presidential Elections} & president party election vote republican organization johnson democratic health candidate office republicans jackson pierce presidential win support senate elect 
 \\
         & präsident partei wahl republikaner johnson wählen gewinnen jackson republikanisch pierce demokratisch demokrat arthur stimme senat kandidat politisch coolidge staat \\

\end{longtable}

\end{document}